\documentclass[journal]{IEEEtran}
\ifCLASSINFOpdf
\else
\fi
\usepackage{url}
\usepackage[T1]{fontenc}
\usepackage{titlesec}

\usepackage{multirow}
\usepackage{adjustbox}
\usepackage{float}
\usepackage[font=footnotesize]{caption}
\usepackage{xcolor}
\usepackage{cite}
\usepackage{amsmath,amssymb,amsfonts}
\usepackage{algorithm}
\usepackage{algpseudocode}
\algnewcommand\algorithmicforeach{\textbf{for each}}
\algdef{S}[FOR]{ForEach}[1]{\algorithmicforeach\ #1\ \algorithmicdo}
\algblockdefx[ForP]{ForP}{EndForP}[1]{\textbf{for each} #1 \textbf{do in parallel}}{\textbf{end for}}
\usepackage{longtable}
\usepackage{lettrine}

\usepackage{graphicx}
\usepackage{subfig}
\usepackage[figuresright]{rotating}
\usepackage{textcomp}

\usepackage{tabularx,booktabs}
\newcolumntype{C}{>{\centering\arraybackslash}X} % centered version of "X" type
\usepackage{lipsum}
\usepackage{etoolbox}
\usepackage{tikz}
\usetikzlibrary{tikzmark}
\usetikzlibrary{calc}

\errorcontextlines\maxdimen

\newcommand{\ALGtikzmarkcolor}{black}% customise this, if you want
\newcommand{\ALGtikzmarkextraindent}{4pt}% customise this, if you want
\newcommand{\ALGtikzmarkverticaloffsetstart}{-.5ex}% customise this, if you want
\newcommand{\ALGtikzmarkverticaloffsetend}{-.5ex}% customise this, if you want
\makeatletter
\newcounter{ALG@tikzmark@tempcnta}

\newcommand\ALG@tikzmark@start{%
    \global\let\ALG@tikzmark@last\ALG@tikzmark@starttext%
    \expandafter\edef\csname ALG@tikzmark@\theALG@nested\endcsname{\theALG@tikzmark@tempcnta}%
    \tikzmark{ALG@tikzmark@start@\csname ALG@tikzmark@\theALG@nested\endcsname}%
    \addtocounter{ALG@tikzmark@tempcnta}{1}%
}

\def\ALG@tikzmark@starttext{start}
\newcommand\ALG@tikzmark@end{%
    \ifx\ALG@tikzmark@last\ALG@tikzmark@starttext
    \else
        \tikzmark{ALG@tikzmark@end@\csname ALG@tikzmark@\theALG@nested\endcsname}%
        \tikz[overlay,remember picture] \draw[\ALGtikzmarkcolor] let \p{S}=($(pic cs:ALG@tikzmark@start@\csname ALG@tikzmark@\theALG@nested\endcsname)+(\ALGtikzmarkextraindent,\ALGtikzmarkverticaloffsetstart)$), \p{E}=($(pic cs:ALG@tikzmark@end@\csname ALG@tikzmark@\theALG@nested\endcsname)+(\ALGtikzmarkextraindent,\ALGtikzmarkverticaloffsetend)$) in (\x{S},\y{S})--(\x{S},\y{E});%
    \fi
    \gdef\ALG@tikzmark@last{end}%
}

\apptocmd{\ALG@beginblock}{\ALG@tikzmark@start}{}{\errmessage{failed to patch}}
\pretocmd{\ALG@endblock}{\ALG@tikzmark@end}{}{\errmessage{failed to patch}}
\makeatother

\begin{document}
%\userpackage{caption}
\title{}
\title{Machine Learning for Autonomous Vehicle's Trajectory Prediction: A comprehensive survey, Challenges, and Future Research Directions} 
\author{Vibha Bharilya, and Neetesh Kumar,~\IEEEmembership{Member, IEEE} % <-this % stops a space
        \thanks{Vibha Bharilya is with Department of Computer Science and Engineering, Indian Institute of Technology, Roorkee, Uttarakhand- 247667, India (e-mail: vibha\_b@cs.iitr.ac.in).}
        
\thanks{Neetesh Kumar is with Department of Computer Science and Engineering, Indian Institute of Technology, Roorkee, Uttarakhand- 247667, India (e-mail: neetesh@cs.iitr.ac.in).}% <-this % stops a space
%\thanks{This work is supported by SERB, DST, Govt. of India under the Project ``Grant no: EEQ/2019/000182, Software defined controlled and dynamic traffic load balanced scheduling framework for IoT enabled Intelligent Transportation System (ITS)".}% <-this % stops a space
}

\maketitle

\begin{abstract}
The significant contribution of human errors, accounting for approximately 94\% (with a margin of ±2.2\%), to road crashes leading to casualties, vehicle damages, and safety concerns necessitates the exploration of alternative approaches. Autonomous Vehicles (AVs) have emerged as a promising solution by replacing human drivers with advanced computer-aided decision-making systems. However, for AVs to effectively navigate the road, they must possess the capability to predict the future behavior of nearby traffic participants, similar to the predictive driving abilities of human drivers. Building upon existing literature is crucial to advance the field and develop a comprehensive understanding of trajectory prediction methods in the context of automated driving. To address this need, we have undertaken a comprehensive review that focuses on trajectory prediction methods for AVs, with a particular emphasis on machine learning techniques including deep learning and reinforcement learning-based approaches. We have extensively examined over two hundred studies related to trajectory prediction in the context of AVs. The paper begins with an introduction to the general problem of predicting vehicle trajectories and provides an overview of the key concepts and terminology used throughout. After providing a brief overview of conventional methods, this review conducts a comprehensive evaluation of several deep learning-based techniques. Each method is summarized briefly, accompanied by a detailed analysis of its strengths and weaknesses. The discussion further extends to reinforcement learning-based methods. This article also examines the various datasets and evaluation metrics that are commonly used in trajectory prediction tasks. Encouraging an unbiased and objective discussion, we compare two major learning processes, considering specific functional features. By identifying challenges in the existing literature and outlining potential research directions, this review significantly contributes to the advancement of knowledge in the domain of AV trajectory prediction. Its primary objective is to streamline current research efforts and offer a futuristic perspective, ultimately benefiting future developments in the field.

% The development of autonomous vehicles (AVs) has led to the need for accurate trajectory prediction to ensure safe and efficient driving. Machine learning (ML) has emerged as a promising solution for predicting the future trajectories of objects in the environment. This paper presents a survey of the state-of-the-art ML-based trajectory prediction methods for AVs.The paper begins with an introduction to the general problem of predicting vehicle trajectories and provides an overview of the key concepts and terminology used throughout. After a brief overview of conventional methods, we comprehensively evaluate machine learning-based approaches, including deep learning and reinforcement learning based-methods. The article also examines the performance of several well-known solutions, identifies the challenges present in the literature, and outlines potential new research directions.
 %The performance of several well-known solutions is discussed, and research gaps in the literature are identified. Finally, potential new research directions are outlined.

\end{abstract}

% Note that keywords are not normally used for peer review papers.

\IEEEpeerreviewmaketitle
\section{Introduction}
\lettrine[findent=2pt]{\textbf{A}}{nnually}, approximately 1.35 million deaths occur due to road crashes, with 1,140 reported deaths in 2018 according to the Australian Automobile Association (AAA) \cite{123}. There were 1,194 fatal car accidents in 2022 in Australia. This is an increase of 5.8\% from 2021. National fatalities have stayed basically flat during the past ten years \cite{345}. In the United States, the NHTSA's investigation reveals that around 94\% of severe road crashes can be attributed to driver errors [4]. Further, human error is consistently identified as a major factor in road crashes, emphasizing the need to address this preventable distress. To assist human drivers in avoiding critical situations, advanced motorized vehicles employ Advanced Driver Assistance Systems (ADAS), which have rapidly evolved since their inception in the 1950s. Researchers are actively exploring the efficiency of ADAS in warning drivers and preventing crashes. The rapid technological progress, including the use of high-end sensors, powerful machine learning techniques, and innovations from companies like Google and Tesla Motors, has significantly impacted the automation industry. Automotive and tech companies have demonstrated the feasibility of Automated Driving Systems (ADS) through successful test fleet operations. The Society of Automotive Engineers (SAE) classifies ADS into six levels of vehicle automation \cite{sae2018taxonomy}, with a focus on full automated operation. 

Autonomous Vehicles (AVs) are expected to play a significant role in reducing crashes and enhancing road safety in the foreseeable future.
The rapid development of perception, planning, and control systems for AVs in recent years is noteworthy. However, the production of AVs in large quantities will not be feasible until their safety is fully established. One of the critical technologies in AVs is the ability to forecast the future states of the surrounding environment in real time, as human drivers can. This capability will further enhance safety measures.
Before beginning a new driving operation, such as acceleration or a lane change, a human driver typically scans the surrounding traffic to predict how it will behave in the future. Future trajectories can be used to model future traffic participant states, which can then be used to construct decision-making or planning algorithms as well as to foresee potential dangers. However, accurately predicting future traffic participant trajectories is attracting a lot of attention and is quickly becoming one of the key points to improving the safety of autonomous driving. This is because of the variety of maneuvers that traffic participants make, the complex interactions between traffic participants and environments, the uncertainty of sensory information, the computation burdens, and the computing time requirements of AVs.
\subsection{Motivation}
Numerous techniques have been developed for the trajectory prediction, and several scholars have pursued this area of research. Some of the review papers have covered various trajectory prediction techniques, in the same line, Lefèvre \textit{et al.} \cite{lefevre2014survey} provided an analysis of motion prediction and risk assessment techniques used for AVs before 2014. Mohammad \textit{et al.} \cite{shirazi2016looking} discussed strategies for behavior prediction at crossings based on drivers' actions. Further, Mozaffari \textit{et al.} \cite{mozaffari2020deep} offered a review of deep learning-based approaches focused on vehicle behavior analysis. Leon \textit{et al.} \cite{leon2021review} and Liu \textit{et al.} \cite{liu2021survey} wrote reviews on trajectory prediction for AVs, where, Leon \textit{et al.} covered deep learning and stochastic methods, and Liu \textit{et al.} focused solely on deep learning methods. Karle \textit{et al.} \cite{karle2022scenario} offered three distinct prediction models as a classification of these models and compared them based on the underlying study methodology. Gomes \textit{et al.} \cite{gomes2022review} reviewed the literature on Intention-Aware and Interaction-Aware trajectory prediction for autonomous vehicles and examined how maneuver goals and their interaction with other maneuvers affect the performance of trajectory prediction techniques. In Ghorai \textit{et al.} \cite{ghorai2022state}, a survey covered the identification and monitoring of dynamic agents and objects encountered by an autonomous ego vehicle. The main topics of the review were delved into 2D and 3D dynamic object identification techniques based on DL employed in AV research. Huang \textit{et al.} \cite{huang2022survey} thoroughly examined trajectory prediction techniques for AVs put forth over the last two decades, excluding vision-based techniques. Recently, Benrachou \textit{et al.} \cite{benrachou2022use} encompassed research on both data-driven and model-based algorithms, which aim to forecast the movement of surrounding traffic. Table I provides a summary of the related state-art-of surveys, along with the different categorizations approaches, and contributions.   
\newline
\newline
 Motion prediction involves anticipating the behavior, maneuvers, or trajectory of an object, depending on the desired level of abstraction. The term "behavior" encompasses general actions and their execution style, such as "following the road and maintaining a safe distance." On the other hand, "maneuvers" refer to discrete actions that an object can perform without requiring a detailed specification, such as "turning right." Trajectories, on the other hand, provide the most detailed type of prediction by describing an object's position over discrete time steps \cite{karle2022scenario}. Previous surveys have predominantly emphasized motion prediction and behavior prediction in the realm of Autonomous Vehicles (AVs). Further, other state-of-the-art surveys are mixed of trajectory prediction for vehicles and pedestrians. A comprehensive and dedicated review on autonomous vehicles trajectory prediction accounting machine learning methods is remained relatively unexplored. Furthermore, several advancements in the domain of trajectory prediction, in recent years, including computer vision-based methods, reinforcement learning etc., have not been addressed in the existing surveys which are also needed to be explored. Consequently, there exists substantial potential for further exploration and investigation within this domain. Thus, the motivation behind writing this survey paper is to actively contribute to the research in the trajectory prediction field specifically for AVs. 
\begin{figure}[]
    \centering
    \includegraphics[width=0.5\textwidth]{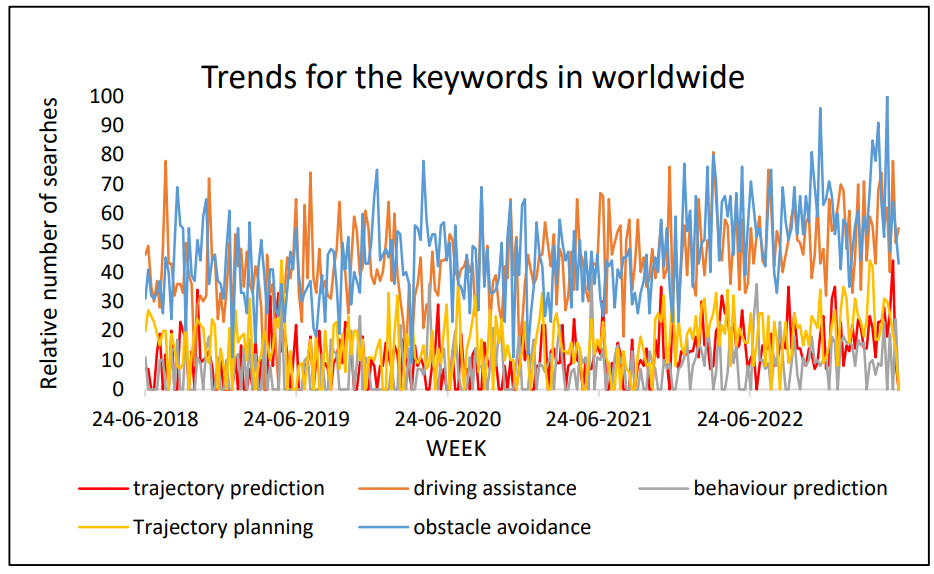}
    \caption{Google trends for specific keywords.}
    \label{fig:my_label}
\end{figure}
\begin{figure}[]
    \centering
    \includegraphics[width=0.5\textwidth]{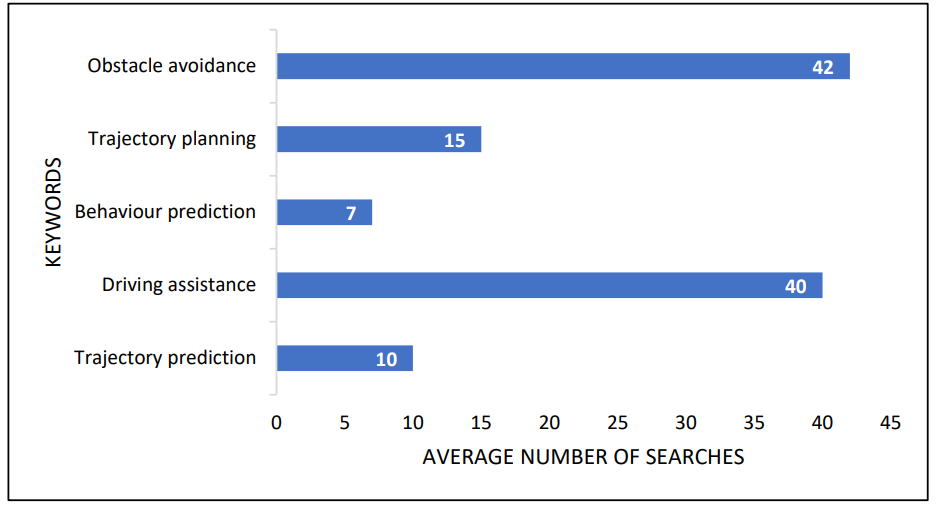}
    \caption{Average count for the specific keywords related to trajectory prediction.}
    \label{fig:my_label}
\end{figure}
\subsection{Google Trends}
In recent years, Autonomous Driving (AD) has become increasingly popular in the automotive industry. Prominent automobile manufacturers, including Tesla, General Motors, and BMW, have made significant investments and focused on trajectory prediction and related technologies for Autonomous Vehicles (AVs) for the development of AD technology. The worldwide search trends for keywords related to AD, such as trajectory prediction, driving assistance, behavior prediction, trajectory planning, and obstacle avoidance, are illustrated in Fig. 1 and Fig. 2. In Fig. 1, the comparison of related keywords demonstrates changes over the same time period. It shows that driving assistance and obstacle avoidance generate similar levels of excitement worldwide. However, within the field of AD, there have been recent advancements in keywords such as trajectory prediction, trajectory planning, and behavior prediction. Notably, trajectory prediction focuses on a more specific domain within autonomous driving. In Fig. 2, the average number of searches related to the keyword worldwide is depicted. Notably, driving assistance and obstacle avoidance keywords receive a higher number of searches compared to other terms like trajectory prediction, trajectory planning, and behavior prediction. Trajectory prediction is currently evolving in the field of AD, indicating increasing interest and development in this area.

\begin{table*}
 \caption{Summary of related state-art-of the surveys: Related work, the title of the work, categorization of each work and their contribution  }
    \centering

   \resizebox{17.5cm}{9cm}{\begin{tabular}{|p{0.5cm}|p{1.8cm}|p{3 cm}|p{5 cm}|p{5 cm}|}
     \hline
        \textbf{S.No.}&\textbf{Ref. \& Year} & \textbf{Topics Discussed} &\textbf{Categorisation}&\textbf{Contributions} \\
        \hline
         1. & \cite{mozaffari2020deep}  \& 2020 & Deep Learning-Based Vehicle Behavior Prediction for Autonomous Driving Applications: A Review & Classification based on three criteria: input representation, output type, and prediction method.&  An assessment of contemporary deep learning techniques for predicting vehicle behavior.\\
%It lists various standards for categorizing a subset of common techniques based on input and output data but excludes certain recently released techniques.
         2. & \cite{leon2021review} \& 2021 & A Review of Tracking and Trajectory Prediction Methods for
Autonomous Driving & Categorized based on its primary prediction approach, including neural networks, stochastic methods, and hybrid techniques. & This article offers a study of tracking and trajectory prediction approaches that only include deep learning and stochastic methodology methods. \\
         3. & \cite{liu2021survey} \& 2021 & A Survey on Deep-Learning Approaches for Vehicle Trajectory
Predictions in Autonomous Driving & Grouped according to the manner in which data is represented, learning techniques utilized, and objective functions employed.&  Examine and characterize current deep learning-based trajectory forecasting techniques in this work.\\
%& Classify current learning-based methods for trajectory prediction based on their representation, modeling, and learning approaches.
         4. &\cite{karle2022scenario} \& 2022 & Scenario Understanding and Motion Prediction for
Autonomous Vehicles—Review and Comparison & A categorization of prediction models are presented, which includes physics-based, pattern-based, and planning-based approaches.& Examine and compared the three specific prediction methods and Demonstrate a trade-off between holism and interpretability in contemporary approaches.\\
%and Three distinct prediction models are offered as a classification of these models, with comparisons drawn between them based on the underlying study methodology.
         5. & \cite{gomes2022review} \& 2022 & A Review on Intention-aware and Interaction-aware Trajectory
Prediction for Autonomous Vehicles & The current classification of prediction methods is framed within the context of intention-aware and interaction-aware trajectory prediction techniques.& Examine and characterize the physics-based methods and ML techniques.\\

% into Intention-Aware and Interaction-Aware trajectory prediction for autonomous vehicles is the subject of this paper's review of the literature, which examines the  papers released since 2008.\\
         6. &\cite{ghorai2022state} \& 2022 & State Estimation and Motion Prediction of Vehicles and Vulnerable Road Users for Cooperative Autonomous Driving: A Survey & A categorization of prediction models are presented into three classes: physics-based, maneuver-based, and interaction aware motion models. & The major focus of this review is on perception sensors and navigation.\\
         % sensorsThe identification and monitoring of dynamic agents and objects that an autonomous ego vehicle comes into contact with, as well as the evaluation of their states and intentions, are the main topics of this review.\\
         % A Survey & This study focuses on Motion
% prediction and state estimation methods of vehicles and VRUs and cover deep learning methods\\
         7. & \cite{huang2022survey} \& 2022 & A Survey on Trajectory-Prediction Methods
for Autonomous Driving & Classification based on physics-based
methods, the classic machine learning-based methods, the deep
learning-based methods, and reinforcement learning-based
methods & Choose both heuristic and contemporary trajectory prediction techniques for a specific timeframe and provide a comparative summary.\\
8. & \cite{benrachou2022use} \& 2022 & Use of Social Interaction and Intention to Improve Motion Prediction Within Automated Vehicle Framework: A Review &Motion prediction solutions should
be categorized into four principal strategies: (1) Prediction
methods (intention-aware or interaction-aware), (2) Classes
(model-based or data-driven), (3) Algorithms (which type of model), and (4) Datasets (classified according to the point of
view: top-down view data or vehicle-view data).

 & Cover the machine learning techniques and the primary emphasis of this paper has been on approaches that are intention-aware or incorporate interaction-awareness.\\
 %categorization of existing motion prediction solutions based on 1. Prediction methods, 2. Classes, 3. Algorithms \\
         9. &  \textbf{This survey}  &Machine Learning for Autonomous Vehicle's Trajectory Prediction: A comprehensive survey, Challenges, and Future Research Directions &  The categorization involves the utilization of conventional methods, computer vision-based methods, cutting-edge deep learning techniques, and reinforcement learning-based methods.& This assessment presents a concise overview of traditional methods, computer vision-based approaches, and conducts a comprehensive assessment of prevalent deep learning and reinforcement learning-based techniques employed in trajectory prediction for autonomous vehicles.
         Additionally, it includes a discussion of the advantages and disadvantages associated with these methods.\\
          \hline
    \end{tabular}
   }
    \label{tab:my_label}
\end{table*}

\subsection{Key Contributions}
  This comprehensive survey on the state-of-the-art machine learning-based trajectory prediction methods for Autonomous Vehicles (AVs) provides a taxonomy of the different approaches, as shown in Fig. 3, including conventional methods, deep learning-based methods, and reinforcement learning-based methods, and discusses the advantages and limitations of each method. This study focuses on vehicle trajectory prediction algorithms, as other traffic participants, such as adjacent vehicles, directly affect the ego vehicle. 
  In the end, the paper highlights the challenges and future research directions in this field. 
 The significant contributions of this survey are enlisted as follows:
\begin{enumerate}
   \item This survey offers an empirical study on autonomous vehicles trajectory prediction methods and extensively focuses on machine learning-based methods. For the better understanding, an overview on AV's trajectory prediction problem, related terminology, and conventional methods are also briefly provided.  
    \item A concise assessment of conventional methods such as Physics-based methods, Sampling methods, and Probabilistic models in trajectory prediction is presented, along with a discussion of their advantages and disadvantages.

\item  A comprehensive evaluation is provided for the prevalent deep learning and reinforcement learning based-methods used in trajectory prediction for autonomous vehicles.
 
\item  An analytical summary is provided for the metrics and datasets used to evaluate the performance of trajectory prediction methods.

\item  A comparison of the methods is conducted, analyzing the strengths and weaknesses of each approach. Furthermore, challenges and potential research avenues are identified.
\end{enumerate}
 
%In brief, this survey serves as a useful reference for researchers and practitioners working on trajectory prediction for autonomous vehicles. 
\begin{figure*}[hbt!]
    \centering
     
     \includegraphics[width=1.0\textwidth] {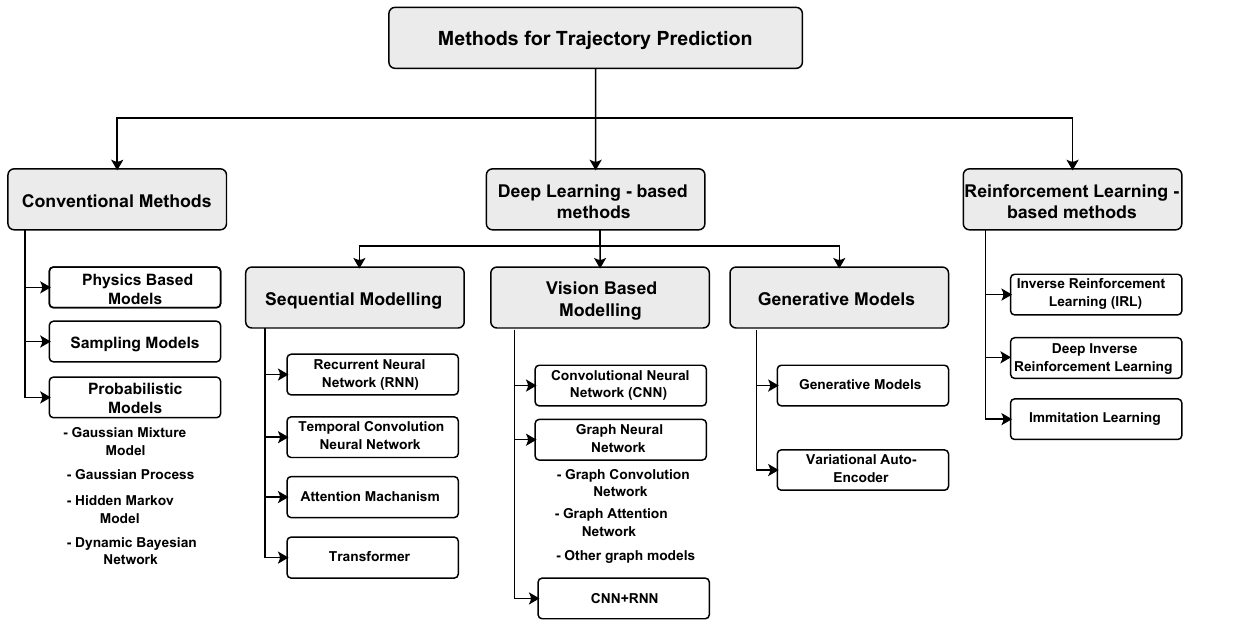}
    \caption{Categorization of methods for trajectory prediction task.}
    \label{fig:my_label}
\end{figure*}

\subsection{Paper Organization}
 The road map of the survey has been presented in pictorial form in Fig. 4. There are nine sections in this paper. Section II presents a generic problem formulation, provides definitions of the terminologies used, and the methods are categorized based on various criteria. In Sections III, IV, and V of the paper, comprehensive reviews on conventional-based methods, deep learning-based methods, and reinforcement learning-based methods are conducted respectively. Section VI discusses the commonly used evaluation metrics and datasets. Section VII discusses the performance of different methods, and Section VIII highlights the current challenges in
the literature and potential new research directions. The key concluding remarks are given in section IX.

\begin{figure}[hbt!]
      
     \includegraphics[width=0.5\textwidth] {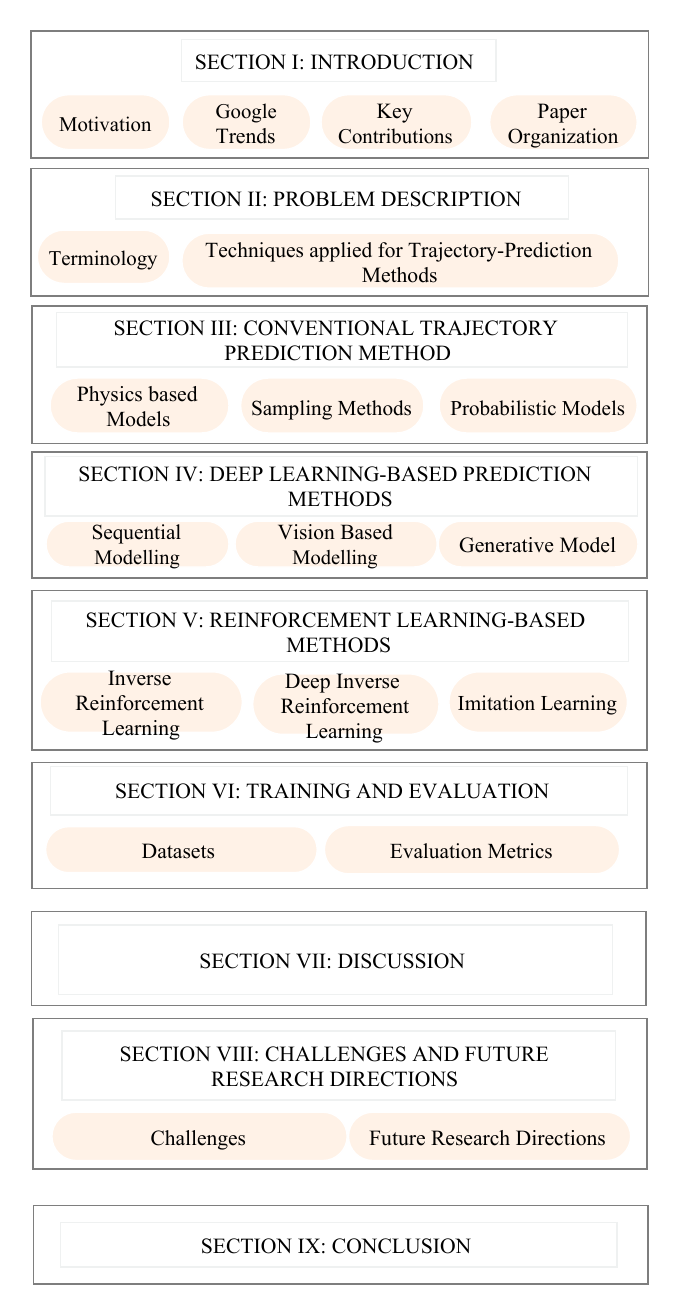}
    \caption{Road Map of the Paper.}
    \label{fig:my_label}
\end{figure}

\section{Problem Description }
\label{prelims}
In the context of Automated Driving (AD), accurately predicting the trajectories of other road users poses a significant challenge for AV's software. It requires a comprehensive understanding of the spatio-temporal dynamics of the environment, including the past states of observable road users and their interaction patterns, irrespective of their quantity and types. %This survey paper specifically focuses on trajectory prediction methods for vehicles, considering various constraints related to the road scene and the stochastic nature of human behavior. 
Trajectory prediction involves two main steps. First, it is essential to track and gather relevant information about neighboring road users to obtain precise and reliable trajectories. Second, based on the acquired knowledge, future trajectories of these neighboring road users need to be predicted. To accomplish these tasks, the AV's software must have access to mapping data encompassing the road scene and the surrounding area (referred to as road context). This includes information such as road and crosswalk locations, lane directions, and other relevant map-related details. Additionally, the software needs to identify and monitor Surrounding Vehicles (SVs) as well as Target Vehicles (TVs) for accurate trajectory prediction.
To tackle the inherent ambiguity of the problem, we approach vehicle trajectory prediction as a probabilistic task. We define the future trajectories of TVs as the sequence of their future states, denoted as $Y_{TVs}$:
\begin{equation} Y_{TVs} = \{e_j^t, e_j^{t+1}, e_j^{t+2}, ..., e_j^{t+f}\}_{j=1}^N \end{equation}
Here, N represents the number of TVs, f is the size of the prediction window, and $e_j^t$ denotes the state of vehicle j at time step t. The problem is formulated by computing the posterior distribution P($Y_{TVs}$|C), where $C = {X \cup I}$ represents the available information to the ego vehicle.
The historical states, captured in X, encompass the observations of N traffic participants up to time step t-1:
\begin{equation} X = \{e_j^0, e_j^{1}, e_j^{2}, ..., e_j^{t-1}\}_{j=1}^N \end{equation}
These historical states typically include attributes such as position, velocity, acceleration, orientation, etc. Additionally, I denote optional environmental information that can be considered or omitted based on availability.
In this formulation, the goal is to estimate the future trajectories $Y_{TVs}$ of the traffic participants given the available information C. The posterior distribution P($Y_{TVs}$|C) represents the probability distribution of the future trajectories conditioned on the available information.
To manage computational complexity, the prediction of each TV can be performed independently. At each stage, one vehicle is selected as the target TV, and its trajectory distribution, P($Y_{TV}$|C), is computed:
\begin{equation}
% [X_{TVs}={{fhg}}^n_{i=1}]
Y_{TVs} = \{e_j^t,e_j^{t+1},e_j^{t+2},......,e_j^{t+f}\}_{j=1}^N
\end{equation}
Here, T represents the chosen TV, and the trajectory prediction for that specific vehicle is determined.

\begin{figure}[hbt!]
    \centering
    \includegraphics[width=0.5\textwidth]{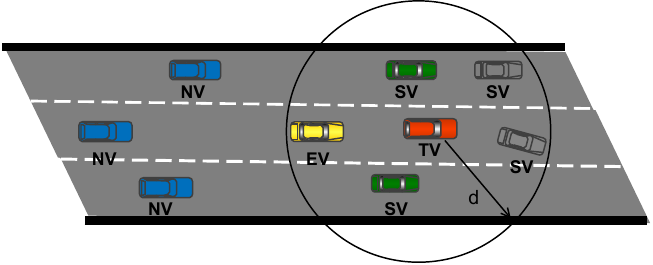}
    \caption{This illustration showcases the terminology used and the restricted view of the onboard sensors in EVs. In order to categorize vehicles as either SVs or NVs, a criterion is used where vehicles within a certain distance threshold (d) of the TV are considered to have an impact on its behavior. The two SVs that are not observable by the EV are depicted in gray. However, limited observability can lead to inaccurate predictions, such as when the preceding vehicle of the TV changes lanes, which is not visible to the EV, allowing the TV to accelerate \cite{mozaffari2020deep}. }
    \label{fig:my_label}
\end{figure}
\subsection{Terminolgy}
In the field of trajectory prediction, several terminologies are commonly used to describe different aspects of the prediction process. Here are some key terminologies.
% The relevant keywords and concepts used in this paper are as follows:
\begin{enumerate}
\item \textbf{Trajectory:} A trajectory refers to the path or motion of an object or entity over time. It represents the series of positions or states that an object traverses.
\item \textbf{Manoeuvre:} The term "maneuver" refers to the specific actions or movements performed by a vehicle or object as it navigates through its environment. Maneuvers can include various actions, such as lane changes, turns, merges, accelerations, decelerations, and stops.
    \item The vehicles whose trajectory we are interested in forecasting are called \textbf{Target Vehicles (TVs)}.
    \item \textbf{Ego Vehicle (EV)} is an autonomous vehicle that monitors its surroundings to forecast TV trajectory.
    \item The prediction model examines \textbf{Surrounding Vehicles (SVs)} since they may have an impact on how TV will behave in the future. SVs may be chosen using a variety of criteria depending on the modelling assumptions used in the study.
\item The remaining vehicles in the driving environment that are deemed to have no bearing on the behaviour of the TV are known as \textbf{Non-Effective Vehicles (NVs)}.
\item \textbf{Unimodal Trajectory} - Generate the single trajectory of single or multiple traffic participants in the given scenes.
\item \textbf{Multimodal Trajectory} - Generate the multiple trajectories of single or multiple traffic participants in the given scenes.
\end{enumerate}

The proposed terminology is illustrated in Fig. 5, through a driving scenario. The vehicles in the scenario are divided into SVs and NVs using a distance-based criterion as an example.

\subsection{Techniques applied for Trajectory-Prediction Methods} 
Trajectory prediction methods in autonomous driving can be broadly classified into the following categories:
\subsubsection{Conventional Methods}
Conventional methods for trajectory prediction refer to traditional approaches that have been commonly used to forecast the future trajectories of road users in Autonomous Driving (AD). These methods typically rely on well-established mathematical and statistical techniques to make predictions based on historical data and predefined models.
Some of the commonly used conventional methods are:
\begin{enumerate}
    \item \emph{Physics-based Models}: These methods rely on the laws of physics and kinematics principles to predict the future trajectory of a vehicle. They consider factors such as current position, velocity, acceleration, and road constraints to estimate the future path \cite{xie2017vehicle}.
\item \emph{Kinematic models}: These models assume that the motion of objects can be described by simple mathematical equations, such as constant velocity or constant acceleration models. They estimate future positions based on the object's current state and its assumed motion dynamics \cite{batz2009recognition}.
\item \emph{Kalman filters}: Kalman filters are widely used for tracking and prediction tasks. They combine measurements from sensors with predictions from a mathematical model to estimate the current state of an object and make predictions about its future trajectory \cite{lefkopoulos2020interaction}.

\item  \emph{Markov models}: Markov models capture the probabilistic dependencies between successive states of an object. They use historical data to estimate transition probabilities and make predictions based on the most likely sequence of states \cite{wang2021decision}.

\item \emph{Probabilistic Models}: Probabilistic approaches consider uncertainty in trajectory prediction by representing the future trajectories as probability distributions. These models leverage statistical techniques to estimate the most likely trajectory and provide a measure of confidence \cite{zhang2020research}.

\item \emph{Bayesian Filters}: Bayesian filters, such as Kalman filters and particle filters, are widely used for trajectory prediction. These filters combine measurements from sensors with a dynamic model to estimate the future trajectory of a vehicle. They can handle noisy sensor data and provide real-time predictions \cite{li2019dynamic}.
\textit{}
\end{enumerate}
Conventional methods for trajectory prediction are often computationally efficient and relatively easy to implement. However, they may have limitations in handling complex scenarios with intricate interactions and uncertainties. As a result, there has been a growing interest in exploring more advanced machine learning-based approaches, such as deep learning and reinforcement learning, to improve the accuracy and robustness of trajectory predictions.

\subsubsection{Deep learning-based methods}
Deep learning-based methods have gained significant attention in recent years for trajectory prediction in Autonomous Vehicles (AVs). These methods leverage the power of artificial neural networks to learn complex patterns and relationships from large amounts of data. Here are some common deep learning-based approaches for trajectory prediction:
\begin{enumerate}
    \item \emph{Recurrent Neural Networks (RNNs)}: RNNs are widely used in trajectory prediction due to their ability to model sequential data. Models such as Long Short-Term Memory (LSTM) and Gated Recurrent Unit (GRU) can capture temporal dependencies and predict future trajectories based on past observations \cite{graves2013generating}.

\item \emph{Convolutional Neural Networks (CNNs)}: CNNs are primarily used for image processing tasks, but they can also be applied to trajectory prediction by treating trajectory data as image-like representations. CNNs can extract spatial features from trajectory data and learn to predict future trajectories based on these features \cite{deo2018convolutional}.

\item \emph{Generative Adversarial Networks (GANs)}: GANs consist of a generator network and a discriminator network. They can be employed for trajectory prediction by training the generator to generate realistic future trajectories and the discriminator to differentiate between real and generated trajectories. GANs can capture the distribution of training data and generate diverse and plausible trajectory predictions \cite{zhao2019multi}.

\item \emph{Variational Autoencoders (VAEs):} VAEs are generative models that learn a latent representation of the input data. They can be used for trajectory prediction by learning the latent space representation of past trajectories and generating future trajectories conditioned on this latent representation. VAEs enable the generation of diverse and probabilistic trajectory predictions \cite{bhattacharyya2019conditional}.

\item \emph{Transformer Models:} Transformer models, originally introduced for natural language processing tasks, have also shown promise in trajectory prediction. These models can capture long-range dependencies and interactions between different agents in the scene. By attending to relevant spatial and temporal information, transformer models can generate accurate trajectory predictions \cite{quintanar2021predicting}.

\end{enumerate}

Deep learning-based methods have demonstrated improved performance in capturing complex patterns, handling diverse scenarios, and generating more accurate trajectory predictions compared to conventional approaches. However, they require large amounts of labeled training data and computational resources for training and inference. Additionally, the interpretability of the learned models can be a challenge, making it important to validate the predictions and understand the model's limitations in real-world scenarios.

\subsubsection{Reinforcement learning-based methods}
Reinforcement Learning (RL) methods have been explored for trajectory prediction in Autonomous  Driving (AD), offering a unique approach to learn optimal policies for predicting future trajectories. While RL is traditionally associated with decision-making and control, it can also be utilized in the context of trajectory prediction. Here are some RL methods used for trajectory prediction:
\begin{enumerate}
    \item \emph{Inverse Reinforcement learning (IRL):} The key idea behind IRL is to observe and analyze expert demonstrations, typically provided by human drivers, and then infer the underlying reward function that motivates their actions. This inferred reward function can be used to predict future trajectories that align with the observed expert behavior \cite{ziebart2008maximum}.
    \item \emph{Deep Inverse Reinforcement Learning (Deep IRL):} Deep IRL is an extension of Inverse Reinforcement Learning (IRL) that combines deep neural networks with the IRL framework to predict trajectories in AD. Deep IRL aims to infer the underlying reward function from expert demonstrations using deep learning techniques, allowing for more complex and high-dimensional representations of the reward function \cite{you2019advanced}.
    \item \emph{Imitation learning (IL):} IL for trajectory prediction enables autonomous systems to mimic the behavior of human drivers and generate trajectories that align with expert demonstrations. It leverages the knowledge and expertise of human drivers to make more human-like predictions and navigate the environment in a manner that is similar to how humans would drive \cite{hu2022model}. 
\end{enumerate}
By applying RL methods to trajectory prediction, models can learn from data and interactions with the environment to make accurate predictions about future trajectories. However, it is important to consider the trade-off between the complexity of RL algorithms and the availability of training data, as well as the challenges of generalization to various driving scenarios and uncertainties in the real-world environment.
\begin{table*}[]
\caption{Summary of the Conventional Trajectory prediction methods}
   \resizebox{17.5cm}{5cm}{ \centering
    \begin{tabular}{|p{3cm}|p{3cm}|p{5cm}|p{5cm}|}
    \hline
        \textbf{Based-on} & \textbf{Sub-category} & \textbf{Limitations} & \textbf{Advantageous}\\
        \hline
        Physics-Based trajectory prediction & Dynamic models \cite{rajamani2011vehicle,brannstrom2010model,lin2000vehicle,huang2006vehicle,pepy2006reducing,kaempchen2009situation} & The complexity of dynamics models can be very large and they can have many intrinsic properties. & Instead of prediction, dynamic models are employed for motion control.\\
        & Kinematics model \cite{ammoun2009real,schubert2008comparison,polychronopoulos2007sensor,lytrivis2008cooperative,barth2008will,batz2009recognition} & Only capable of predicting short-term trajectory of traffic agents. & Kinematics models are simple structure as compared to dynamic models.  \\
        & Kalman Filtering methods \cite{zhang2017method},\cite{lefkopoulos2020interaction} & It presupposes that the equations for prediction models are  linear, which is unrealistic in many real-world circumstances.& The noise of the current state of the vehicle can be handled. \\
        & Monte Carlo methods \cite{okamoto2017driver},\cite{wang2019trajectory}& Computationally inefficient when considering the large number of parameters. & This strategy is straightforward and understandable. \\
        \hline
        %293,307,308 [34] [53-55]
        Sampling Methods & \cite{houenou2013vehicle}, \cite{tran2014online},\cite{wissing2018interaction},\cite{albeaik2022limitations} & These models cannot be easily generalised to various scenes because they were only trained for certain ones. & They can withstand noise and uncertainty better.\\
        % As the algorithms' output is a dispersion of the vehicle's future states as opposed to predicting a single trajectory
        \hline
        Probabilistic Models & Gaussian Mixture Models \cite{hu2018probabilistic},\cite{augustin2019prediction},\cite{wirthmuller2020teaching},\cite{deo2018would},\cite{zhang2020research} & They might not be appropriate for modelling distributions with more complexity. & Several different probability distributions can be modelled using Gaussian Mixture Models.\\
        & Gaussian Process \cite{laugier2011probabilistic}, \cite{trautman2010unfreezing},\cite{guo2019modeling},\cite{vasquez2004motion},\cite{hermes2009long} & Before incorporating the data into the prediction models, make certain assumptions about it. & They are able to accurately assess their own uncertainty.\\
        & Hidden Markov Models  \cite{qiao2014self},\cite{deng2018improved},\cite{wang2021decision} & They do not take into account how interaction-related aspects affect the process of prediction. & HMM techniques have been quite successful at predicting driving maneuvers.\\
        & Dynamic Bayesian Network \cite{murphy2002dynamic,gindele2015learning,schreier2016integrated,bahram2015game,he2019probabilistic,li2019dynamic} & DBN struggles with error from maneuver recognition through trajectory generation. & DBN simulates the impact of both vehicle states and traffic participant interactions.\\
        \hline
    \end{tabular}}
    
    \label{tab:my_label}
\end{table*}
\begin{figure}[hbt!]
    \centering
    \includegraphics[width=0.5\textwidth]{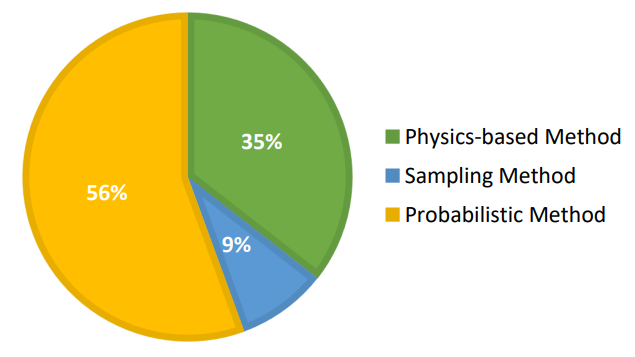}
    \caption{Participation of Research articles in trajectory prediction task using conventional approaches.}
    \label{fig:my_label}
\end{figure}

\section{Conventional trajectory prediction method}
This section classifies prediction methods into three dominant classes, Physics-based Models, Sampling Methods, and Probabilistic models, and Table II presents a brief overview of the Conventional methods for trajectory prediction with their limitations and advantages. In Fig. 6, several conventional methods and their involvement in addressing the trajectory prediction task in Autonomous Vehicles (AVs) are depicted. The analysis of the papers reveals that 56\% of the papers focus on probabilistic methods, 35\% of the papers focus on sampling methods, and the remaining 9\% of the papers are dedicated to sampling methods in this survey.

\subsection{Physics based Models}
The first class of suggested physics-based prediction models uses classical mechanics' motion equations as a foundation for modeling the target object's future motion. Either dynamic or kinematic models can be used to describe the physical behavior. A dynamics model considers the lateral and longitudinal tire forces causing the motion, but a basic dynamics model is typically chosen to balance predictive accuracy and computational effort. In contrast, kinematics models are more commonly used due to their simple form, and the Kalman Filtering (KF) techniques can handle disturbances, such as uncertainty or noise, in the vehicle's current condition. For instance, Zhang \textit{et al.} \cite{zhang2017method} proposed a vehicle-to-vehicle communication and KF-based approach to enable a host vehicle to predict the trajectories of remote vehicles and avoid obstacles. Lefkopoulos \textit{et al.} \cite{lefkopoulos2020interaction} also introduced the Interacting Multiple Model Kalman Filter (IMM-KF), a new technique that incorporates interaction-related parameters for more accurate trajectory prediction using a physics-based model over a few seconds.
The Monte Carlo approach can be used to roughly simulate the state distribution by applying a physics model to a sample of input variables at random, generating potential future trajectories. This method can be used to predict traffic participant trajectories from either a fully known or an unknown state evaluated by a filtering mechanism. Okamoto \textit{et al.} \cite{okamoto2017driver} use the Monte Carlo approach in their maneuver-based model to predict future trajectories based on the recognized maneuver. Similarly, Wang \textit{et al.} \cite{wang2019trajectory} use the Monte Carlo approach to predict trajectories and then use Model Predictive Control (MPC) to refine the reference trajectories. 
In summary, Physics-based Models are characterized by their excellent explainability, robust performance, and high accuracy, particularly for short-term prediction tasks. These models are well-suited for safety assessment purposes, typically focusing on predictions within a time horizon of no more than 1 second.
However, they may have limitations in capturing complex maneuvers, often relying on simplified assumptions and limited adaptability in unknown  or dynamic environments.

\subsection{Sampling Methods} These techniques involve sampling the future states of traffic participants. Instead of predicting a single trajectory, these approaches generate a distribution of possible vehicle states, which makes them more robust to noise and uncertainty. There are two main types of sampling: generating multiple trajectory segments or particle states. In their study, Houenou \textit{et al.} \cite{houenou2013vehicle} combined a maneuver-based approach with a model-based approach assuming Constant Yaw Rate and Acceleration (CYRA) to develop a trajectory prediction method. They identified the maneuver and selected the best trajectory from a set generated by minimizing a cost function.
Meanwhile, Tran and Firl \cite{tran2014online} utilized Monte Carlo Simulation (MCS) to predict multimodal trajectories and a normalized three-dimensional Gaussian Process (GP) regression model to learn vehicle behavior at a junction. Similarly, Wissing \textit{et al.} \cite{wissing2018interaction}  proposed an interaction-aware trajectory prediction method that used MCS to simulate interactions and forecast the distribution of potential future positions for the target vehicle. This approach leveraged the Intelligent Driver Model (IDM) \cite{albeaik2022limitations} to account for the interacting behaviors of traffic participants, and the particles were disseminated using a lane-change driving model that considered three different lateral moves and aspects of the driving scenario with each run of the MCS. To summarize, sampling methods are essential tools for trajectory prediction, and the choice of method depends on the specific problem and the properties of the distribution of trajectories. However, these methods face challenges such as computational complexity, the requirement for efficient sampling strategies, and the possibility of overlooking important trajectory regions.

\subsection{Probabilistic Models} 
%Probabilistic frameworks are utilized for trajectory prediction, enabling the creation of probability distributions for the future positions of traffic participants. Similar to sampling models, these frameworks account for uncertainties and noise in both the data and prediction models. They are particularly valuable when simulating and modeling these uncertainties. Additionally, probabilistic techniques are effective for modeling conditional relationships, such as the correlation between the future state of the target vehicle and its intended maneuver. By incorporating probabilistic frameworks, trajectory prediction can provide more comprehensive and accurate insights into the anticipated behaviors of traffic participants.

A probabilistic framework in trajectory prediction refers to the use of probability theory to model and estimate the likelihood of future trajectories of objects or entities, such as vehicles, pedestrians, or other moving objects. It involves representing uncertainty and variability in the prediction process and providing probabilistic distributions or confidence measures for the predicted trajectories.
In a probabilistic framework, trajectory prediction is typically formulated as a conditional probability problem, where the goal is to estimate the probability distribution of future trajectories given the observed past trajectories, sensor measurements, and other relevant information. This involves incorporating probabilistic models, statistical techniques, and machine learning algorithms to capture the uncertainties and dependencies in the data.
%Trajectory prediction is also done using probabilistic frameworks. It has the ability to create a probability distribution across the upcoming positions of traffic participants, much like sampling models. When prediction frameworks are created to simulate the noise and other uncertainties in the data and prediction models, they are very useful. These techniques are also effective resources for conditional relationship modeling. For instance, the relationship between the target vehicle's future state and the intentions of the maneuver.

\subsubsection{Gaussian Mixture Model}
A Gaussian Mixture Model (GMM) is a probabilistic model that is often used in trajectory prediction to capture the uncertainty and complexity of the data. It represents the distribution of the trajectories as a combination of multiple Gaussian distributions, each representing a possible mode or cluster of trajectories.
A Semantic-based Intention and Motion Prediction (SIMP) was proposed by Hu \textit{et al.} \cite{hu2018probabilistic}. It uses multiple 2D GMM to model the probability distribution of movement patterns in driving scenarios and Deep Neural Networks (DNNs) to calculate the likelihood of entering the intersection area. GMM was also employed in other methods to model specific motion patterns \cite{augustin2019prediction},\cite{wirthmuller2020teaching}. Although classic Hidden Markov Model (HMM) approaches have been quite successful at predicting drivers' moves, they do not take the impact of interaction-related aspects into account during the prediction process, therefore the results of their predictions are insufficiently accurate in real-world traffic situations. An interaction-related vehicle trajectory prediction model based on HMM and Variational GMM is proposed by Deo \textit{et al.} \cite{deo2018would}.
The knowledge about vehicle interactions is discovered by locating the energy function's ideal solution. A GMM-HMM manoeuvre prediction model that takes interaction-aware elements into account is proposed by Zhang \textit{et al.} \cite{zhang2020research} based on game theory. Jiang \textit{et al.} \cite{jiang2022vehicle} developed a GMM-HMM recognizer based on joint mutual information maximization to estimate the driver's lane-changing intention. This recognizer was incorporated as a node in the Dynamic Bayesian Network (DBN) framework. To summarize, GMMs provide a versatile and robust method for trajectory prediction by capturing complex patterns and variations in the data. They are capable of handling multimodal distributions, which allows for representing different maneuver types or behavior patterns exhibited by vehicles. However, it's important to note that training and inference with GMMs can be computationally demanding. Additionally, determining the optimal number of Gaussian components or modes in the model can be a challenging task.
%In summary, GMMs offer a flexible and powerful approach for trajectory prediction, as they can capture complex patterns and variations in the data. They can handle multimodal distributions, allowing for the representation of different maneuver types or behavior patterns exhibited by the vehicles. However, it's important to note that training and inference with GMMs can be computationally intensive, and determining the appropriate number of Gaussian components or modes in the model can be a challenging task.

\subsubsection{Gaussian Process}
When utilizing Gaussian Process (GP) for trajectory forecasting, trajectories are considered as samples taken from a GP along the time axis. These samples are represented by N discrete points, which are mapped to an N-dimensional space. In this N-dimensional space, the samples adhere to a Gaussian distribution. During the modeling step, the GP model's main objective is to estimate the GP parameters based on these samples. By fitting the GP to the observed trajectory samples, the model captures the underlying patterns and dynamics of the data. The GP parameters, such as the mean and covariance, define the characteristics of the GP and determine the shape and uncertainty of the predicted trajectories \cite{hewing2020simulation}. In the study by Laugier \textit{et al.} \cite{laugier2011probabilistic}, GPs were employed to predict trajectories following the evaluation of likely behaviors using Hidden Markov Models (HMMs). Trautman \textit{et al.} \cite{trautman2010unfreezing} addressed the frozen robot problem by utilizing GP for joint collision avoidance. Additionally, GP can be utilized to simulate interaction-related aspects in trajectory prediction tasks.
%In \cite{laugier2011probabilistic}, GP is utilised to forecast the trajectories after HMM is used to evaluate likely behaviours. The frozen robot problem is solved by Trautman  \textit{et al.} \cite{trautman2010unfreezing} using GP for joint collision avoidance and GP can also be utilised to simulate interaction-related aspects.
%Guo \textit{et al.} \cite{guo2019modeling} use the GP and Dirichlet process (DP) to construct motion processes and a non-parametric Bayesian network to extract possible motion patterns. The prototype set can be trained to represent each trajectory through approaches based on prototype trajectories. Thus, the key distinction between these approaches is how to create the prototype trajectory. 
Guo \textit{et al.}  \cite{guo2019modeling} employed GPs and the Dirichlet process (DP) to construct motion processes and utilized a non-parametric Bayesian network to extract potential motion patterns. The prototype set was trained to represent each trajectory using methods based on prototype trajectories. The primary differentiation among these approaches lies in the technique employed to generate the prototype trajectory. Govea \textit{et al.} \cite{vasquez2004motion}'s statistical analysis of the mean and variance of each sample of a trajectory yields the prototype trajectories. In their study, Hermes \textit{et al.} \cite{hermes2009long} focused on capturing variations in vehicle movement through training. They divided the sample trajectories into multiple subsets and generated several prototype trajectories as an outcome of their research. In summary, GP is a valuable tool in trajectory prediction for AVs, providing several advantages such as flexibility, probabilistic forecasts, adaptability, and potential integration with other techniques. However, one limitation of approaches based on trajectory samples is their limited applicability to new contexts, which hinders their adaptability to diverse scenarios and environments.

\subsubsection{Hidden Markov Model}
In trajectory prediction using the Hidden Markov Model (HMM), the observation sequences are comprised of the previous states of the traffic participants. The HMM algorithm is applied to estimate the most likely future observation sequence based on these past observations. Qiao \textit{et al.}  \cite{qiao2014self} offers a technique called HMTP* based on HMM that adaptively selects parameters to replicate real situations at a pace that changes over time. In \cite{deng2018improved}, HMM and fuzzy logic are utilised to anticipate driver maneuvers. HMM can also be included in planning and decision-making processes. HMM is employed in \cite{wang2021decision} for risk assessment and trajectory prediction, with the outcomes being supplied into the system for making decisions. A behavior prediction method based on the HMM is proposed by Li \textit{et al.} \cite{li2022autonomous}, considering the direction of incoming cars. To ensure the reliability of the prediction results, multiple sets of initial values are generated. Additionally, this approach aims to reduce the model's dependency on data, resulting in improved prediction performance. Ren \textit{et al.} \cite{ren2022method} propose the lane-changing behavior recognition model based on the Continuous Hidden Markov Model (CHMM) is developed to identify the lane-changing behavior of nearby vehicles. In summary, The HMM is highly beneficial for trajectory prediction due to its ability to capture temporal dependencies, handle missing or noisy data, and account for the uncertainty involved in predicting future trajectories. However, an assumption of HMM is that the hidden states are Markovian, implying that the probability of transitioning to a future state depends solely on the current state.

%The HMM is particularly useful in trajectory prediction as it can capture temporal dependencies, handle missing or noisy data, and model the uncertainty associated with predicting future trajectories. However, it assumes that the hidden states are Markovian, meaning that the probability of transitioning to a future state only depends on the current state.
% To forecast the driver's moves, Holger et al. [52] fed the steering angle and global coordinates into an HMM
% HMM

\subsubsection{Dynamic Bayesian Network}
%Using the Bayesian Network and taking time sequence into account, Dynamic Bayesian Network (DBN) is a maneuver-based approach. DBN and Bayesian networks share the same fundamental ideas and methods for making probabilistic inferences. The distinction is that although Kevin \textit{et al.} \cite{murphy2002dynamic} establish the idea of time templates to address timing concerns in probabilistic models, Bayesian Networks represent static systems. A time segment is a time template that has been materialised in accordance with DBN, discretizing continuous time into countable points with predetermined time granularity.
By incorporating time sequence and leveraging the Bayesian Network framework, the Dynamic Bayesian Network (DBN) offers a maneuver-based approach for trajectory prediction. DBN and Bayesian networks share fundamental concepts and methodologies for conducting probabilistic inferences. However, one distinction is that Kevin \textit{et al.} \cite{murphy2002dynamic} introduced the concept of time templates to address timing considerations in probabilistic models, while Bayesian Networks typically represent static systems. In the context of DBN, a time segment refers to a time template that discretizes continuous time into discrete points with a predetermined time granularity.
Multiple vehicles driving manoeuvres are modelled by Gindele \textit{et al.} \cite{gindele2015learning}. All vehicle states,  interaction relationships, observation statuses, road structures, etc. are included in the input data. DBN is used by Schreier\textit{et al.} \cite{schreier2016integrated} to evaluate driving manoeuvres, and they use the kinematics model associated with each maneuver to forecast the trajectory. Game theory is used in \cite{bahram2015game} to anticipate the vehicle movement, and DBN, which takes interaction-related elements into account, then judges the vehicle motion. He \textit{et al.} \cite{he2019probabilistic} employ DBN to recognise lane-change and vehicle-following motions and to forecast the trajectory. In \cite{li2019dynamic}, DBN is designed to consider the kinematic factors of vehicles, Vehicle manoeuvres, and their interdependence, and road-related information.
To summarize, when utilized for trajectory prediction, DBN takes into account the interactions between traffic participants, leading to improved performance in conventional machine learning-based methods. However, DBN still encounters challenges in accurately recognizing maneuvers and generating trajectories. Many existing methods are limited to distinguishing only two or three maneuvers, such as lane-keeping and lane-changing, which restricts the model's ability to generalize to a wide range of scenarios. Consequently, there is a need for further advancements in DBN-based approaches to enhance their maneuver recognition capabilities and improve the model's generalization ability in trajectory prediction tasks.

%When used for trajectory prediction, DBN takes into account the impact of interactions between traffic participants and has shown good performance compared to traditional machine learning-based methods. However, some methods are limited in their ability to classify maneuvers, often only being able to distinguish between two or three, such as lane-keeping and lane-changing, which reduces the model's ability to generalize to new situations.

\section{Deep learning-based prediction methods}
Conventional prediction techniques are only effective in basic prediction scenarios and short-term prediction assignments. Deep learning-based trajectory prediction models have gained popularity due to their ability to consider various factors that contribute to accurate predictions. These models take into account physical factors, such as the position, velocity, acceleration, size, and shape of vehicles. They also consider road-related factors like traffic signs, traffic lights, road geometry, and road obstacles. Additionally, interaction-related factors, including the distance between vehicles, relative speeds, and the presence of communication systems, are considered. Fig. 7 provides a general overview of these methods. The following sections outline the most prevalent deep learning-based methods used for trajectory prediction in Autonomous Vehicles (AVs).

%In the subsequent sections, this paper presents a summary of the prevailing deep learning-based methods used for trajectory prediction in Autonomous Vehicles (AVs).

%With the ability to consider a multitude of factors such as physical factor (ex. position, velocity, acceleration, size, and shape of the vehicles), road-related (ex. traffic signs, traffic lights, road geometry, and road obstacles), and interaction-related factors (ex. distance between vehicles, relative speeds, and the existence of communication), deep learning-based trajectory prediction models have become increasingly popular. As depicted in Fig. 7, these models provide a comprehensive overview of the various approaches. This paper outlines the most prevalent deep learning-based trajectory prediction techniques for autonomous vehicles.

\begin{table*}
\caption{Summary of Recurrent Neural Network based methods: Related Work \& Year, No. of Predicted Trajectories for each vehicle, Prediction Horizon (PH), Advantages and Limitation of each work, Summary of each work and Evaluation Metric (EM).}  

 \resizebox{17.5cm}{7cm}{
  \centering
\begin{tabular}{|p{0.7cm}|p{1cm}|p{0.5cm}|p{4cm}|p{4cm}|p{4cm}|p{1.5cm}|}
\hline
        \textbf{Ref.}&\textbf{\# Trajectories}&\textbf{PH}&\textbf{Advantages}& \textbf{Limitation} & \textbf{Summary of Prediction Method} & \textbf{EM}\\
        \hline 
\cite{zyner2018recurrent} \newline 2018  &     1   & 1.3s &  To more clearly see the areas of conflict and inaccuracies, the data is shown as an overlay on the map.                                                                                              &   Interaction between vehicles does not take into account.                                                                                                         & Single RNN: The fully connected layer of the model and the three LSTM layers are utilised for the output maneuvers. &    Accuracy   \\
\cite{ding2019online}  \newline 2019     &   1     & 4s   & Makes use of a policy anticipation network to make high-level policy decisions.               &  Interaction modelling for prediction is not investigated.                                                   & Single RNN: RNN to encode the vehicle's historical data.                                                                                                           & RMSE                 \\

\cite{chang2019argoverse} \newline 2019     & 1,3,6,9 & 3s   & Investigation into the use of rich maps for 3D object tracking and motion prediction.          &  Interaction between vehicles and road structure is absent.                                                   & Single RNN: LSTM encoder-decoder that incorporates social and geographic information.                                                                   & minADE,\newline minFDE, DAC, MR \\
\cite{zyner2019naturalistic} \newline 2020     & 3       & 5s   &  Clustering technique that captures the range of potential routes from the prediction output.   &  Modeling for only roundabout scenario.                                                                       & Single RNN: Encoder-Decoder three-layer LSTM model incorporates  all observations for a single track.                                                    & Euclidean Distance, MHD   \\
\cite{xing2019personalized}   \newline 2020   & 1       & 5s   &  Inter-vehicle communication signals were used.                                                 & Prediction for the leading vehicle only.                                                                    & Single RNN: LSTM layer to extract the time-sequence patterns for vehicle trajectory prediction.                                                         & RMSE,MHD             \\

\cite{xin2018intention} 2018 & 1       & 5s   & 1. Not impacted by internal dataset imbalance; 2. Lateral position's small prediction boundaries. &  Model is not generalise for crossings and unstructured roads.                                            & Multiple RNN: One LSTM to forecast the target lane and another LSTM to predict the trajectory based on the estimated target lane of the target vehicle. & RMSE                 \\
\cite{deo2018multi} 2018 & 1,6     & 5s   &  Trajectories based on maneuver classes.                                                        &  Assumption regarding six surrounding vehicles.                                                               & Multiple RNN:  One LSTM encoder for the input sequence, six LSTM decoders to a different manoeuvre, and One LSTM decoder, to forecast  trajectory.      & RMSE                 \\

 \cite{dai2019modeling}  2019   & 1       & 6s   & Add shortened connections between the two  LSTM layers.                                        & Predicted trajectories don't show lane change procedure characteristics.                                     & Multiple RNN: one group of LSTM is used to simulate the motion of nearby vehicles, and the other is used to simulate how nearby vehicles interact.      & RMSE                 \\
\cite{ding2019predicting} 2019 & 1       & 4s   &  Modeling the pair-wise interaction using GRU.                                                  &  Experiments restricted to a dataset of highways.                                                             & Multiple RNN: The trajectory encoding for each individual vehicle is obtained using an RNN encoder network.                                           & NLL, accuracy        \\
 \cite{tang2019multiple} 2019 & 1,12    & 5s   &  All agents in the scenario do interactive, parallel step-wise rollouts.                       &  The combination of discrete and continuous latent variables in the model is not investigated.               & Multiple RNN: RNNs running in parallel to depict the agents in a scene.                                                                                              & NLL,\newline minADE,\newline minFDE    \\
\cite{zhang2021vehicle} 2021 & 1       & 5s   &  Statistically examining numerous earlier traffic flow paths.                                  & 1. vehicle turning signals is not included; 2. This is not encounters in the conflict zone from approaching vehicles. & Multiple RNN: Two LSTM blocks: One for intention prediction, another for trajectory prediction.                                                                      & MHD                  \\
      
         \hline
\end{tabular}}
\end{table*}

\begin{figure}[]
    \centering
     \includegraphics[width=0.5\textwidth, height =0.15\textwidth]{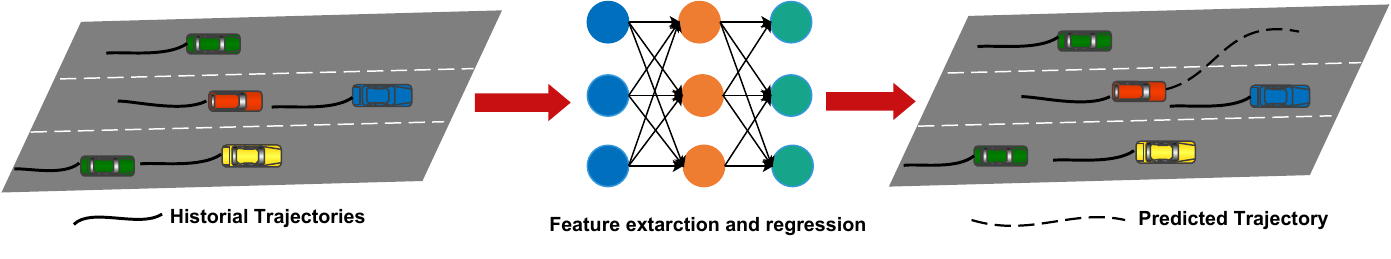}
    
    \caption{The illustration of Deep learning-based methods.}
    \label{fig:my_label}
\end{figure}
\begin{table*}[]
\caption{Summary of Temporal Convolution Neural Network based methods: Related Work \& Year, No. of Predicted Trajectories for each vehicle, Prediction Horizon (PH), Advantages and Limitation of each work, Summary of each work and Evaluation metric (EM) - This information is not available for work}
%/ predicting any number of trajectories for the work
 \resizebox{17.5cm}{4cm}{
    \centering
    \begin{tabular} {|p{0.5cm}|p{1cm}|p{0.5cm}|p{5cm}|p{3.5cm}|p{4cm}|p{1.5cm}|}
    \hline
        \textbf{Ref.}&\textbf{\# Trajec\newline tories}&\textbf{PH}&\textbf{Advantages}& \textbf{Limitation} & \textbf{Summary of Prediction Method} & \textbf{EM}\\
        \hline 
        \cite{zhang2020lane}\newline2020&       1    & - & 1. Estimating the vehicle's position and steering-wheel angle; 2. Consider the steering wheel angle to be a sign of lane-changing behaviour. & Merging and weaving zones are not evaluated.                                              & TCN extracts features from the past and current states of actor of interest. & MAE,MSE   \\
        \cite{strohbeck2020multiple}\newline2020 &1,3,6 & 3s   &  Employ rasterized graphics to depict the location of the actor of interest and its changing surroundings.                                  & The effects of surrounding vehicles on the ego vehicle is not clear.                      & TCN to predict the long-term lane-changing trajectory and driving behavior   & minADE,\newline minFDE,\newline MR, DAC \\ 
           
         \cite{katariya2022deeptrack}\newline2022&  1     & 5s   & 1. Utilize depthwise convolution;  2. Lowering the size and operational complexity of models.                                                 & Due to ambiguity in driver behaviour, the model failed to forecast in several situations. & TCN utilized as an encoder to encode the vehicle dynamics.                    & RMSE,ADE\newline FDE         \\
         \cite{li2022vehicle}\newline2022&  1     & 5s   &This takes into account the effects of the vehicle's driving behavior.                                                                          & The model is only appropriate for highway scenes.                                          & Many TCN blocks make up a TCN, which encodes the input as a context vector.  & RMSE                 \\
         \cite{azadani2022novel}\newline2022& 1,3  & 4.8s    & 1. Included intersections without signals; 2. The model doesn't depend on a specific action.                                                 & Interactions between the surrounding vehicles and other road users are absent.            & Working with a mixed-density layer and dilated convolutional networks.       & RMSE                 \\ \hline
    \end{tabular}}

    \label{tab:my_label}
\end{table*}
\label{stofartssdvns}

\begin{table*}[]
    \caption{Summary of Attention based methods: Related Work \& Year, No. of Predicted Trajectories for each vehicle, Prediction Horizon (PH), Advantages and
Limitation of each work, Summary of each work, and Evaluation Metric (EM) }
%- This information is not available for work/ predicting any number of trajectories for the work
    
    \resizebox{17.5cm}{10cm}{
    \centering
    \begin{tabular}
     {|p{0.7cm}|p{1cm}|p{0.5cm}|p{4cm}|p{4cm}|p{4cm}|p{2cm}|}
    \hline
         \textbf{Ref.}&\textbf{\# Trajectories}&\textbf{PH}&\textbf{Advantages}& \textbf{Limitation} & \textbf{Summary of Prediction Method} & \textbf{EM}\\
        \hline
        \cite{hao2020attention}\newline 2020 & 1 & 4s & When the predicted time length increases, the attention mechanism keeps historical data from being lost.& Forecasts future position by using only the vehicle's prior position as input.&Attention mechanism combined with GRU decoder to forecast vehicle position.& RMSE\\
       \cite{yan2020trajectory}\newline2020 & 1& 5s &A lane attention system combines real-time lane information.&1. The model is tested for highway scenarios; 2. The road information excluded.  &Context attention and lane attention are the two spatial-attention strategies used to explain how vehicles interact.&RMSE\\
       \cite{kim2021lapred}\newline 2021 & 1,5,6,12 & 6s& Extracting the joint properties related to the lane and the trajectories of the nearby agents.&Speed and yaw angle are not included with the positional coordinate of vehicles.&  The self-attention technique used to concentrate on features from the target vehicle's preferred lane.&ADE,FDE\\
       
      \cite{fu2021trajectory}\newline2021 & 1& 3s&1. Construct a spatial and temporal navigation map; 2. Predict current location and velocity of surrounding vehicles.&1. Unstructured and urban roads are not addressed; 2. Select the most influential six vehicles around the
ego-vehicle. &  Attention mechanism occurs between the encoder and decoder components.&RMSE \\
\cite{yu2021dynamic}\newline2021&1&5s& Constraint net used to extract and model the external environmental constraints.&It is challenging to generalise the model in complex conditions.&The most remarkable vehicles are chosen at each time step using an attentional decoder with LSTM.& RMSE\\
% \cite{} &1&4.8s& 1. modelling spatial and temporal interactions simultaneously. 2. Analysis in Highways and Roundabout  Scenarios& The map information does not include & The two different attention mechanisms are in the encoder to extract better features,& ADE,FDE\\
\cite{wu2021hsta}\newline2021&1&5s & 1. Concurrently model spatiotemporal interactions; 2. Trained the model in an end-to-end fashion & 1. Referring to the entire scene's agents as the neighbourhood; 2. Certain cases at the roundabout fail. 
& Multi-head attention is utilised to represent temporal correlations of interactions, and a State Gated Fusion (SGF) layer is applied to combine spatial and temporal interactions.& ADE,FDE
 \\
\cite{meng2021intelligent}\newline2021& 1 &5s & Reflecting the spatial relationship between the surrounding vehicles and the target vehicle. & Predicting the longitudinal trajectory is challenging.
& Describe the spatial and temporal attention modules separately. & RMSE,NLL
\\
     \cite{kim2020multi}\newline2020 &1&5s& 1. Interactions can unsupervisedly learn to focus on a few key vehicles; 2. Model is scalable with any number of nearby vehicles & The model is for highways only. &Both a vehicle attention layer and a lane attention layer are present in the encoder.The decoder has a vehicle attention layer only.&RMSE \\
      
       %To selectively draw attention to particular context vector properties before the decoder accepts it as input, 
       
       % Not take into account the effects of infrastructure elements, such as road structure and traffic signs
      %  & & & ,[],[] & It is capable of learning the most crucial temporal and spatial elements for prediction and predicting the trajectory of surrounding vehicles\\
       % & & & [116] & Multi-head attention to extracting lane and vehicle attention to determine future trajectory distributions\\
   
     \cite{messaoud2020attention}\newline2020& 3& 5s & 1. Non-local social pooling modeling the interaction with a multi-head attention mechanism; 2. Incorporate the data about the vehicle class &1. There are no scenarios for mixed or heterogeneous traffic; 2.Road structure is not included. & the attention layer in between the encoder-decoder based on LSTM ,models the interactions between the target and the neighboring vehicles&RMSE,Min and Max RMSE\\
     \cite{messaoud2021trajectory}\newline2021 &1,15 &6s&1. Combined representation of the agents and the static scene; 2. Attention maps are shown visually. &Specify the interaction space explicitly using a defined distance. &Each attention head simulates an interaction between the combined context features and the target.& minADE,\newline minFDE, MR,off-road rate\\
       \cite{lin2021vehicle}\newline2021&1&5s&The inception-based module was used to maintain the spatial information of the surrounding vehicles.&1. Fewer nearby vehicles taken into account; 2. Trajectory prediction depend on maneuver.& LSTM-based attention is used to extract the target vehicle's driving intentions and temporal driving behaviours.&RMSE\\ 
        \cite{yang2022lane}\newline2022 &1&5s& 1. Encourages a deeper understanding of the lane-change process in vehicles; 2.Target vehicle's time series attention module and the nearby vehicles' spatial attention module.& When changing lanes, neither the pose nor the change in speed of a target vehicle is taken into account.& LSTM networks with a spatiotemporal attention mechanism for extracting trajectory features. & RMSE\\
       \cite{kim2022diverse}\newline2022&1&5s&Modeling the link between the vehicles using grid-based discretization. & Road structure and their interaction is not included. &LSTMs and spatial-temporal attention mechanisms are combined for explainability.&RMSE\\
    \cite{hu2022trajectory}\newline2022&1,6&3s &1. Considers all interactions, including agents-agents, lanes-agents, lanes-lanes, and agents-agents; 2. A novel quantitative evaluation metric proposed. & Performance margin is insignificant with the other models. & An attention module combines the characteristics of many proposals that reflect various actions or behaviours.& minADE,\newline minFDE,\newline minLaneFDE\\
   
   \cite{hasan2023mals}\newline2023 &1&5s& Encoding social and temporal interaction using multi-head attention. &1. Model is not adopted for urban driving; 2. This is not using spatial HD maps, traffic signals, or lane kinds & Social and temporal attention layer in both encoder-decoder. &RMSE \\\hline
    \end{tabular}}
   
    \label{tab:my_label}
\end{table*}
\label{stofartssdvns}
\subsection{Sequential Modelling}
Deep learning-based trajectory prediction methods often involve using a sequential network to extract features from historical trajectories and can serve as the output layer. These networks typically include Recurrent Neural Networks (RNNs), Temporal Convolutional Neural Networks (TCNs), Attention Mechanisms (AMs), and Transformers. Fig. 8 provides a visual representation, in percentages, of the distribution of research papers utilizing different algorithms in sequential modeling for trajectory prediction. It can be observed that TCNs are less commonly used in the AVs trajectory prediction task compared to other algorithms such as RNNs, AMs, and Transformers.

%In Fig. 8, the number of papers, show in percentages, in sequential modeling for trajectory prediction is depicted, showcasing the utilization of various algorithms. 

%representation of sequential modeling and their algorithms to show their participation in solving the trajectory prediction task in AVs.
\begin{figure}[hbt!]
    \centering
     \includegraphics[width=0.5\textwidth]{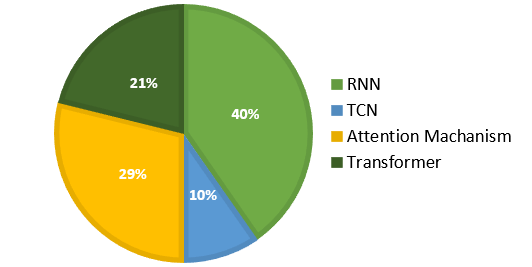}
    \caption{Participation of Research articles in trajectory prediction task using Sequential learning-based approaches.}
    \label{fig:my_label}
\end{figure}
\subsubsection{Recurrent Neural Network}
The Recurrent Neural Network (RNN) was designed to handle temporal information, as opposed to conventional machine learning methods and Convolution Neural Networks (CNNs), which excel at processing spatial information \cite{graves2013generating}, \cite{sutskever2014sequence}. It maintains a record of past-time step data and combines input and hidden states to generate the desired output. However, when dealing with a large number of time steps, the RNN's gradient can either weaken or explode, causing issues. To tackle this problem, gated RNNs like the Long Short-Term Memory Network (LSTM) and Gated Recurrent Unit (GRU) have been developed. RNN-based trajectory prediction models are categorized as either single RNN models or multiple RNN models. 
%Fig. 8 presents the algorithms for sequential modeling and their role in addressing the trajectory prediction task in autonomous vehicles (AVs).

1) \textit{ Single RNN:} To predict trajectories based on maneuvers or single-modal trajectory prediction, a single RNN model is employed. Additionally, it can be incorporated into auxiliary models to facilitate more complex capabilities, such as interaction-aware forecasting. In various studies \cite{zyner2018recurrent, zyner2017long, phillips2017generalizable}, LSTM has been utilized as a sequence classifier for vehicle maneuver prediction. In these studies, LSTM cells extract vehicle attributes, and the output layer predicts movements using the final cell's hidden states. Fully connected layers are used to extract features and input them into three-layer LSTMs in \cite{zyner2018recurrent, zyner2017long}, while two LSTM layers without embedding are employed in \cite{phillips2017generalizable}. Altché \textit{et al.}  \cite{altche2017lstm} used a single-layer LSTM to estimate the target vehicle's trajectory. Ding \textit{et al.} \cite{ding2019online} used an LSTM encoder to predict the target vehicle's maneuver by encoding its states. 
% Trajectory prediction was performed using the expected maneuver and map data, and nonlinear optimization methods were used to maximize the initial future trajectory based on interaction-related variables, traffic regulations, and map data. 
The LSTM encoder-decoder incorporating social and geographic information was compared to the Nearest Neighbor (NN) regression method in \cite{chang2019argoverse}. In \cite{zyner2019naturalistic}, Zyner \textit{et al.} utilized a weighted Gaussian Mixture Model (GMM) to forecast multi-modal trajectories. The GMM's parameters were obtained using an encoder-decoder three-layer LSTM model, and the predicted trajectories were clustered using the modal with the highest probability. Xing \textit{et al.} \cite{xing2019personalized} used GMM to identify driving styles, LSTM and fully connected regression layers to assess sequence data and driving styles to predict vehicle trajectory, with the first vehicle in the fleet following its predicted trajectory. Kawasaki \textit{et al.} \cite{kawasaki2020multimodal} integrated LSTM and KF for multi-modal trajectory prediction while considering lane information.

2) \textit{ Multiple RNNs}: The development of neural networks has resulted in the widespread usage of various types of RNN architectures. In Xin \textit{et al.} \cite{xin2018intention}, two separate LSTMs are employed to predict the target lane and trajectory of a vehicle based on its current state and expected lane. Deo \textit{et al.} \cite{deo2018multi} suggest six LSTM decoders, each connected to a different maneuver, to forecast multi-modal trajectories. Dai \textit{et al.} \cite{dai2019modeling} utilizes two groups of LSTM networks to simulate the motion and interaction of nearby vehicles. Ding \textit{et al.} \cite{ding2019predicting} present a group of GRU encoders to characterize paired interactions between vehicles. Min \textit{et al.} \cite{min2019rnn} use multiple RNNs and fully connected layers to generate the cubic polynomial coefficients that describe the target vehicle's future trajectory. Tang \textit{et al.} \cite{tang2019multiple} employ an attention mechanism to create a dynamic state encoder consisting of multiple RNNs sharing parameters to predict the multi-modal trajectory. Multi-modal trajectories are generated using an LSTM encoder-decoder and a multi-head attention layer in \cite{mercat2020multi}. A paradigm with several LSTMs is proposed by Zhang \textit{et al.} in \cite{zhang2021vehicle} for both trajectory and intention prediction. Xu  \textit{et al.} \cite{xie2021congestion} introduce a student-teacher network for trajectory prediction, where the student algorithm is based on an LSTM Encoder-Decoder model, and the instructor algorithm is based on a Convolutional Graph Network. Although RNNs are widely used for analyzing and predicting data series, such as trajectory prediction, they have limitations in simulating spatial relationships, such as vehicle interaction, and processing image-like data, such as the context of a driving scene. This is why complex RNN-based solutions often require multiple techniques to overcome the limitations of a single RNN. In summary, RNNs offer a powerful approach for trajectory prediction by effectively modeling temporal dependencies. They can handle variable-length sequences and provide interpretability. However, they can suffer from vanishing or exploding gradients and sequential computation limitations. Understanding these factors is crucial when applying RNNs to trajectory prediction tasks.
Table III summarizes the RNN-based approaches for trajectory prediction, providing information on the Prediction Horizon (PH) in seconds (s) and the number of predicted trajectories. The table also includes the Evaluation Metrics (EM) used for training and testing, along with highlighting the strengths and weaknesses of each study.

%Table 3 provides a summary of RNN-based approaches for trajectory prediction, along with the Evaluation Metrics (EM) used for training and testing. The table provides information on the Prediction Horizon (PH) in seconds (s) and the number of trajectories that are predicted. It also highlights the strengths and weaknesses of each study.

%\begin{figure}
    %\centering
    % \includegraphics[width=0.5\textwidth]{dl1.png}
   % \caption{The illustration of Deep learning-based methods  }
    %\label{fig:my_label}
%\end{figure}

\subsubsection{Temporal convolutional networks} 

Temporal Convolutional Networks (TCNs) are a popular type of deep neural network architecture used in trajectory prediction tasks. In trajectory prediction tasks, TCNs are trained on historical trajectory data and are used to predict the future trajectory of a vehicle or pedestrian.  Compared to recurrent networks, TCNs have been shown to outperform them in tasks such as handwritten recognition \cite{sharma2021towards}, audio synthesis \cite{pandey2019tcnn}, and time-series data \cite{zheng2022denoising}. One advantage of TCNs is their ability to handle variable-length sequences without information leakage. Bai \textit{et al.} \cite{bai2018empirical} employed causal convolution, dilated convolution, residual connection, and a completely connected network to create TCN. Zhang \textit{et al.} \cite{zhang2020lane} utilized TCN to predict lane-change maneuvers and trajectories. In \cite{strohbeck2020multiple}, CNN processes the rasterized image while TCN collects features from historical trajectory data that are combined with the raster feature and the present state. DeepTrack \cite{katariya2022deeptrack} is a lightweight deep learning algorithm with accuracy comparable to top trajectory prediction algorithms, but with a much smaller model size and reduced computational complexity. DeepTrack encodes the vehicle dynamic using TCN and reduces model complexity by using depthwise convolution as the fundamental building block. In \cite{li2022vehicle}, a TCN encoder and a Multi-Layer Perceptron (MLP) decoder are used, where the position and speed of the vehicles are sequentially entered and encoded as a context vector during the encoding procedure. To improve prediction accuracy, an intention recognition module is included with a TCN encoder. In \cite{azadani2022novel}, Mozhgan \textit{et al.} integrate a dilated convolutional network-based encoder-decoder with a mixture density network to predict potential multimodal pathways taken by target vehicles. It is evident that TCN possesses benefits when it comes to handling time-series data. In summary, TCNs offer a powerful approach for modeling temporal dependencies in trajectory prediction tasks. They excel at capturing short-term and long-term dynamics, perform efficient parallel computation, and have interpretable receptive fields. However, spatial relationships and long-term memory might require additional considerations. 
Table IV presents a summary of TCN-based approaches for trajectory prediction, including the prediction horizon (in second (s)), the number of trajectories predicted, and the evaluation metrics used for training and testing. The table also highlights the strengths and weaknesses of each study.
%Table 4 provides a summary of TCN-based approaches for trajectory prediction, along with the evaluation metrics used for training and testing. The table also highlights the strengths and weaknesses of each study.
\begin{table*}[]
 \caption{Summary of Transformer based methods: Related Work \& Year, No. of Predicted Trajectories for each vehicle, Prediction Horizon, Advantages and Limitation of each work,
Summary of each work, and Evaluation Metric (EM) \newline - Predicting any number of trajectories for the work} \resizebox{17.5cm}{9cm}{
    \centering
    \begin{tabular} {|p{0.7cm}|p{1cm}|p{0.5cm}|p{4cm}|p{4cm}|p{4cm}|p{1.5cm}|}
    \hline
         \textbf{Ref.}&\textbf{\# trajectories}&\textbf{PH}&\textbf{Advantages}& \textbf{Limitation} & \textbf{Summary of Prediction Method} & \textbf{EM}\\
        \hline
       \cite{quintanar2021predicting}\newline2021 &1  & 5s &Providing more information (position and direction), to assess its impact on the model's performance.& The inputs are only the track history of the target vehicle.& Modified a standard transformer and utilising the augmented data for the context of vehicles.&ADE, FDE\\
        \cite{liu2021multimodal}\newline2021&  1,6 & 3s &Create a region-based training technique that guarantees that each proposal can capture a certain mode. &Interaction of vehicle to infrastructure is missing. & Stack transformers are used to model multimodality at feature level with fixed independent proposals.&minADE minFDE MR\\
       
      \cite{zhao2021spatial}\newline2021& 1 &  5s&To gauge how socially interactive agents are, a channel-wise module is inserted.&Trajectory prediction depends on historical data, yet future trajectories are still unpredictable.& Transformer networks with residual layers are used to predict the trajectory and learn interaction aspects.& ADE, FDE, RMSE  \\
      \cite{chen2021s2tnet}\newline 2021 &1&3s &Extract information about interactions both in terms of spatial and temporal dimensions. & There needs to be integrated map data. & Transformer is made to record all traffic agents' spatiotemporal interactions, not just their spatial neighbours.& WSADE,\newline WSFDE
\\
       %1. a module for intention-aware nonaggressive decoding queries.
       \cite{chen2022vehicle}\newline2022&  - & 5s & The graph attention and the sparse self-attention mechanisms are used for social interaction and temporal interaction, respectively.&Environmental data, such as detailed maps and traffic conditions, are excluded.& Transformer uses social and temporal attention modules to capture correlations from raw trajectory data.&RMSE\\
       
       \cite{hou2022structural}\newline2022 &1 & 5s &1. Simultaneously learns the temporal and spatial relationships between various SVs; 2. An approach without recurrences to enhance the speed-accuracy trade-off in long-term prediction. & The effectiveness of the attention weights in the attention layers is unconfirmed in terms of their ability to reveal the driver's true attention to the other SVs.& Structural Transformer employed by utilising the two-layer encoder-decoder architecture for parallel trajectory prediction for multiple SVs. &Final position error, Time cost\\

         \cite{huang2022multi} \newline2022&  1,6& 3s&Modifies the multi-head attention method to support multi-modal prediction.& Traffic conditions and rules are excluded.& Extracting relationships between interacting actors using a transformer encoder.&minADE minFDE MR \\
       
      \cite{ngiam2022scene} \newline2022&  1,6 &5s&1. Model the query using a masking method; 2. Use the attention method to integrate features from different road components, agent interactions, and time steps.&Captures  agent-to-agent interactions, excluding infrastructure interactions with agents. & transformer to carry out conditional motion prediction, goal-conditioned prediction, and motion prediction.& minADE, minFDE, miss rate, and mAP.\\

      \cite{wang2023safety}\newline2023&-&5s &This takes into account the driving habits of other vehicles. & Insufficient generality for trajectory planning.
&Transformers are used to encode the vehicle interaction.& ADE,FDE\\

         \cite{gao2023dual}\newline2023&1 &4s&The intention prediction of the target vehicle used as input for the trajectory prediction module.&Only highway scenarios may be predicted using this approach.&Dual transformer utilised: one for intention prediction, the other for trajectory prediction.& RMSE\\
      % to put into practise forecasting the lane-change intentions and trajectories of target vehicles
     \cite{wang2023lane}\newline2023& 6&3s&Reduce the model's parameters to make it smaller in size.&The number of generated trajectories for the target vehicle has to be predetermined.& Stack of transformers to connect the highways' and agents' features.&minADE, minFDE, miss rate\\
       % & &\checkmark  &[],[], [2023] & Transformer layer is introduced to extract the relationships between the agents and road graph to predict the trajectory\\
   \hline
    \end{tabular}}
   
    \label{tab:my_label}
\end{table*}

\subsubsection {Attention Mechanism} 

The Attention Mechanism (AM) is a cognitive model that approximates human thought processes by allowing for the efficient extraction of high-value information from a large volume of data using limited attentional resources. It is frequently used in deep learning tasks such as speech recognition \cite{chorowski2015attention}, image classification \cite{mnih2014recurrent}, and natural language processing \cite{hu2020introductory}, with self-attention \cite{vaswani2017attention} being a popular method for identifying the weights and new context vectors based on the input sequence.
% According to Vaswani \textit{et al.} \cite{vaswani2017attention}, self-attention, also known as intra-attention, is the attention function that identifies the weights and new context vector based on the input sequence. This approach is commonly used in various applications due to its effectiveness and simplicity.
Several recent studies have employed the attention mechanism for trajectory prediction and intention estimation. Hao \textit{et al.} \cite{hao2020attention} proposed an encoder-decoder architecture combining GRU and self-attention, while Yan \textit{et al.} \cite{yan2020trajectory} investigated a self-attention architecture with two types of self-attention mechanisms for the driving lane and driving context. Kim \textit{et al.} \cite{kim2021lapred} used self-attention to concentrate on features from the target vehicle's preferred lane, and Fu \textit{et al.} \cite{fu2021trajectory}and Yu \textit{et al.} \cite{yu2021dynamic} employed attention between the encoder and decoder components to selectively draw attention to particular context vector properties. 

According to Wu \textit{et al.} \cite{wu2021hsta} and Meng \textit{et al.} \cite{meng2021intelligent}, the model can learn important spatial and temporal components for predicting and anticipating the movements of nearby vehicles. These models use a spatial attention layer to combine data from surrounding vehicles and a temporal attention layer to account for the temporal relationships between object agents.
Lin \textit{et al.} \cite{lin2021vehicle} proposed the STA-LSTM, which combines spatial and temporal information with an attention mechanism to explain how past trajectories and nearby vehicles affect the ego vehicle. Additionally, Kim \textit{et al.} \cite{kim2022diverse} proposed a model with a Baseline Network and Trajectory Proposal Attention, which is designed to simulate interaction-aware prediction. More recent work includes TP2Net \cite{hu2022trajectory}, a trajectory prediction network that uses temporal pattern attention to extract latent multimodal driving information, and yang \textit{et al.} \cite{yang2022lane} investigating the spatiotemporal dynamics between the ego vehicle and nearby cars,  utilized spatiotemporal attention mechanisms in LSTM networks to perform lane change prediction and the trajectories of the vehicles.
Several studies have employed multi-head attention and AM to extract information from lanes and vehicles, and model traffic interactions by analyzing attentions extracted from LSTM encoders. For instance, Kim \textit{et al.}  \cite{kim2020multi} utilizes multi-head attention to extract lane and vehicle information to predict future trajectory distributions. Messaoud \textit{et al.} 
 \cite{messaoud2020attention} also employs attention extracted from LSTM encoders to model traffic interactions. In Messaoud \textit{et al.}'s \cite{messaoud2021trajectory} model, each attention head simulates a possible interaction between the target and context features.
Hasan \textit{et al.} \cite{hasan2023mals} involves two Multi-Head Attention layers to capture the social and temporal interactions among vehicles. Additionally, incorporated a Multi-Head Attention-based decoder that includes an LSTM layer to decode the social and temporal interactions of the vehicles in a step-by-step manner. In summary, the attention mechanism in trajectory prediction improves the model's ability to focus on relevant information, handle variable-length sequences, provide interpretability, and enhance robustness to noise. However, it comes with potential drawbacks related to computational overhead, model complexity, attention bias, and data dependency. 
%These factors should be carefully considered and balanced when applying the attention mechanism in trajectory prediction.
%By attending to important features, the model can make more accurate and informed predictions, contributing to safer and more reliable autonomous driving systems.
Table V presents a comprehensive summary of Attention-based approaches for trajectory prediction. It includes important information such as the prediction horizon (measured in seconds (s)), the number of trajectories predicted, and the evaluation metrics used for training and testing. Additionally, the table provides insights into the strengths and weaknesses of each study.
%Table 5 provides a summary of Attention-based approaches for trajectory prediction, including the prediction horizon (in second (s)), the number of trajectories predicted, along with the evaluation metrics used for training and testing. The table provide the strengths and weaknesses of each study.

\subsubsection{Transformer}Transformer is a neural network design that utilizes an attention mechanism concept and has been employed in various projects such as object detection \cite{han2022few}, image segmentation \cite{cheng2022masked}, posture estimation \cite{tran2022combination}, tracking, and trajectory prediction \cite{weng2022whose}. It was initially utilized for machine translation in Natural Language Processing (NLP) \cite{bracsoveanu2020visualizing} and outperformed recurrent neural networks. Researchers have found the Transformer model to be effective for trajectory prediction, with Quintanar \textit{et al.} \cite{quintanar2021predicting} modifying a standard transformer to incorporate past trajectories as an input feature extracted from aerial view photo datasets. Another approach suggested by Liu \textit{et al.} \cite{liu2021multimodal} involves a multi-modal prediction architecture consisting of stacked transformers that gather features from historical trajectories, road data, and social interaction. Meanwhile, Zhao \textit{et al.} \cite{zhao2021spatial} utilized a transformer network with residual layers to predict trajectories that account for interaction, using fully linked feed-forward networks and pooling operations to integrate geographical data and enable the transformer to learn interaction aspects. 
The Spatio-Temporal Transformer Networks (S2TNet) \cite{chen2021s2tnet} uses spatio-temporal Transformer to represent spatio-temporal interactions and temporal Transformer to handle temporal sequences. Chen \textit{et al.} \cite{chen2022vehicle} propose a novel non-autoregressive model for predicting vehicle trajectories based on transformers, utilizing a self-attention module to define the dynamic variation in social behavior and a graph attention module to represent the interactions between vehicles. The Structural Transformer suggested by Hou \textit{et al.} \cite{hou2022structural} is a recurrence-free multi-sequence learning network that grasps interactions between surrounding vehicles along both temporal and geographical dimensions simultaneously. Huang \textit{et al.} \cite{huang2022multi} propose a Transformer-based multi-modal trajectory prediction model using a multi-head attention Transformer layer to model the relationship between interacting agents. The SceneTransformer  is a transformer-based model introduced by Ngiam \textit{et al.} \cite{ngiam2022scene} that uses attention to mix features from agent interactions and road graphs in both space and time. The LaneTransformer proposed by Wang \textit{et al.} \cite{wang2023lane} combines the characteristics of the features between the roadways and the agents using a stack of transformer blocks, and high-order interactions are aggregated using an attention-based block. Wang \textit{et al.} \cite{wang2023safety} propose a mixture-of-experts approach utilizing a transformer to model the interactions between vehicles explicitly considering their driving styles for building a multimodal motion planner. The study in \cite{gao2023dual} proposes a dual Transformer model to demonstrate the relationship between intentions and trajectories for the target vehicle. As demonstrated by these studies, the use of transformers provides several advantages in handling time-series data in trajectory predictions. To summarize, transformers have shown their potential in trajectory prediction by capturing complex dependencies and interactions. They offer scalability, transfer learning capabilities, and the ability to handle multiple agents. However, they require substantial computational resources and may have challenges in interpretability and data efficiency. 
%Careful consideration should be given to the specific requirements and constraints of the trajectory prediction task when applying transformer models. 
Table VI summarizes the Transformer-based approaches for trajectory prediction, presenting key details such as the prediction horizon (measured in seconds (s)), the number of trajectories predicted, and the evaluation metrics employed for training and testing. Furthermore, the table highlights the strengths and weaknesses of each study. 
%Table 6 provides a summary of Transformer-based approaches for trajectory prediction, prediction horizon (measured in seconds), the number of trajectories predicted along with the evaluation metrics used for training and testing. The table also highlights the strengths and weaknesses of each study.

\subsection{Vision Based Modelling}
There are two types of prediction methods, which differ in how they formulate predictions. The first is the Bird-Eye-View (BEV) approach, which uses an algorithm to process data from a top-down, map-like view. The second is ego-camera prediction, which involves viewing the world through the perspective of the ego-vehicle. However, the ego-camera approach is generally more challenging than the BEV approach due to various factors \cite{teeti2022vision}. Firstly, the BEV approach offers a broader field of view and more accurate predictions, whereas the ego-camera approach has a narrower field of vision and a limited prediction horizon. Additionally, the ego-camera approach is more prone to obstructions than the BEV approach. Despite these difficulties, the ego-camera approach is still more beneficial than the BEV approach because most vehicles do not have access to cameras that can locate target agents and BEVs on the road. 
Therefore, a prediction system should be able to view the world from the perspective of the ego vehicle, as demonstrated in Fig. 9. The illustrations of various vision-based techniques and their contribution to solving the trajectory prediction task in Autonomous Vehicles (AVs) are depicted in Fig. 10. Each technique makes a roughly equal contribution to the trajectory prediction in AVs research paper. This section highlights the inclusion of Convolutional Neural Networks (CNNs) and Graph Neural Networks (GNNs) in addressing this domain.
%The Convolutional Neural Network (CNNs) and Graph Neural Network (GNNs) are illustrated in this section.
\begin{figure}[]
    \centering
     \includegraphics[width=0.5\textwidth]{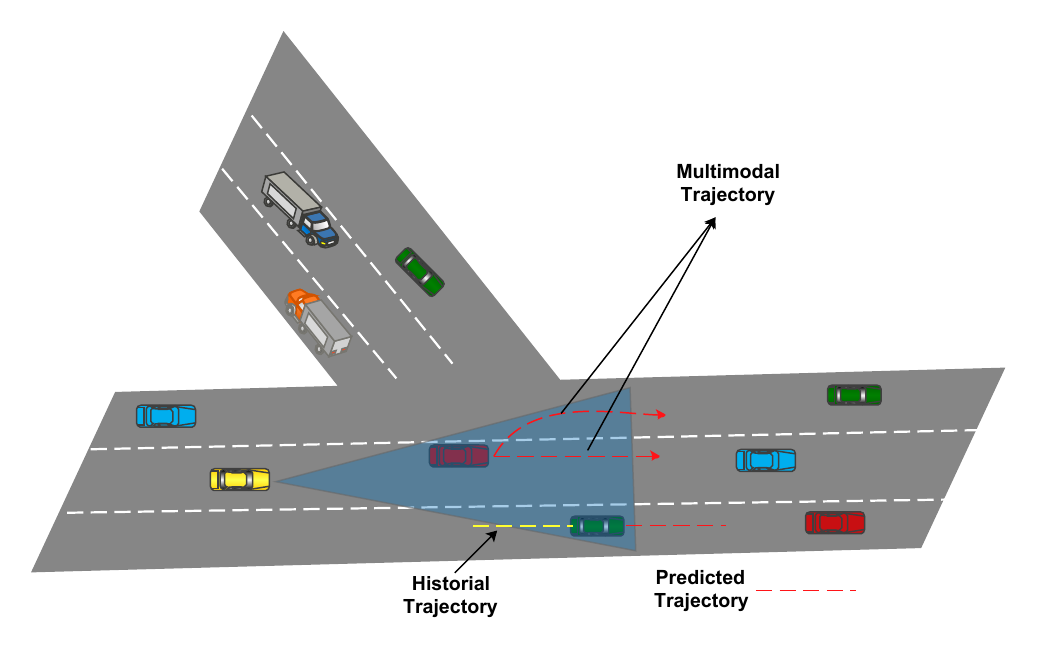}
    
    \caption{The ego-camera prediction algorithm adopts the perspective of the ego-vehicle to observe the environment, identifying relevant information about the target and surrounding vehicles, such as bounding boxes, RGB frames, position, speed, type, etc. This enables the algorithm to make predictions about their future locations. }
    \label{fig:my_label}
\end{figure}
\begin{figure}[]
    \centering
     \includegraphics[width=0.5\textwidth]{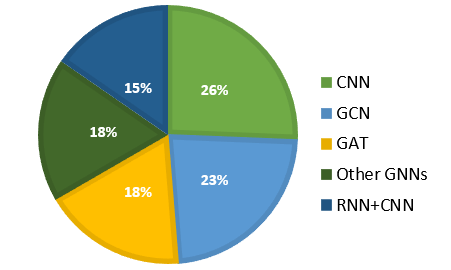}
    \caption{Participation of Research articles in trajectory prediction task using Vision-based approaches. }
    \label{fig:my_label}
\end{figure}

\begin{table*}[]
\caption{Summary of Convolution Neural Network based methods: Related Work \& Year, No. of Predicted Trajectories for each vehicle, Prediction Horizon (PH), Advantages and Limitation of each work,
Summary of each work, and Evaluation metric (EM)
\newline - Predicting any number of trajectories for the work/This information is not available for work } \resizebox{17.5cm}{4.5cm}{
    \centering
    \begin{tabular}{|p{0.7cm}|p{1.2cm}|p{0.5cm}|p{4cm}|p{4cm}|p{4cm}|p{1.7cm}|}
    \hline
         \textbf{Ref.}&\textbf{\# trajectories }&\textbf{PH}&\textbf{Advantages}& \textbf{Limitation} & \textbf{Summary of Prediction Method} & \textbf{EM}\\
        \hline
       \cite{cui2019multimodal}\newline2019  &- &6s &In closed-course tests, the suggested technique was tested within self-driving cars.&The process creates a raster image that encodes the context of each actor,& Using MobileNetV2 \cite{sandler2018mobilenetv2} as a feature extractor.&Displacement errors\\
      
          \cite{djuric2020uncertainty}\newline2020& 1 & 3s&A fleet of self-driving cars were fitted with the system after rigorous offline and online testing.&Vehicles interactions are not taken into account.& CNN inputs a raster image to extract significant features.& Displacement errors\\ 
              
        \cite{phan2020covernet}\newline2020&  1,5,10,15 & 6s\newline  &Frame the problem of trajectory prediction as a classification task over a wide range of trajectories.&The model outperforms for the urban driving datasets only.&The model's foundation is ResNet-50 \cite{targ2016resnet}.&minADE,FDE,\newline HitRate \\
        % Using CNNs, plot the future trajectory of agents based on their history and scene context
      
        \cite{marchetti2020mantra}\newline2020 &1,5 & 4s&The ability of the memory to assimilate fresh samples lowers the error on unobserved data.&The past trajectory of the target vehicle and road-related information are considered.&A MANN with a non-episodic memory.&ADE,FDE\\

        %CNN architecture utilizing context rasterization techniques for the trajectory prediction of VRUs.
        \cite{zhang2021resnet}\newline2021& 1,3&-&To forecast various agents' driving trajectories.&Model's three output trajectories only and their associated confidence intervals. & Apply the ResNet\cite{targ2016resnet} model to the input images to learn complicated feature representations.&NLL\\
        \hline
    \end{tabular}}
  
    \label{tab:my_label}
\end{table*}
\begin{table*}[]

 \caption{Summary of Graph Convolution Network-based methods: Related Work \& Year, No. of Predicted Trajectories for each vehicle, Prediction Horizon, Advantages and Limitation of each work,
Summary of each work, and Evaluation Metric (EM)} \resizebox{17.5cm}{6cm}{
    \centering
    \begin{tabular}{|p{0.7cm}|p{1cm}|p{0.5cm}|p{4cm}|p{4cm}|p{4cm}|p{1.5cm}|}
    \hline
         \textbf{Ref.}&\textbf{\# trajectories}&\textbf{PH}&\textbf{Advantages}& \textbf{Limitation} & \textbf{Summary of Prediction Method} & \textbf{EM}\\
        \hline
         \cite{li2019grip}&1 & 5s &The relationships between various traffic agents are described by using fixed graph. &When employed in urban traffic settings, it could see some performance decrease.&Extracts features using several graph convolutional blocks.&RMSE\\
    \cite{li2019grip++} &1&5s&The relationships between various traffic agents are described using fixed and dynamic graphs.&Incorporate GRIP++ into a perception and route planning module to further assess overall performance.&Extracts features using several graph convolutional blocks.& RMSE,\newline WSADE, WSFDE\\
        \cite{jeon2020scale}& 1  &4s&EGCN-based interaction embedding analyses the inter-vehicle interactions inherently.&Model cannot take into account the road structures.& Model is  fully scalable to handle any number of vehicles in driving scenario. &RMSE\\
        % With the help of edge features, EGCN updates the node feature of nodes.
         %[133]
      \cite{chandra2020forecasting} &1 &5s  &1. The interactions between any two road agents are represented by the adjacency matrix; 2. There is no assumptions regarding the dimensions and geometry of the road-agents.& Training is slow  for computing the
traffic-graphs. & A two-layer GNN-LSTM structure is used to resolve the trajectory prediction task.&ADE,FDE\\
%Using two different streams, it haphazardly predicts trajectories as well as road-agent behaviour.corresponding Laplacian matrices
        \cite{zhao2020gisnet} & 1& 5s &To forecast future trajectory, a model can combine the information gathered from both non-Eculidan and Euclidean domains.&When constructing the topology of the communication network, the time information was not taken into account.& GCN network allows vehicles to communicate information to take into account how the surrounding vehicles are changing and adapting to the environment.&RMSE\\
        \cite{sheng2022graph}& 1& 5s &Adjacency matrix used to describe the
intensities of mutual influence between vehicles. &The impact of neighbouring automobiles from various angles are ignored. & This network uses a GCN to address spatial interactions and a CNN to capture temporal data. &RMSE\\
   \cite{xu2022group}&1 &5s &A matrix that may be adjusted and supplemented to make up for the shortcomings of extracting characteristics from fixed topology& Spatial and temporal features are not processed simultaneously.&  The stacked GCN module extracts the global spatiotemporal characteristics of the historical vehicle trajectory data.&RMSE,ADE,\newline FDE\\
  \cite{xu2023mvhgn}&1&5s&By describing various logical correlations of numerous road agents, a multi-view logical network is proposed.&It is a difficult challenge to confidently estimate the trajectory based on the missing data and noisy samples.&The GCN module mines the logical-physical properties at the micro level.&WSADE,\newline WFADE\\ \hline
%[2023]
    \end{tabular}
 }
    \label{tab:my_label}
\end{table*}
\subsubsection{Convolutional Neural Network}
Convolutional Neural Networks (CNNs) have been successfully applied to various computer vision tasks, including trajectory prediction. Although CNNs are primarily designed for image data, they can be adapted for trajectory prediction by treating the trajectory sequence as a structured grid-like input. Recently, CNN has shown success in various tasks, including machine translation \cite{gehring2016convolutional} and computer vision \cite{bhatt2021cnn}. In the context of trajectory prediction for autonomous vehicles, CNN is commonly used for vision-based prediction, where features are extracted from images captured by frontal cameras. Nikhil \textit{et al.} \cite{nikhil2018convolutional} found that using CNN for trajectory prediction is superior to RNN, as trajectory has significant spatio-temporal continuity. They stacked the convolutional layer after a fully connected layer to create time continuity and used a fully connected layer to output the future trajectory, taking the past trajectory as input. This CNN-based network operates faster, according to experiments.

However, most techniques that use the CNN framework take a Bird's-Eye View (BEV) as their input, displaying a top-down view of the traffic situation. BEV images can be created using multiple data sources, including LiDAR point clouds, Occupancy Grids (OG), and High-Definition Maps (HD-Maps). Some recent studies utilized CNN to extract features from sophisticated BEV representations. For example, MobileNetV2  \cite{sandler2018mobilenetv2}, a memory-effective CNN designed for mobile apps, was used in \cite{cui2019multimodal},\cite{djuric2020uncertainty} to output potential trajectories and their likelihoods. The trajectory prediction of Vulnerable Road Users (VRUs) is addressed in \cite{chou2020predicting} through a new rapid CNN architecture that utilizes context rasterization techniques \cite{djuric2020uncertainty}. In \cite{phan2020covernet}, the vehicle state and the raster image were used to build a set of potential future trajectories, and the trajectory with the highest probability was selected as the future trajectory by examining semantic properties. A novel rapid CNN architecture was proposed in \cite{marchetti2020mantra} for trajectory prediction of vulnerable road users, where a Memory Augmented Neural Network (MANN) was used to produce multimodal trajectories.

Recent studies have also proposed new techniques that forecast trajectory using CNN and produce cutting-edge results. For instance, Gilles \textit{et al.} \cite{gilles2021home} generates a heatmap of the agent's potential future, while Ye \textit{et al.}  \cite{ye2021tpcn} uses the point cloud learning method to incorporate both spatial and temporal data into trajectory prediction. Zhuoren \textit{et al.} \cite{zhang2021resnet} used ResNet-50 \cite{targ2016resnet} to anticipate the trajectories of AVs such as vehicles and pedestrians. ResNet-50 \cite{targ2016resnet} can effectively collect information from multiple dimensions to produce superior forecasts with the three trajectories and their confidence levels. While processing raster maps with CNN involves significant computational costs and information loss, vector maps can be used as nodes in Graph Neural Networks (GNN) for trajectory prediction.
To summarize, CNNs offer advantages in capturing spatial patterns and recognizing spatial relationships in trajectory data. They are efficient in terms of parameter sharing and can handle larger datasets. However, they may struggle with modeling temporal dependencies and handling variable-length sequences. Table VII presents a summary of CNN-based approaches for trajectory prediction, including the prediction horizon measured in seconds (s) and the number of trajectories predicted. The table also provides an overview of the evaluation metrics used for training and testing, as well as highlighting the strengths and weaknesses of each study. 
%Table 7 provides a summary of CNN-based approaches for trajectory prediction, along with the evaluation metrics used for training and testing. The table also highlights the strengths and weaknesses of each study. 
The approaches for predicting vehicle trajectory based on GNN will be covered in the following sections.
% Careful consideration should be given to the specific requirements and characteristics of the trajectory prediction task when applying CNNs.
\begin{figure}
    \centering
    \includegraphics[width=0.5\textwidth , height=0.2\textwidth]{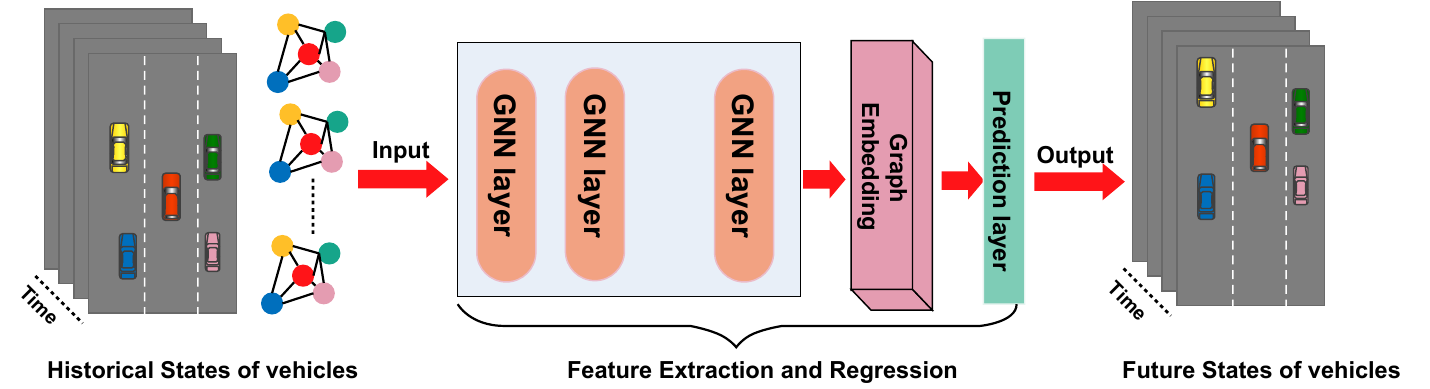}
    \caption{Depiction of the graph neural network for trajectory prediction task.}
    \label{fig:my_label}
\end{figure} 
\begin{table*}
 \caption{Summary of Graph Attention Network-based methods: Related Work \& Year, No. of outputs, Prediction Horizon (PH), Advantages and Limitation of each work,
Summary of each work, and Evaluation Metric (EM)} \resizebox{17.5cm}{5cm}{
    \centering
     \begin{tabular}{|p{0.7cm}|p{1cm}|p{0.5cm}|p{4cm}|p{4cm}|p{4cm}|p{1.5cm}|}
      \hline
         \textbf{Ref}&\textbf{\# trajectories}&\textbf{PH}&\textbf{Advantages}& \textbf{Limitation} & \textbf{Summary of Prediction Method} & \textbf{EM}\\
      
        % Forecast the future paths of heterogeneous traffic-agents\\
      
  %  {p{1.5cm}|p{2cm}|c|p{0.5cm}|p{6cm}}
    %  \textbf{Class} & \textbf{Advantages} & \textbf{Disadvantages} & \textbf{Ref.} & \textbf{Summary of Prediction Method} \\
      % Capable of handling spatial dependencies
       %[94]  &  && & &&&A bicycle vehicle kinematics model is embedded with CNN for trajectory prediction \\

        \hline
        % takes into account how interactions between traffic agents may affect motion
       
       \cite{mo2020recog}\newline2020 &1 & 5s &Simulate the diverse interactions between infrastructures and vehicles.& A more detailed HD map not included rather than a picture-based one for interaction with the environment.& A heterogeneous digraph is used by GAT to derive spatial interaction information.& ADE,FDE\\
        \cite{ding2021ra}\newline2021& 1 & 5s &On the scene graph, two types of relation edges are created separately to describe various affects that cars and places have on one another. &The model is not suitable for problems with unbalanced data learning.& A dual GAT is used to simultaneously describe the space attraction and vehicle-wise repulsion.& RMSE\\
        %A graph-attention module with two independent layers are created to represent the information update process using various forms of impacts separately.
        \cite{mo2022multi}\newline2022& 1&8s &A gate mechanism allowing for the selective sharing of maps among all target agents.& Hardware configurations in practical applications will have an impact on the data's quality and availability.& Inter-agent interaction modeling with a heterogeneous edge-enhanced GAN. &RMSE,ADE,\newline FDE\\
        \cite{liu2022multi}\newline2022& 1,5 & 3s &1.adjacency matrix outlining agent connection; 2.vector representation method for the map information. &1.dynamic
features are excluded; 2. Prediction accuracy can be further improved.& Graph Isomorphism Network has been used to update the node feature and include information from nearby nodes.& minADE,\newline minFDE\\
        \cite{zhang2022ai}\newline2022& 1&  5s& 1. Forecasts agent speeds rather than spatial coordinates; 2. Asymmetrical adjacency matrix depicts the relationships between traffic agents.& There are no rasterized maps or LiDAR data included.& These models display traffic scenes as graphs and GAT to explicitly record the interactions from the environment. & RMSE,\newline WSADE,\newline WFADE\\
        \cite{zhang2022trajectory}\newline2022 &1&5s&Accurately simulating spatial-temporal interactions from the environment.& Model fixed the maximum number of traffic agents that it could handle.&With the use of attention mechanisms, the GAT block is used to represent the interactions between agents and infrastructure.&ADE,FDE\\
        \cite{meng2023trajectory}\newline2023&1,6,10& 6s &The introduction of an ice and snow mask system that simulates situations where lane lines are covered.& It is crucial to the accuracy of the absolute forecast that the proportions of different input data be considered. & Information gathering is made possible via a two-layer GAT, which also explains node correlation.&minADE,\newline minFDE,\newline MR\\
         \hline
    \end{tabular}}
   
    \label{tab:my_label}
\end{table*}

\subsubsection{Graph Neural Network}
When considering prediction techniques that take interaction-related factors into account, each element of the environment can be viewed as a node in a graph. However, many real-world applications generate data from non-Euclidean spaces, and traditional deep learning-based methods that analyze Euclidean spatial data perform poorly in such cases. Each scene can be represented as an irregular graph with variable-sized unordered nodes, and some crucial operations, such as convolution, are not directly applicable to the graph due to variations in the number of nearby nodes. Nevertheless, every node in the graph is connected to other nodes by edges, which can be used to determine the interdependence of various objects. Graph Neural Networks (GNNs) are highly suited for vehicle trajectory prediction challenges based on interaction-related elements \cite{wu2020comprehensive}. The methodology is described in Fig. 11. This idea is supported by Diehl \textit{et al.} \cite{diehl2019graph}, who demonstrate the effectiveness of trajectory prediction using two well-known graph networks: the Graph Convolutional Network (GCN) and the Graph Attention Network (GAT).

\paragraph{Graph Convolutional Network} 
The Graph Convolutional Network (GCN) is a popular technique in the field of graph neural networks. It extends the convolution operation from traditional image data processing to graph data processing. The key idea is to create a mapping function that can extract interaction-aware features from the node features in the network and their neighboring nodes.
Li \textit{et al.} \cite{li2019grip} proposed GRIP, a graph convolutional network-based trajectory prediction model that considers the interaction-related factors by treating vehicles as nodes within the network at each sampling time. GRIP utilizes a fixed graph network to describe the interaction-related characteristics between traffic participants and employs an LSTM encoder-decoder to forecast the trajectory of nearby vehicles using the output of GCN. To improve the accuracy of GRIP, Li \textit{et al.} \cite{li2019grip++} proposed GRIP++, which uses both fixed and dynamic graph networks and achieved top ranking in the Baidu Apolloscape dataset \cite{huang2019apolloscape} at the end of 2019. Jeon \textit{et al.} \cite{jeon2020scale} proposed SCALE-Net, which can predict the trajectories of any number of nearby vehicles while maintaining performance by using an Edge-Enhance Graph Convolutional Network (EGCN) \cite{gong2019exploiting} to learn edge features in the traffic flow. Chandra \textit{et al.} \cite{chandra2020forecasting} proposed a two-layer GNN-LSTM structure to resolve the trajectory prediction issue by using an LSTM encoder-decoder in the first layer to predict the future trajectories of traffic participants and a weighted dynamic geometric graph network in the second layer to represent the interaction-related characteristics of traffic participants. Zhao \textit{et al.} \cite{zhao2020gisnet} proposed a spectrum-based GCN network that allows all vehicles in the scene to communicate information to take into account how the surrounding vehicles are changing and adapting to the environment. Sheng \textit{et al.} \cite{sheng2022graph} proposed the GSTCN network, which uses a GCN to address spatial interactions, a CNN to capture temporal data, and a gated recurrent unit network to encrypt and decrypt the spatiotemporal properties to produce future trajectory distributions. Xu \textit{et al.} \cite{xu2022group} proposed a group vehicle trajectory prediction model with a global spatiotemporal graph that can thoroughly analyze the temporal and geographical association between previous vehicle trajectories. Dongwei \textit{et al.} \cite{xu2023mvhgn} suggested the MVHGN forecast, a graph neural network-based model for predicting the future paths of heterogeneous traffic-agents that employs a multi-view logical network by fusing various logical correlations and the multi-view logical characteristics derived by the graph convolution module. In summary, GCNs offer a promising approach for trajectory prediction by explicitly modeling the spatial dependencies and relationships among objects. They can effectively capture contextual information and handle irregular graph structures. However, scalability, graph construction, and temporal dependency modeling should be carefully considered when applying GCNs to trajectory prediction tasks. Table VIII provides a summary of GCN-based approaches for trajectory prediction including the prediction horizon measured in seconds (s) and the number of trajectories predicted, along with the evaluation metrics used for training and testing. The table also highlights the strengths and weaknesses of each study.

     % the GCN network enables the extraction of vehicle dependencies and the simultaneous prediction of the trajectories of surrounding vehicles.
    
\begin{table*}[]
 \caption{Summary of other Graph Neural Network-based methods: Related Work \& Year, No. of Predicted Trajectories for each vehicle, Prediction Horizon (PH), Advantages and Limitation of each work,
Summary of each work, and Evaluation Metric (EM)} \resizebox{17.5cm}{4cm}{
    \centering
    \begin{tabular} {|p{0.7cm}|p{1cm}|p{0.5cm}|p{4cm}|p{4cm}|p{4cm}|p{1.7cm}|}
    \hline
         \textbf{Ref.}&\textbf{\# Trajec\newline tories}&\textbf{PH}&\textbf{Advantages}& \textbf{Limitation} & \textbf{Summary of Prediction Method} & \textbf{EM}\\
        \hline
       \cite{gao2020vectornet} 2020 &1&3s& Vectorized representation of the HD map and agent dynamics.& Recalculating the VectorNet features for every target would increase the  computational cost with the number of targets being predicted.&  Hierarchical GNN, where the first level takes advantage of the spatial proximity of certain road elements and the second level simulates the high-order interactions between all elements.&ADE\\
        
         \cite{liang2020learning} 2021&1,6 & 3s&1.Create a lane graph using the raw map data; 2. exploit all four different lane-agent interactions. &Case of extreme acceleration  does not captures well by the model.&1D CNN to handle the input trajectory data and uses along-lane dilation and numerous adjacency matrices to extend graph convolutions to provide the map features. & minADE,\newline minFDE, MR \\
       \cite{zhao2021tnt} 2021& 1,6&3s& Based on the provided scene context, TNT can model the scene context using any acceptable context encoder.&Forecasting over a medium time horizon. & There are three stages to TNT: target prediction, target-conditioned motion estimation and scoring and selection. & minADE,\newline minFDE,MR\\
          \cite{gu2021densetnt} 2021& 1,6&3s&Provide an offline optimization-based method to supply our final online model with several future pseudo-labels.&The model is trained in urban datasets only.&Estimates dense target candidate probabilities without using heuristic anchoring.&minADE,\newline minFDE,MR\\
         \cite{zeng2021lanercnn} 2021& 1,6&3s& The actor-to-actor and actor-to-map relations are distributedly and map-awarely captured by LaneRCNN. &Predict only single vehicle future trajectories. & To encode each actor's previous motions and the topology of the local map, learn a local lane graph representation for each actor.&minADE,\newline minFDE,MR\\ 
\hline
    \end{tabular}
   }
    \label{tab:my_label}
\end{table*}
\paragraph{Graph Attention Network} 
% Sequence-based tasks are now frequently using the attention mechanism. Its benefit is that it can increase the impact of the data's most crucial component. 
The method for collecting data from the one-hop neighborhood varies greatly between Graph Attention Network (GAT) and GCN, with GAT employing the attention mechanism in place of the statically normalized convolution process. Velickovi'c \textit{et al.} \cite{veličković2018graph} proposed the GAT. In \cite{mo2020recog}, an encoder-decoder design was used along with GAT to extract spatial interaction information from a heterogeneous digraph, consisting of automobile and local road map vertices. The Repulsion and Attraction Graph Attention (RAGAT) model was introduced in \cite{ding2021ra}, which uses two stacked GATs to predict trajectories based on free space and vehicle condition information. In \cite{mo2022multi}, a three-channel system with a heterogeneous edge-enhanced graph attention network was developed to address the heterogeneity of vehicles in a scene. A directed edge-featured heterogeneous graph was used to represent inter-agent interactions in traffic, and a gate mechanism was added for selective map sharing among target agents. Liu \textit{et al.} \cite{liu2022multi} proposed a multi-agent, multi-modal trajectory prediction framework using Graph Attention Isomorphism Networks (GAIN), which consisted of three attention blocks. AI-TP was introduced in \cite{zhang2022ai} to forecast multiple SV trajectories using GAT for interaction information, followed by two convolutional Gated Recurrent Units (GRU) networks. Zhang \textit{et al.} suggested the Gatformer model in \cite{zhang2022trajectory} for predicting future movements of nearby traffic agents while considering spatial-temporal connections, using graphs and GAT to capture environmental interactions, and integrating the Transformer encoder-decoder. In \cite{meng2023trajectory}, a two-layer GAT was used for information aggregation and node correlation explanation, with a multi-head attention mechanism to project surrounding states to the graph and explain interactions between vehicles and the traffic flow state. %In summary, GATs offer an effective approach to trajectory prediction by leveraging attention mechanisms to capture important spatial dependencies in the scene. They adaptively integrate features from neighboring nodes and provide interpretable attention weights. However, computational complexity and limited temporal dependency modeling should be taken into account when applying GATs to trajectory prediction tasks. 
In summary, GATs enable the model to attend to relevant nodes (e.g., vehicles, pedestrians) in the graph, assigning different weights to capture the importance of each node's features for predicting the trajectory of a specific object. However, The performance of GATs heavily depends on the quality and representation of the graph structure. Designing an appropriate graph representation and considering the selection of nodes and edges is crucial for achieving optimal results.
Table IX summarizes the GAN-based approaches for trajectory prediction, highlighting the number of trajectories predicted and the prediction horizon measured in seconds (s). The table also provides insights into the strengths and weaknesses of each study, along with the evaluation metrics used for training and testing.
%Table 9 provides a summary of GAN-based approaches for trajectory prediction, showcasing the number of trajectories predicted, the prediction horizon measured in seconds (s), and the evaluation metrics used for training and testing. The table also highlights the strengths and weaknesses of each study
%Table 9 provides a summary of GAN-based approaches for trajectory prediction, along with the evaluation metrics used for training and testing. The table also highlights the strengths and weaknesses of each study.

\paragraph{Other Graph Neural Network}
High Definition (HD) maps play a crucial role in trajectory prediction for autonomous vehicles. HD maps provide detailed information about the road network, including lane markings, traffic signals, and road boundaries, which can help predict the future trajectory of a vehicle or pedestrian more accurately. Initially, Benz \textit{et al.} \cite{ziegler2014making} utilized HD maps for predicting trajectories, followed by determining the vehicle's future trajectory along the lane based on map topology using related lane information. However, this method did not consider interaction-related factors. To improve trajectory prediction accuracy, researchers have incorporated GNN to capture interaction features between vehicles and maps as well as between vehicles, following the introduction of the Argoverse dataset \cite{chang2019argoverse} with vector maps.
Gao \textit{et al.} \cite{gao2020vectornet} proposed VectorNet, a GNN-based system that employs nodes to represent both the vector maps and vehicles in the scene for trajectory prediction. Liang \textit{et al.} \cite{liang2020learning} integrated CNN-extracted vehicle features and GCN-extracted lane features from vector maps for trajectory prediction. Zhao \textit{et al.} \cite{zhao2021tnt} presented a target-driven method called target-driven trajectory prediction (TNT) that selects sparse goal anchors and the optimal route to the target using VectorNet-extracted map features. DenseTNT 
\cite{gu2021densetnt} outperforms TNT in performance by evaluating dense goal candidates. Zeng \textit{et al.} \cite{zeng2021lanercnn} utilized LaneRCNN to represent local lane maps and interaction modules to account for interaction factors between participants' historical trajectories and local map topology. Researchers are exploring ways to integrate multiple sources of information, including HD maps, sensor data, and machine learning algorithms, to improve the accuracy and robustness of trajectory prediction for autonomous vehicles. 
Table X provides a summary of other graph neural network-based approaches for trajectory prediction, focusing on the number of trajectories predicted and the prediction horizon measured in seconds (s). The table also provides the strengths and weaknesses of each study, along with the evaluation metrics used for training and testing.

\begin{table*}[]
    \caption{Summary of CNN+RNN based methods: Related Work \& Year,No. of Predicted Trajectories for each vehicle, Prediction Horizon (PH), Advantages and Limitation of each work,
Summary of each work, and Evaluation Metric (EM)
\newline - Predicting any number of trajectories for the work}
 \resizebox{17.5cm}{6cm}{
    \centering
    \begin{tabular}{|p{0.7cm}|p{1cm}|p{0.5cm}|p{4cm}|p{4cm}|p{4cm}|p{1.5cm}|}
    \hline
         \textbf{Ref.}&\textbf{\# Trajec\newline tories}&\textbf{PH}&\textbf{Advantages}& \textbf{Limitation} & \textbf{Summary of Prediction Method} & \textbf{EM}\\
        \hline
      \cite{deo2018convolutional}\newline2018& - &5s&For effectively learning interdependencies in vehicle motion, convolutional social pooling is an enhancement over social pooling layers.& It depends on information from vehicle tracks and ignores visual and map-based cues for predicting manoeuvre classes and future trajectories.& On each vehicle trajectory, LSTM is used. The outcome is displayed in a BEV grid structure before being fed to a CNN. A total of six LSTM decoders receive the output.&RMS,\newline NLL\\
    
         \cite{zhao2019multi}\newline2019&1 & 5s&
         1. The use of convolutional fusion enables the modeling of interactions among multiple agents; 2. The incorporation of adversarial loss facilitates stochastic prediction learning. & Despite the limited agent-scene interactions and straight road lanes in the dataset, our model did not surpass NGSIM.& A concatenated vector comprising an agent's movement and a static scene encoded by a CNN and LSTM. & RMSE\\
       
         \cite{schreiber2019long}\newline2019&  -& 2s  &It created a recurrent skip architecture to handle missing input data.& Only a stationary sensor records the data.& 
 The spatial features are first extracted from the input image using a convolutional network. These features are then fed as input to the encoder-decoder LSTM.
 
 & F1-scores \\
 % From the input image, a convolutional network extracts spatial features. The encoder-decoder LSTM receives these properties as input. The output picture from the deconvolution network is the same size as the input image after the result is fed into it.
        
         \cite{chandra2019traphic}\newline2019& -&5s& Forecasting the trajectories of road agents in busy traffic footage. & 1. Dense diverse traffic serves as a model design inspiration; 2. Understanding the diverse sizes and shapes of road agents is necessary for simulating heterogeneous constraints. & A hybrid LSTM-CNN network is used to model the horizon-based and heterogeneous-based weighted interactions between road agents.& RMSE,\newline ADE,\newline FDE\\
      
         \cite{xie2020motion}\newline2020 &1  &30s &  1. The box plot is used to identify and get rid of anomalous vehicle trajectory values; 2. The model's hyper-parameters are optimised using a grid search approach. & Consider the historical information of surrounding vehicles. & Convolutional and maximum pooling layers in the CNN-LSTM framework extract interaction-aware features, while an LSTM and a fully connected layer is used for prediction.& RMSE, MAE, and deviation\\
   
         \cite{xu4135360vehicle}\newline2022  &1 &5s & The interaction of cars is depicted using a traffic graph; 2. The prediction mechanism takes into account road and speed parameters.& Future trajectories are not thought to be affected by the diversity of traffic agents, traffic regulations, or climate change. & The temporal characteristics of a vehicle are captured using a convolutional layer, spatial features are captured using graph operation layer and LSTM encoder-decoder to predict the future trajectories.& ADE,FDE\\
         \hline
     
    \end{tabular}}

    \label{tab:my_label}
\end{table*}

\begin{table*}[]
    \caption{Summary of Generative Adversarial Network-based methods: Related Work \& Year, No. of Predicted Trajectories for each vehicle, Prediction Horizon (PH), Advantages and Limitation of each work,
Summary of each work, and Evaluation Metric (EM) 
\newline - Predicting any number of trajectories for the work / This information is not available for work }
 \resizebox{17.5cm}{6cm}{
    \centering
 
    \begin{tabular}{|p{0.7cm}|p{1cm}|p{0.5cm}|p{4cm}|p{4cm}|p{4cm}|p{1.5 cm}|}
    \hline
      \textbf{Ref.}&\textbf{\# Trajec\newline tories}&\textbf{PH}&\textbf{Advantages}& \textbf{Limitation} & \textbf{Summary of Prediction Method} & \textbf{EM}\\
        \hline
        \cite{hegde2020vehicle}\newline 2020 & -&4.8s&1.The vehicle interactions are handled by the pooling mechanism; 2. based on the vehicle's past performance to forecast its upcoming course. & The model requires processing of the images before sending them to the GAN, so it cannot operate in real-time. & Future paths are generated by the generator, which is composed of an encoder and decoder network and a pooling module. The discriminator, which consists of an encoder, can classify the trajectories more precisely as genuine or fake.&ADE,FDE\\
      
          \cite{zhao2020novel}\newline 2020  & 1&30s&1. The conversion of each vehicle's position coordinates into normalised coordinates; 2. Based on the psychology of the driver, the vehicle turning model can improve the driving path.& 1. Does not include the interaction between the 
 vehicles and road information; 2. Consider only the urban road scenarios.&   Using past trajectory data, GAN is used to train and understand the driver's behavior.& MAE, RMSE, \newline average accuracy\\

         \cite{li2021vehicle} \newline 2021 &1& 6s&Prediction models that use rules as inductive biases. &Fail to take trajectory uncertainty and rule prioritization into account.& Signal Temporal Logic is viewed as a collection of discriminator features and a generator auxiliary loss.& ADE,FDE, \newline MaxDist \\
    
         % By adding noise to the generator's inputs, commercial trajectory predictors can be used to build it. Syntax tree features are utilised to reduce the training costs for the generator and/or improve the discriminator's performance.\\
         \cite{guo2023map}\newline 2023& 1,6 & 3s&To allow the global map to be reused, a graph query mechanism is presented.&The potential of HD maps for trajectory prediction tasks has to be investigated.&The generator side, the proposed model creates contextual features by fusing vehicle motion with high-definition maps. The addition of the map enhances the discriminator's basis for making judgments about the resulting trajectories.&MinADE,\newline MinFDE,\newline MR,DAC\\

          \cite{wang2020multi}\newline 2020&- &  5s  & 1. To record multi-vehicle interactions in both spatial and temporal dimensions, two parallel fusion modules have been constructed; 2. Under various conditions, display the effects of nearby vehicles on trajectory prediction.&1. Separately consider the spatial and temporal features for prediction; 2. Consider only the historical information of multi-vehicles. &  The generative adversarial network is used to handle the agent motion behavior's innate multi-modal properties.& RMSE\\
        
        \cite{wang2021multi}\newline2021 & -  &5s& 1. Considering the precise spatial distributions of agents during movement; 2. Multiple trajectory sequences are used to capture socially temporal relationships; 3. To formally describe the temporal correlations between interactions, use the social recurrent mechanism.& Only historical trajectories for the scenario's observed agents are taken; scene information for future trajectories is not included.&  GAN to produce multi-modal trajectory distribution. & ADE,FDE\\
         \hline

    \end{tabular}}

    \label{tab:my_label}
\end{table*}
\subsection{Combination of CNNs and RNNs}
Several researchers have proposed models that use a combination of RNN and CNN to handle temporal and spatial information for trajectory prediction. For instance, Deo \textit{et al.} \cite{deo2018convolutional} use an LSTM encoder to extract temporal data from nearby vehicles, which is then fed into a social pooling layer that collects interaction-related parameters between vehicles. A social tensor is created and fed into a collection of CNNs to determine the spatial correlation of vehicles. MATF \cite{zhao2019multi} introduces a fully convolutional network that resembles a U-net \cite{zhou2018unet++} for Multi-Agent Tensor Fusion (MATF) encoding and decoding. The fused vectors of each vehicle are taken from the output layer of the U-net \cite{zhou2018unet++}-like network, added to the LSTM-encoded vectors of the vehicles' dynamics, and then supplied to LSTM decoders. Schreiber \textit{et al.} \cite{schreiber2019long} use a CNN on condensed BEV images and an Encoder-Decoder LSTM to learn the temporal dynamics of the input data. TraPHic \cite{chandra2019traphic} uses a CNN-LSTM hybrid network to derive features from the state and nearby objects of the primary vehicle. Xie \textit{et al.} \cite{xie2020motion} use a "box" to find and remove outliers in the vehicle's trajectory and extract interaction-aware features by feeding them into a convolutional layer and a maximum pooling layer. Xu \textit{et al.} \cite{xu4135360vehicle} propose a model that uses a convolutional network and a graph operation layer to capture spatiotemporal features and an LSTM encoder-decoder to forecast the traffic-related future trajectories of multiple vehicles. Table XI presents a summary of CNN-based approaches for trajectory prediction, emphasizing the number of trajectories predicted and the prediction horizon measured in seconds (s). The table also highlights the strengths and weaknesses of each study including evaluation metrics used for training and testing.

\subsection{Generative Model}
Predicting multi-modal trajectories presents challenges and uncertainties due to the potential diversity of outcomes. To address this issue, some researchers have turned to generative models to create multi-modal trajectories that can capture the underlying diversity. However, in order for a multi-modal trajectory prediction model to be effective, its output distribution must meet certain requirements, including diversity, social acceptability, and controllability. Achieving an optimal distribution using only one ground truth can be difficult and may lead to less diverse and unacceptable predictions. To overcome this challenge, Generative Adversarial Networks (GANs) and Variational Auto Encoders (VAEs) have been proposed as solutions. Fig. 12 illustrates the involvement of research papers, depict in percentages, of both generative models in assisting Autonomous Vehicles (AVs) with the task of trajectory prediction. Both models contribute approximately equally to the prediction process, showcasing their shared responsibility in generating accurate trajectory predictions.
%Fig. 12 show the participation of reaserach paper of both the generative models and their responsibility in helping AVs solve the task of trajectory prediction. Both models show the approx equal participation.
 \begin{table*}[]
 \caption{Summary of Variational Auto Encoder-based methods: Related Work \& Year, No. of Predicted Trajectories for each vehicle, Prediction Horizon (PH), Advantages and Limitation of each work,
Summary of each work, and Evaluation Metric (EM) 
\newline - Predicting any number of trajectories for the work / This information is not available for work}
 \resizebox{17.5cm}{6cm}{
    \centering
    
    \begin{tabular}{|p{0.7cm}|p{1cm}|p{0.5cm}|p{4cm}|p{2.8cm}|p{4cm}|p{1.5 cm}|}
    \hline
      \textbf{Ref.}&\textbf{\# Trajec\newline tories}&\textbf{PH}&\textbf{Advantages}& \textbf{Limitation} & \textbf{Summary of Prediction Method} & \textbf{EM}\\
      \hline
        \cite{cho2019deep}\newline2019 &- &5s&Jointly reason about future vehicle trajectories as well as the degree to which each rule is satisfied.& Supposing that the control of the ego vehicle is impacted by six nearby vehicles in adjacent lanes.&Using the prior trajectory and feature representation made up of lane deviation distance and distances to nearby vehicles, CVAE uses these to learn a distribution of future trajectories.& ADE\\
         %Using LSTM and CVAE, various vehicle positional hypotheses are estimated.\\
       
         %[144]
         \cite{hu2019multi}\newline2019  & 1,3&-&The ability to connect the model to the underlying motion pattern makes it interpretable.& The prediction system is tuned for roundabout situations.&Using historical scene details and driving intentions as conditional inputs in the model structure, the prediction of joint trajectories of two cars.&MSE,NLL\\
       
         \cite{zhang2020multimodal}\newline2020 &1,10 & 5s &The spatial interactions of vehicles are modelled by dilated convolutional social pooling.& Vehicle statuses are the only components of the multimodal information input.& SSAE can automatically extract deep and high-level features from input data, which reflects the interaction between different states of vehicles.&RMSE\\
        
         \cite{sriram2020smart}\newline2020& 1,5 & 5 s &Taking into account scene semantics and inter-agent interactions, with constant-time inference regardless of the agent count.& &Using CVAE to predict data diversity for each type of agent.& ADE, FDE,NLL\\ 
        
         \cite{dulian2021multi}\newline2021  &-& 6s  &1. Instead of being limited to generating a finite number of deterministic trajectories, generate an endless number of varied motion samples; 2. show benefits of using the Minimum over N (MoN) cost function. &On-road participants' social interactions are excluded. &Constrained by an agent's previous mobility and a rasterized scene context encoded with the Capsule Network.&minADE, \newline minFDE\\
         
         %Using CVAE and CapsNet together to generate a variety of stochastic trajectories in challenging conditions
         \cite{liu2022interactive}\newline 2022& 1&5s&Considering the drivers' unknown trajectory intentions given the driving risk map.& Predict the future path for vehicles only.&Using ground truth and historical vehicle trajectories, CVAE based on GRU will produce candidate trajectories.&ADE,FDE\\
         \hline
     
    \end{tabular}}
   
    \label{tab:my_label}
\end{table*}

\begin{figure}
    \centering
     \includegraphics[width=0.5\textwidth]{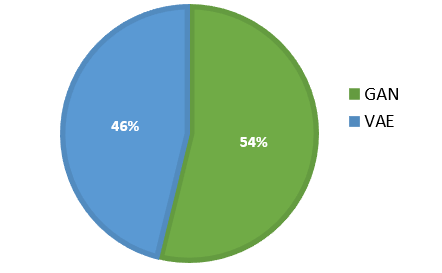}
    \caption{Participation of Research articles in trajectory prediction task using Generative models.}
    \label{fig:my_label}
\end{figure}

\paragraph{Generative Adversarial Network} 
In trajectory prediction tasks, Generative Adversarial Networks (GANs) are used to generate realistic trajectories based on the input data. The generator takes in the historical trajectory data as input and generates a future trajectory, while the discriminator evaluates the generated trajectory for realism. The generator is trained to improve the realism of the generated trajectories by fooling the discriminator into believing they are real. This methodology is shown in Fig. 13. GAN was introduced by Ian Goodfellow in 2014 \cite{wang2017generative}. When GANs are used for trajectory prediction, the discriminator assesses the accuracy of the predicted trajectory while the generator constructs it. In \cite{hegde2020vehicle}, Hegde \textit{et al.} forecast vehicle trajectories using the vehicle's coordinate information. The TS-GAN model presented by wang  \textit{et al.}  \cite{wang2020multi} utilizes a self-developed convolutional social mechanism and a recurrent social mechanism to extract vehicle spatial and temporal information from the GAN network. To create model-based multi-modal trajectories, Song \textit{et al.} \cite{song2022learning} employ vector maps and vehicle status information and apply a learning-based discriminator to extract information about vehicle interactions for providing the best trajectories. In \cite{zhao2020novel}, the GAN-VEEP model is proposed for short-term vehicle trajectory prediction, utilizing a vehicle coordinate normalization model to convert position coordinates into normalized coordinates. In \cite{li2021vehicle}, two strategies for incorporating Signal Temporal Logic (STL) rules into a GAN-style trajectory predictor are presented. In \cite{wang2021multi}, the STSF-Net framework is proposed, which utilizes a GAN for multi-modal trajectory distribution, with a generator that has an LSTM encode-decoder framework with a 3D CNN network for temporal correlations modeling and a discriminator that uses a Multi-Layer Perceptron (MLP) to identify the true trajectory. Additionally, Guo \textit{et al.} \cite{guo2023map} suggests using a map-enhanced GAN for trajectory prediction by fusing vehicle motion with high-definition (HD) maps to create contextual features. Table XII presents A summary of GAN-based approaches for trajectory prediction, focusing on the number of trajectories predicted and the prediction horizon measured in seconds (s). The table also provides the strengths and weaknesses of each study and highlights the evaluation metrics used for training and testing.

\begin{figure}
    \centering
     \includegraphics[width=0.5\textwidth, height=0.2\textwidth]{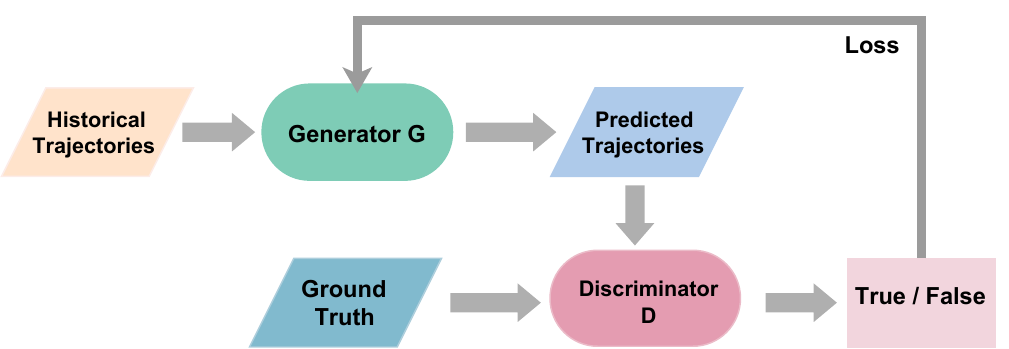}
    
    \caption{Depiction of Generative Adversarial network for trajectory prediction task.}
    \label{fig:my_label}
\end{figure}
\paragraph {Variational Auto Encoder} 
The Auto Encoder (AE) compresses data using an Encoder and decodes it with a Decoder to produce a reconstructed output with minimal reconstruction errors. However, AE has been criticized for merely "memorizing" data and having limited data generation capacity. In contrast, the Variational Autoencoder (VAE) has a generative capability that spans the entire space, and it addresses the issue of non-regularized latent space in autoencoders. VAE aims to minimize both reconstruction loss and similarity loss. Bhattacharyya \textit{et al.} \cite{bhattacharyya2019conditional} proposes the use of a Conditional Variational Autoencoder (CVAE) for structured prediction tasks.
Cho \textit{et al.} \cite{cho2019deep} suggested using CVAE and LSTM to estimate possible future positions of vehicles. To ensure compliance with traffic laws and social navigation principles, they also utilized Signal Temporal Logic (STL) to eliminate irrational scenarios. Hu \textit{et al.} \cite{hu2019multi} proposed a multi-modal trajectory prediction framework based on CVAE, but it only considered situations where two vehicles were involved. Zhang \textit{et al.} \cite{zhang2020multimodal} proposed using Stacked Sparse AutoEncoders (SSAE) to handle a high-dimensional input vector with motion and interaction data in a multi-modal scenario. Sriram \textit{et al.} \cite{sriram2020smart} presented an architecture that predicts the multi-modal trajectory of all traffic participants simultaneously using Convolutional LSTM and CVAE for scene context feature extraction and trajectory prediction, respectively. Dulian and Murray \cite{dulian2021multi} utilized CNN networks to extract spatial information from Bird's Eye View (BEV) images of an HD-Map and used a CVAE to predict future trajectories, sampling the conditional variable from a prior distribution during the testing phase. liu \textit{et al.} \cite{liu2022interactive} developed a CVAE-based model to generate potential trajectories while considering motion uncertainty and then created a driving risk map. They also developed a probability model based on the trajectory risk value and used a random selection method to produce a unifying rendering of the scene's traffic agents' interactions. Based on the findings of these studies, CVAEs can take into account various conditions such as the current state of the vehicle, surrounding traffic, road conditions, or any other relevant contextual information. These conditions can be encoded as additional inputs to the CVAE model, which then learns to generate future trajectories conditioned on these inputs. Additionally, the performance of CVAEs heavily relies on the effectiveness of the chosen conditioning inputs.
%Integrating structured maneuvers into our model framework can enhance the interpretability of our predictions. We aim to analyze the factors that impact the performance of our model.
%Despite the limited agent-scene interactions and straight road lanes in the dataset, our model did not surpass NGSIM. 
Table XIII summarizes the Variational Autoencoder-based approaches for trajectory prediction, highlighting the number of trajectories predicted and the prediction horizon measured in seconds (s). The table also provides insights into the strengths and weaknesses of each study, along with the evaluation metrics used for training and testing.
%Table 13 provides a summary of Variational Autoencoder-based approaches for trajectory prediction, along with the evaluation metrics used for training and testing. The table also highlights the strengths and weaknesses of each study.
% capable of handling both spatial and temporal dependencies

 \begin{figure}[]
    \centering
    \includegraphics[width=0.5\textwidth]{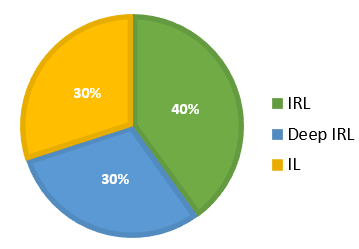}
    \caption{Participation of Research articles in trajectory prediction task using reinforcement learning-based approaches.}
    \label{fig:my_label}
\end{figure}
%\begin{table*}[]
\begin{table*}[]

    \centering
    \caption{Summary of Reinforcement Learning based methods: Related Work \& Year, Major Techniques, Scenarios, and their applications, Environment, Advantages, and Limitation of each work}
   \resizebox{!}{0.66\textwidth} {
    \begin{tabular}{|p{0.5cm}|p{3cm}|p{3cm}|p{3cm}|p{2cm}|p{4.5cm}|p{4.5cm}|}
    \hline
        \textbf{Work} &\textbf{Techniques} & \textbf{Scenarios} &\textbf{Application} & \textbf{Environ\newline ment} & \textbf{Advantages} & \textbf{Limitation} \\
         \hline
 \cite{wang2018reinforcement}\newline2018 & Deep Q-learning & Highway
segment with three lanes in one direction & Lane change, Vehicle Control. & Simulator & The ability of the vehicle agent to learn a safe and effective lane change driving strategy.& The algorithm's performance degrades in various traffic flow scenarios and road layouts.\\
 \cite{guan2018markov}\newline2018 & Markov decision process,\newline Dynamic programming, & Two-lane highway scenario. & Decision-making, overtaking and avoiding collision. & Simulator & This approach might be used in common situations like highway and park driving, eliminating the need to manually model rules and compare them to rule-based approaches.& Due to the high computational complexity, the time required increases dramatically as the dimension of the state space grows. \\
          \cite{zou2018inverse}\newline2018 & Inverse Reinforcement Learning (IRL), \newline Markov decision process, CNN& Driving curve scenario. & Left turn and right turn, driver behavior modeling & Simulator & With many unvisited states in the new curve scenario, this technique exhibits excellent generalisation efficiency. &In the demonstration trajectory state, the agent is unable to make parallel driving decisions, but it can still perform the decision-making tasks in an unvisited state.\\
           \cite{sun2018probabilistic}\newline2018& Hierarchical IRL & Ramp-merging driving scenario.&The discrete driving decisions such as yield or pass as well as the continuous trajectories. & NGSIM & Driving choice influences are explicitly modelled for both discrete and continuous driving situations. & \\
    
         \cite{gonzalez2018modeling}\newline2018& Maximum Entropy IRL,\newline Semi-Markov Decision Process (SMDP).& Merges to the right, overtakes, becomes snarled in sluggish traffic, and drives carelessly. & Two-lane highway. & Self-gathered data &The suggested method successfully models the problem of highway driving from inefficient driving demonstrations captured using an instrumented vehicle.& This does not extend the original spatiotemporal state lattice concept to handle the stop-and-go scenarios.\\
         
         \cite{xin2019accelerated}\newline2019& Accelerated IRL, \newline  maximum entropy & Ego vehicle and one front car in the scenarios. & Lane change, \newline Lane keeping & Simulator & It substitutes choosing the ideal trajectories from the candidate trajectory library produced by random sampling policies for invoking RL to generate the best trajectory during each iteration.& This model is specific to lane changes and lane keeping and does not generalise to other types of driving conditions, such as intersections and unpaved roads; 2. There is a need to generalize the proposed method under a distributed learning framework to address the stochastic problem.\\
         
         \cite{xu2019learning}\newline2019 & Inverse Optimal Control (IOC),\newline Langevin Sampling, \newline Monte Carlo \newline neural network & Greater length of the road segment, more lane curvature, and more highway entrances and exits. & Lane-following, avoiding collisions, and passing, \newline Optimal Control. & NGSIM & 1. Make more stable predictions that are better;\newline 2. Langevin Sampling and the energy-based approach can handle complex cost functions.&This does not predict the joint trajectory distribution for all moving agents; it just predicts individual trajectories.\\
         
      \cite{xu2020learning}\newline2020 &  Numerical Optimization algorithms & With a length of 65.3 km and an intended top speed of 80 km/h, the multi-lane highway is devoid of traffic signals  &Lane change and car followings\newline  motion planning & Self-collected  &Heuristic and learning-based lane incentive costs that are suggested and put into practice.& 1. Actors' interaction behaviours are not taken into account; 2. Because the large size vehicles are not included in the dataset, the shape, size, and heading of the vehicles are not adequately addressed.\\
       
       \cite{wu2020efficient} \newline2020&Maximum-entropy IRL&Settings for both interactive and non-interactive driving.& & INTERA\newline CTION &The suggested algorithm is more generalizable and converges much more quickly.&The proposed technology does not extend to generic robotic systems and was created primarily for ground vehicles and other mobile robot systems.\\
       
        \cite{deo2020trajectory}\newline2020 & Maximum-entropy IRL, \newline attention mechanism & Unknown environments.&Multimodal trajectories  &Stanford drone, NuScenes datasets&Using MaxEnt IRL to seek a strategy that can jointly predict agents' intentions and paths on a rough 2-D grid defined across the scene.& It does not consider the interaction factors of surrounding agents for future trajectory prediction.\\
          \cite{you2019advanced}\newline2019&Markov decision process, \newline IRL&Segments of highway and bend roads, each segment having five lanes.& Lane switching, maintaining lanes and speeds, accelerating and braking, overtaking, and tailgating. & Simulator & The traffic model is easily expandable to accommodate more vehicles and lanes.& unable to distinguish between different car kinds in the simulator.\\
           
      \cite{zhu2020off} \newline2020 & Maximum Entropy Inverse Reinforcement Learning,\newline CNN & Straight and flat road scenes and negative obstacles scenes. & Normal driving behavior and avoid or cross the negative obstacles. &Self collected & 1. To facilitate effective forward reinforcement learning, two new CNN are proposed. This addresses the issue of Complexity of state-space grows exponentially; 2. Different cost
functions of traversability analysis are learned to guide the trajectory planning of different behaviors & Extensive experimental research will only be performed in certain circumstances.\\
  \cite{jung2021incorporating}\newline 2021 & Deep IRL,\newline ConvLSTM & Keeping to the lane and moving through intersections whether or not there are other agents. & Traversability map. & Self-collected & 1. It takes into account several circumstances at once for an autonomous vehicle's social navigation; 2. It is independent of expensive prior environmental knowledge. & 1. The model did not take into account traffic signals and stop signs; 2. They did not take into account the  dynamic heterogeneous agents in relation to the nearby automobiles. \\

       \cite{choi2021trajgail}\newline2021& GAIL, imitation learning, POMDP & Four-way intersections & Left, right, terminate & Simulator, self-collected & 1. Enables trajectory generation that can scale to large road network environments; 2. Describing the underlying
route distribution of a traffic network in synthetic data generation problems. &  It does not take into account the effects of traffic conditions or interactions with other vehicles.

\\
        \cite{cheng2022mpnp}\newline2022&Imitation learning & Four training cities, each with four weather conditions& Distribution of diverse futures states and actions. &  Simulator &This strategy uses a camera-only methodology to simulate static and dynamic scenes as well as ego behaviour when driving in urban areas; 2. Hat jointly learns a driving policy and a world model from offline expert demonstrations alone. & They are not driving reward function from expert data.\\
      
       \cite{hu2022model}\newline2022&Maximum entropy IRL  & Cut-ins, sudden stops, and crowded hotel pickup/drop-off areas.& Multi modal& Self collected &1. This system calculates, evaluates, and scores vehicle's trajectory; 2. A straightforward design, easily understandable features, and potent real-world performance.& Evaluation metrics are not good enough to measure the performance of the model.\\
     
       \cite{phan2022driving}\newline2022& GAIL& Hill-climbing.&Trajectory planning & Self-gathered & Underline the value of closed-loop evaluation and training using interactive agents.&1. The ability of the policy may be enhanced by directly teaching the planning agent alongside already taught interactive agents; 2. Generalization to innovative routes is not addressed.\\
     
       \cite{bronstein2022hierarchical}\newline2022& Deep IRL, maximum entropy IRL & Mixed-driving scenario.& Multi modal & Nuscene & 1. This uses scene rasterization to provide the neural network's input scene data; 2. The trajectory generator module now includes a correction factor, by disfavoring trajectories with little difference, can produce more varied trajectories.& Scene rasterization could be hampered by ineffective coding, a lengthy learning curve, and a loss of connection information from occlusion.\\
       \hline
    \end{tabular}}
   
    \label{tab:my_label}
\end{table*}

\section{Reinforcement learning-based methods} Recent years have seen the rapid growth of Reinforcement Learning (RL), which offers a new method for comprehending high-dimensional complex policies \cite{hjaltason2019predicting}. It offers innovative solutions for Autonomous Vehicles (AVs)' challenges involving trajectory prediction \cite{kiran2021deep, fernando2020deep}. The Markov Decision Process (MDP) is typically utilised when RL is used to AVs trajectory prediction to maximise the projected cumulative reward. 
%An MDP model contains: 
%\begin{enumerate}
%   \item States: A set of possible finite states S %=$s_1,s_2,s_s,....s_n$.

 %\item The transition function specifies the likelihood of transitioning to state s(t+1) from state s at time step t when action a is taken.

 %\item Model:  T($s_t,a_t$,s(t+1)), The transition function specifies the likelihood of transitioning to state s(t+1) from state s at time step t when action a is taken.
% Probability to go to state s' when you do the action while you were on state s, is also called the transition model.
% \item Action: A = $a_1,a_2,a_3,....a_n$, A finite state-action space encompasses all the feasible actions that an agent can execute at each time step within a given state.

 %\item Reward: R($s_t,a_t$,s(t+1)), The reward function provides information about the expense associated with taking action a in state s at time step t.

 %\item Policy: A policy $\pi$ refers to a function that maps a state s to an action a. A policy $\pi$ can be either stochastic, represented as $\pi$: S → Prob(A), or deterministic, represented as $\pi$: S → A.  
%\end{enumerate}

% (s)->a , our goal, is a map that tells the optimal action for every state
 % Optimal policy: $π^*$, is a policy that maximizes your expected reward 
%We aim to determine the optimal course of action for each environmental condition through a policy.  
RL techniques are utilized to estimate the underlying cost function or directly identify the optimal policy for trajectory prediction. In either approach, it is assumed that the observed agent always seeks to reach its objective by utilizing the optimal policy based on a specific cost function. Fig. 15 illustrates the application of RL methods in AVs. Within the framework of MDP, RL-based methods can be categorized into Inverse Reinforcement Learning (IRL) methods,  Imitation Learning (IL) methods, and Deep IRL methods, as explained in the following sections. Fig. 14 illustrates the distribution of research articles, expressed as percentages, for different variants of RL and their involvement in addressing the trajectory prediction task in AVs. 
%Fig. 15 depicts the distribution of research articles in percentages of different variants of reinforcement learning and their involvement in addressing the trajectory prediction task in autonomous vehicles (AVs).

% The policy that will provide us with the best course of action for each condition of our environment is what we are trying to find. Reinforcement learning techniques are used to derive trajectory prediction by estimating the underlying cost function or determining the optimal policy directly. In both cases, it is assumed that the observed agent always attempts to reach its goal using the optimal policy based on a specific cost function.
% The two main methods for calculating the trajectory prediction are either estimating the underlying cost function or directly calculating the best course of action. are Inverse Reinforcement Learning (IRL) and Imitational Learning(IL). 

\subsection{Inverse Reinforcement Learning}
The main idea behind Inverse Reinforcement Learning (IRL) is to learn the reward function that explains the observed behavior of the agents. Instead of directly imitating the observed trajectories, IRL aims to understand the underlying motivations or objectives that drive those trajectories. By inferring the reward function, the algorithm can generalize beyond the observed trajectories and make predictions about future trajectories.
Manually specifying the weight of the reward function is inappropriate due to the complex nature of driver behavior, according to Wang \textit{et al.} \cite{wang2018reinforcement} and Guan \textit{et al.} \cite{guan2018markov}. To address this issue, IRL learns the optimal driving policy by inferring the reward function based on expert demonstrations (trajectories), as depicted in Fig. 16. 
%The illustration of IRL method is shown in Fig. 16.
Liting \textit{et al.} \cite{gonzalez2018modeling} utilize a spatiotemporal state lattice to describe driver behavior based on expert demonstrations. The driving maneuvers create a distribution for upcoming trajectories \cite{sun2018probabilistic}. Interaction-related elements are considered to achieve probabilistic prediction for AVs. DriveIRL, presented by Tung \textit{et al.} \cite{phan2022driving}, is the first learning-based planner that uses IRL to control a vehicle in congested urban traffic. They build an architecture divided into ego trajectory generation, checking, and scoring, using simple and reliable techniques to solve the very complex problem of ego trajectory generation.

A significant challenge in IRL is that an optimal policy may be ambiguous, since it can result from multiple
reward functions \cite{abbeel2004apprenticeship}. Because of this, a modified algorithm
called Maximum Entropy IRL (MaxEntIRL) was developed
by Ziebart \textit{et al.}  \cite{ziebart2008maximum}. The MaxEntIRL algorithm aims to resolve the ambiguity in IRL by maximizing the entropy across the distributions of potential state-action pairs for a learned policy. %Various reward functions can cause confusion in IRL for an optimal policy \cite{abbeel2004apprenticeship}. To address this problem, Ziebart \textit{et al.} \cite{ziebart2008maximum} propose a modified method known as maximum entropy IRL (MaxEntIRL) that maximizes entropy over distributions of potential state-action combinations for a learned policy, eliminating ambiguity. 
 Some MaxEnt-IRL techniques use sampled trajectories to carry out prediction tasks. Xu \textit{et al.} \cite{xu2020learning} sample candidate trajectories with the lowest cost that will be selected as the anticipated trajectory. Wu \textit{et al.} \cite{wu2020efficient} propose a method for learning reward functions in the continuous domain by estimating the partition function using the speed profile sampler. State sequences from the MaxEnt policy are sampled in \cite{deo2020trajectory}  and provided to an attention-based trajectory generator to produce valuable future trajectories. To estimate the best policy while reducing computing costs, Xin \textit{et al.} \cite{xin2019accelerated} utilize randomly pre-sampled policies in sub-spaces. Yifei \textit{et al.} \cite{xu2019learning} propose an Inverse Optimum Control (IOC) method utilizing Langevin Sampling to determine the cost function of other vehicles in an energy-based generative model. 
In summary, while IRL has the potential to provide deeper insights and more flexible trajectory predictions, the requirement for expert demonstrations and the challenges associated with their quality and computational complexity should be carefully considered in practical applications.
\begin{figure}[hbt!]
    \centering
    \includegraphics[width=0.5\textwidth,height=0.3\textwidth]{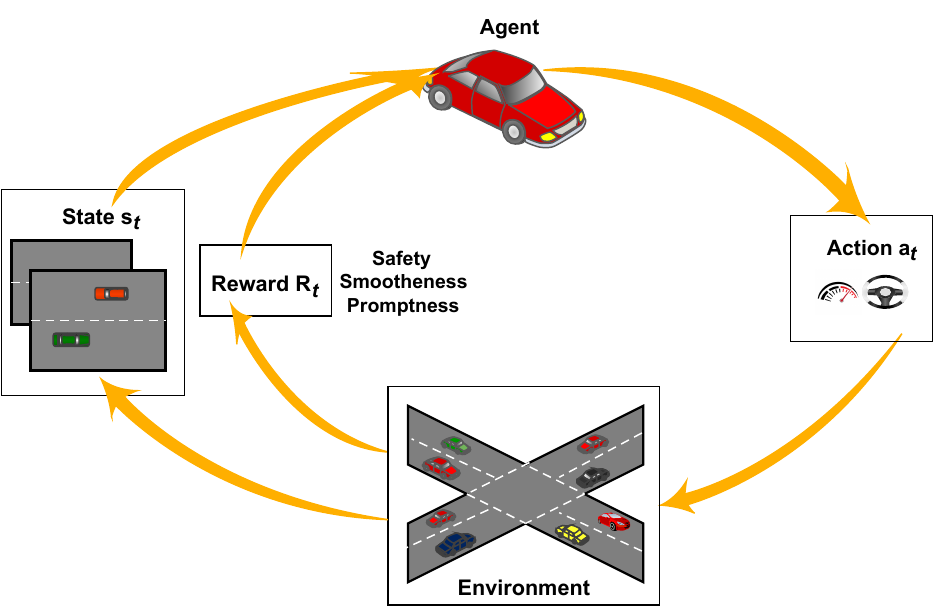}
    \caption{Depiction of the Reinforcement-learning based method.}
    \label{fig:my_label}
\end{figure}

\subsection{Deep Inverse Reinforcement Learning }
Deep Inverse Reinforcement Learning (Deep IRL) is an extension of Inverse Reinforcement Learning (IRL) that incorporates Deep Neural Networks (DNNs) to learn the reward function from expert demonstrations. The deep IRL framework is introduced in \cite{wulfmeier2015maximum} to approximate complex and nonlinear reward functions. To approximate rewards, this article uses a fully Convolutional Neural Network (CCN). You  \textit{et al.} \cite{you2019advanced} consider driving behavior and road geometry, constructing the MDP first using RL, learning the best driving strategy using IRL, and approximating the reward function using DNN. In \cite{zou2018inverse}, driving behavior is represented by Deep IRL utilizing camera images, while CNN extracts the corresponding state information. Zhu  \textit{et al.} \cite{zhu2020off} encode the vehicle's kinematics using RL ConvNet and State Visiting Frequency (SVF) ConvNet by back-propagating the loss gradient \cite{wulfmeier2016watch} between expert SVF from expert demonstration and policy SVF from LiDAR data. Jung \textit{et al.} \cite{jung2021incorporating} using neural LSTM to extract the feature map from the LiDAR and trajectory data, which will then be merged into the output reward map to forecast the traversability map. In \cite{cheng2022mpnp}, a fused dilated convolution module is proposed to improve the extraction of raster features. Subsequently, a reward update policy with inferred goals is enhanced by learning the state rewards of goals and pathways individually instead of the original complex rewards, which can reduce the need for preset goal states. In summary, Deep IRL offers the potential for more powerful and adaptive trajectory prediction models by leveraging deep neural networks. However, challenges related to data requirements, computational complexity, interpretability, and overfitting need to be carefully addressed for successful application in trajectory prediction for autonomous driving.
\begin{figure}[hbt!]
    \centering
    \includegraphics[width=0.5\textwidth,height=0.3\textwidth]{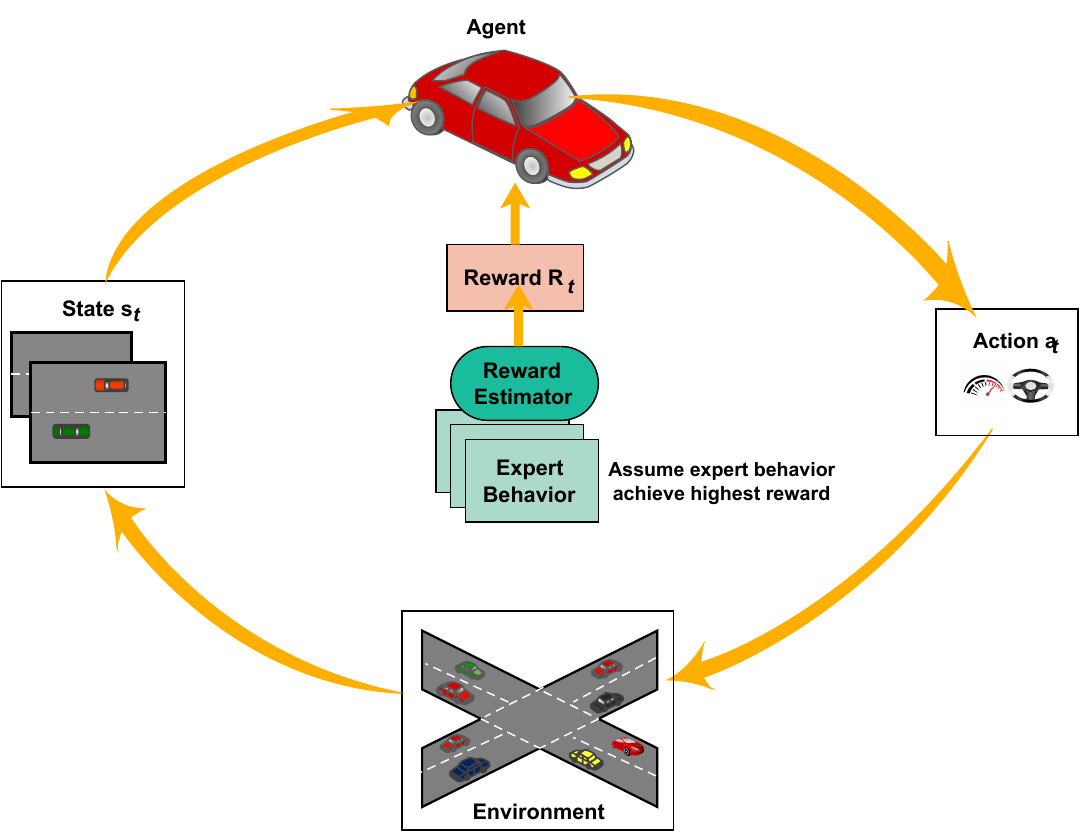}
    \caption{Depiction of the Inverse Reinforcement-learning based method.}
    \label{fig:my_label}
\end{figure}
\subsection{Imitation Learning}
One disadvantage of Inverse Reinforcement Learning (IRL) algorithms is their difficulty in training with scenarios where there are few rewards or no direct reward function. To address this issue, Imitational Learning (IL) has been suggested as a solution. IL aims to quickly determine a policy based on an expert's observation without requiring a cost function. One of the pioneering methods in imitation learning for autonomous driving is ALVINN, developed by Pomerleau \cite{pomerleau1988alvinn}. Another notable approach by Anthony \textit{et al.} \cite{hu2022model} introduces a novel model-based architecture that leverages 3D geometry as an inductive bias. This method is trained solely on an offline dataset of expert driving data, eliminating the need for reward signals or online interaction. This approach shows great promise for real-world applications.
%The first IL method for autonomous driving was ALVINN by Pomerleau \cite{pomerleau1988alvinn}. Anthony \textit{et al.}\cite{hu2022model} propose a novel model-based IL architecture that utilizes 3D geometry as an inductive bias and is trained exclusively using an offline corpus of expert driving data. This approach does not require access to rewards or interaction with an online environment, making it highly promising for real-world applications.

The author utilizes behavior cloning (BC) \cite{devi2020behaviour}, a technique that focuses solely on imitating the expert's policy. BC is straightforward and effective, but it struggles with unknown states, requiring a substantial amount of data. To address this limitation and produce a policy instead of a cost function, 
Generative Adversarial Imitation Learning (GAIL), proposed by Ho \textit{et al.} \cite{ho2016generative}, uses the Generative Adversarial Network (GAN) approach for imitation learning in RL. GAIL extracts policies directly from data rather than relying on expert demonstrations. GAIL, similar to GAN, is based on the fundamental concept of a generator and discriminator. The generator in GAIL produces trajectories that resemble those of an expert as closely as possible, while the discriminator determines whether the generated trajectories are from the expert or not, as shown in Fig. 17. To address GAIL's limitations in only using the current state to model the subsequent state, Choi \textit{et al.} in \cite{choi2021trajgail} propose a method that incorporates a Partially Observable Markov Decision Process (POMDP) within the GAIL framework and uses the reward function from the discriminator to train the model. The high-dimensional solution space of a POMDP makes complex scenario modelling computationally expensive. Additionally, the ambiguity of state observations makes it difficult to differentiate state-action pairs. However, notable advancements have been made in online POMDPs, as demonstrated in \cite{somani2013despot} and \cite{silver2010monte}.
Bronstein \textit{et al.} \cite{bronstein2022hierarchical} modify the default model-based GAIL with a hierarchical model to enable generalization to any goal pathways and evaluate performance with simulated interactive agents in a closed-loop evaluation framework. Kuefler \textit{et al.} \cite{kuefler2017imitating} employ GAIL to model human driving behavior on highways and propose an RNN integrated into the GAIL architecture. Bansal \textit{et al.} \cite{bansal2018chauffeurnet}'s ChaffeurNet utilizes IL to train a robust policy while penalizing implausible events and introduces an explicit loss to prevent the algorithm from solely imitating such undesirable behavior.
In summary, IL and GAIL are promising approaches to address the challenges of training RL algorithms in scenarios with limited rewards or no direct reward function. Their success in modeling human driving behavior and generating realistic predictions opens up possibilities for their application in other real-world scenarios.
Table XIV provides a summary of Reinforcement learning-based approaches for trajectory prediction and also highlights the strengths and weaknesses of each study.
\begin{figure}[]
    \centering
    \includegraphics[width=0.5\textwidth]{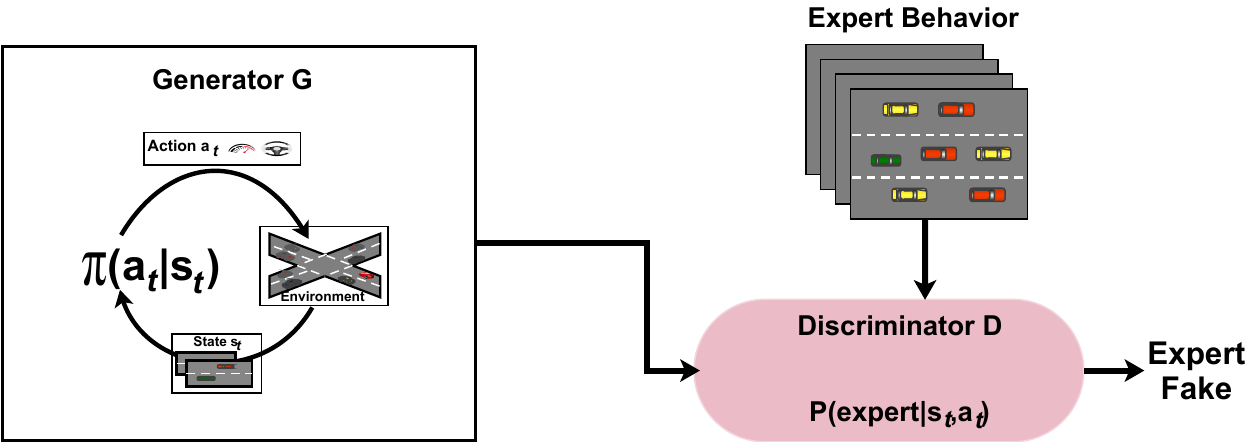}
    \caption{Depiction of the Generative Adversarial Imitation learning based method.}
    \label{fig:my_label}
\end{figure}

\section{ Training and Evaluation}
Various standard datasets are used to test prediction algorithms, and appropriate metrics are used to assess their performance.

\begin{table*}[]
 \caption{Datasets for AVs which are utilized in trajectory prediction} \resizebox{17.5cm}{10.4cm}{
    \centering
    \begin{tabular}{|p{2cm}|p{0.7cm}p{0.8cm}p{0.5cm}p{0.7cm}|p{5.5cm}|p{2.5cm}|p{2.7cm}|}
    \hline
         \textbf{Dataset} &\textbf{LiDAR} & \textbf{Camera} & \textbf{Radar} & \textbf{Drone} & \textbf{Scene Description} & \textbf{Applications} & \textbf{Used by} \\
         \hline
         Nuscenes \cite{caesar2020nuscenes} \newline 2020 &\checkmark &\checkmark&\checkmark && Each scene in nuScenes is 20 seconds long, completely annotated, and has 3D bounding boxes for 8 attributes and 23 classes.& Object identification, tracking, and segmentation &\cite{kim2021lapred},\cite{messaoud2021trajectory},\cite{phan2020covernet},\newline \cite{meng2023trajectory},\cite{li2021vehicle},\cite{guo2023map},\cite{dulian2021multi}
         
         \\ 
           
         Lyft \cite{houston2021one} \newline 2021 &\checkmark &\checkmark& &&During the course of four months, this was gathered by a fleet of 20 autonomous vehicles traveling along a predetermined path in Palo Alto, California. There are 170,000 scenes total, and each scene lasts for 25 seconds.& Motion forecasting and  planning and simulation.&\cite{wang2023safety},\cite{chandra2020forecasting},\cite{zhang2022trajectory}\\

         Waymo Open datasets \cite{ettinger2021large} \newline 2021 & \checkmark &\checkmark & &&Presently contain 1,950 segments of 20 seconds each, accumulated over 390,000 frames in various environments.& Object detection &\cite{ngiam2022scene},\cite{gu2021densetnt} \\
         
         Argoverse \cite{wilson2023argoverse}  \newline 2023 &\checkmark &\checkmark &&&This dataset consists of 360-degree pictures captured by seven cameras with overlapping fields of view, 3D point clouds generated by long-range LiDAR, 6-DOF posture annotations, and 3D track annotations. &Tracking and predicting movements of 3D objects. &\cite{kawasaki2020multimodal},\cite{tang2019multiple},\cite{strohbeck2020multiple},\cite{kim2021lapred}, \cite{kim2022diverse},\cite{liu2021multimodal},\cite{huang2022multi},\cite{ngiam2022scene}, \cite{wang2023lane},\cite{chandra2020forecasting},\cite{liu2022multi},\cite{zhang2022ai},
         \cite{gao2020vectornet},\cite{liang2020learning},\cite{zhao2021tnt},\cite{gu2021densetnt}, \cite{zeng2021lanercnn},\cite{guo2023map},\cite{sriram2020smart}
         
         \\
         INTERACTION \cite{zhan2019interaction} \newline2019 & & \checkmark &&\checkmark  & The interactive driving scenarios cover a wide range of situations, such as merging and lane changes in urban, highway, and ramp environments, roundabouts with yield and stop signs, signalised intersections, etc.& Decision-making, planning, and imitation learning &\cite{wu2021hsta},\cite{quintanar2021predicting},\cite{mo2022multi}, \cite{mo2020recog},\cite{zhao2021tnt}
         \\
         
         HighD \cite{krajewski2018highd}  \newline 2018 & && &\checkmark  & In six different places, traffic was recorded, and more than 110 500 vehicles were present. The trajectory of each vehicle, including its type, size, and manoeuvres, is automatically retrieved. & Traffic pattern analysis or driver model parameterization & \cite{li2022vehicle},\cite{yan2020trajectory},\cite{wu2021hsta},\cite{kim2020multi}, \cite{messaoud2020attention},\cite{quintanar2021predicting},\cite{chen2022vehicle}, \cite{gao2023dual},\cite{liu2022interactive}
         
         \\
         Apolloscape \cite{huang2019apolloscape}  \newline 2018 & \checkmark &\checkmark& & & Amazing collection of more than 140,000 video frames from several sites in China were collected during varied weather situations &Scene parsing, lane segmentation, object detection and tracking in three dimensions, and self-localization  &\cite{li2019grip++},\cite{chandra2020forecasting},\cite{xu2023mvhgn},\cite{zhang2022ai}, \cite{xu4135360vehicle}\\
         
         Kitti \cite{geiger2013vision}\newline 2013 & \checkmark &\checkmark& &  & This dataset are gathered while traveling through rural areas, on highways, and around Karlsruhe, a medium-sized city. Per shot, up to 15 cars and 30 pedestrians can be seen. & 3D object detection and 3D tracking  & \cite{marchetti2020mantra} \\
         
         NGSIM \cite{coifman2017critical} \newline 2017 & \checkmark && &&Road segment lengths of 640 and 500 meters were recorded for US Route 101 and Interstate 80 & Trajectory prediction &\cite{xin2018intention},\cite{dai2019modeling},\cite{ding2019predicting},\cite{deo2018multi},
         \cite{tang2019multiple},\cite{xie2021congestion},\cite{katariya2022deeptrack},\cite{hao2020attention},
         
         \cite{fu2021trajectory},\cite{lin2021vehicle},\cite{yu2021dynamic},\cite{meng2021intelligent},
\cite{wu2021hsta},\cite{yang2022lane},\cite{messaoud2020attention},\cite{hu2022trajectory},
\cite{hasan2023mals},\cite{zhao2021spatial},\cite{chen2022vehicle},\cite{hou2022structural},
\cite{gao2023dual},\cite{deo2018convolutional},\cite{li2019grip},\cite{li2019grip++},
\cite{sheng2022graph},\cite{xu2022group},\cite{zhao2020gisnet},\cite{jeon2020scale},
\cite{chandra2020forecasting},\cite{ding2021ra},\cite{mo2022multi},\cite{zhang2022ai},
\cite{xie2020motion},\cite{zhao2019multi},\cite{xu4135360vehicle},
\cite{wang2020multi},\cite{wang2021multi},\cite{cho2019deep},\cite{zhang2020multimodal}
         \\
         \hline
    
    \end{tabular}}
   
    \label{tab:my_label}
\end{table*}

\begin{table*}[]
   \caption{Evaluation metrics applied in the literature to evaluate the performance of trajectory prediction} \resizebox{17.5cm}{10.4cm}{
    \centering
    \begin{tabular}{|p{2cm}|p{3.5cm}|p{12cm}|}
    \hline
    \textbf{Evaluation Metrics} & \textbf{Formula }& \textbf{Description}\\
    \hline
         % Accuracy Score&-&Determine the accuracy of the prediction: the higher the scoring value, the more accurate the prediction. The measure is the sum of the number of data samples that were successfully categorised divided by the total number of data samples. \\
         % Average Accuracy & &\\
         % The proportions of the prediction error/ displacement error i.e. $e_t$ are averaged by MAE. RMSE determines the average of $e_t$'s square root.: RMSE calculates the square root of the average of $e_t^2$, they are calculated across a specified window of time (t) on the prediction horizon and n is the number of samples in prediction horizon. $e_t$ calculated an error between the actual and anticipated data values. It is used for the regression-based trajectory creation method.
          Mean Absolute Error (MAE)/Root Mean Square Error (RMSE) & RMSE= $\sqrt{\frac{1}{n}\sum_{t=1}^n e_t^2}$ \newline MAE=$\frac{1}{n}\sum_{t=1}^n |e_t|$ &The MAE calculates the average of the prediction error or displacement error, represented by $e_t$. On the other hand,  RMSE computes the square root of the average of $e_t^2$, considering a defined time window (t) on the prediction horizon, and $n$ represents the number of samples in the prediction horizon. The value of $e_t$ represents the difference between the actual and predicted data values and is utilized for the creation of a regression-based trajectory method.
         
         \\

%distance between the predicted final location $\hat{Y_{end}}$ and true final location $Y_{end}$ at the end of the prediction horizon. Even though it does not take into account forecast errors that happened in previous time steps inside the prediction horizon. Minimum FDE (mFDE) is typically used for multimodal prediction to denote the lowest value of FDE over K predictions.

Final Displacement Error (FDE) & FDE = $| \hat{Y}_{end} Y_{end}|$ & 
The FDE measures the discrepancy between the predicted final location $\hat{Y}_{end}$ and the actual final location $Y_{end}$ at the conclusion of the prediction horizon. It solely considers the forecast errors that occur in the last time step of the prediction horizon and disregards any previous errors. When dealing with multimodal predictions, the Minimum FDE (mFDE) is used to refer to the smallest FDE value among K predictions.

\\
%distance between the predicted location $\hat{Y_{t}}$ and true location $Y_{t}$ in the prediction horizon T. Minimum ADE (mADE) is typically used to denote the minimum value of ADE over K predictions in multimodal prediction.

Average Displacement Error (ADE) & ADE = ${\frac{1}{T}\sum_{t=1}^T ||\hat{Y_{t}}-Y_{t}||_2}$  & 
The ADE is the distance between the predicted location $\hat{Y_{t}}$ and the actual location $Y_{t}$ throughout the prediction horizon, which is defined as $T$. When dealing with multimodal predictions, the Minimum ADE (mADE) is used to represent the smallest ADE value among K predictions.

% truth coordinates and predicted onesIt is the average l2 distance between each point that makes up the anticipated trajectory $Y_j^^$ and the matching ground truth $Y_j$.

 % The ratio of predicted trajectories not within 2.0 meters of the ground truth is determined by calculating the Euclidean distance of the final position. In cases where the prediction results are multi-modal, K likely future trajectories are assumed, and ADE, FDE, and MR will be evaluated based on the optimal future trajectory. These evaluations will be denoted as $ADE_K$, $FDE_K$, and $MR_K$, respectively.

\\
%p and q are modelled trajectory distribution and ground truth data distribution respectively.
 Negative Log Likelihood (NLL) & $H(q,p)= E_{x\sim p}-log((p(x))$ & The trajectory distribution of the model is represented by p, whereas q denotes the distribution of the ground truth data.
 
 \\
  &&\\
WSADE, WSFDE \cite{li2019grip++}, \cite{xu2023mvhgn}, 
 \cite{zhang2022ai}&  $WSADE= D_V.ADE_V+ D_P. ADE_P + D_C.ADE_C$ \newline\newline $WSFDE= D_V.FDE_V+ D_P. FDE_P + D_C.FDE_C$ &Given the dissimilar characteristics of vehicle, cyclist, and pedestrian trajectories, the variables $D_V$, $D_P$, and $D_C $ are inversely proportional to the mean speeds of vehicles, pedestrians, and cyclists, respectively, in the dataset. The evaluation metrics used for vehicles include $ADE_V$ and $FDE_V$, for pedestrians $ADE_P$ and $FDE_P$, and for cyclists $ADE_C$ and $FDE_C$.

% As the attributes of vehicle, cyclist, and pedestrian trajectories differ, the variables $D_V$, $D_P$, and $D_C $ are inversely related to the average speeds of vehicles, pedestrians, and cyclists in the dataset. $ADE_V$,$FDE_V$ for the vehicles, $ADE_P$,$FDE_P$ , for pedestrians, and $ADE_C$, $FDE_C$ for cyclists.
\\

 Miss Rate (MR) &&To compute the ratio of predicted trajectories that differ by more than 2.0 meters from the ground truth, the Euclidean distance between their final positions is calculated. In scenarios where the prediction outcomes are multimodal, K feasible future trajectories are taken into account, and the optimal future trajectory is used to evaluate ADE, FDE, and MR, which will be indicated as $ADE_K$, $FDE_K$, and $MR_K$, respectively.
 
 \\
Computation Time &
Based on the hardware configuration or specifications. &Computation time is a critical factor in determining the on-board performance of the method. Although autonomous vehicles have limited computing capabilities, trajectory prediction models are often complex and require substantial computational resources. As the level of autonomous driving increases, it becomes imperative for each module to execute computations at a faster rate to reduce any potential delays. Therefore, the model's real-time performance or computational cost is of utmost importance.

% For the method's on-board performance, computation time is crucial. Despite the low computer capacity of autonomous vehicles, trajectory prediction models are typically complicated and call for a significant amount of computational resources. Higher levels of autonomous driving require each module's computation to run relatively quickly in order to minimise the delay. As a result, the model's real-time performance or computational cost is crucial.
 \\
Prediction Horizon & Based on the specific use case or application.& The time steps into the future that the model can predict are referred to as the prediction horizon. In a dynamic and, at times, unpredictable driving environment, the accuracy of trajectory prediction models typically declines as the prediction horizon increases. Nonetheless, to fulfill the requirements of the planning and control system, the forecast time should not be excessively brief, and it should be consistent with other modules, even in a dynamic and stochastic environment.

% The future time steps the model can forecast are referred to as the prediction horizon. In a dynamical and maybe even stochastic driving environment, accuracy will typically decrease as the prediction horizon lengthens. However, in order for the trajectory prediction results with a given amount of time to satisfy the needs of the planning and control system, the forecast time should not be too short and should be in agreement with other modules.\\
% Minimum of K Metric: 

\\
\hline
    \end{tabular}}
 
    \label{tab:my_label}
\end{table*}

\paragraph{Datasets} 
To evaluate the accuracy of a trajectory prediction model, the projected trajectory and ground truth trajectory are usually compared. These trajectories are obtained from multiple datasets that are collected using sensors such as LiDAR, cameras, radar etc. The vehicle movements in these datasets are either automatically generated or manually annotated.
Modern benchmarks have made significant progress in the AVs prediction field, overcoming the limitations of older datasets which were constrained in terms of environments and agent categories. The NGSIM-180 \cite{coifman2017critical} and highD \cite{krajewski2018highd} dataset are examples of such benchmarks that utilized drones and surveillance cameras to capture cars on highways. These datasets focused on a single type of agent with a limited set of possible actions, which included moving left or right and maintaining a straight path.
The KITTI \cite{geiger2013vision} dataset, introduced by Geiger \textit{et al.} in 2013, was among the earliest multimodality datasets that included LiDAR point clouds in addition to camera frames for input scenes. This development has generated a recent interest in object detection using 3D bounding boxes \cite{chen2017multi}. Moreover, KITTI \cite{geiger2013vision} offers annotations for both cars and pedestrians.

As the depth of AI models increases, more images are required for efficient generalization. Recent datasets such as Lyft \cite{houston2021one}, Waymo \cite{ettinger2021large}, nuScenes \cite{caesar2020nuscenes}, and Argoverse \cite{wilson2023argoverse}  have significantly increased the number of annotated frames, thereby facilitating the training of deep models. These datasets not only include camera and LiDAR data but also provide High Definition (HD) maps that capture the road's topology \cite{seif2016autonomous}. The addition of HD maps has made it possible to investigate global navigation abilities, thus enabling the training of models for longer prediction horizons. Unlike previous datasets, the aforementioned datasets cover more classes, record ego-vehicle odometry data, encompass various cities, different weather and lighting conditions (including rain and night), and provide labels for other agents such as traffic lights and road rules. However, they still lack labels related to intention prediction.

To summarize, modern datasets have effectively addressed many of the challenges associated with prediction by providing a vast amount of diverse, multi-agent, multi-modal data. This data can be used to train models capable of predicting the behavior of various interacting agents in diverse weather conditions. Furthermore, these datasets offer annotations that are useful for high-level comprehension of the driving scene, including information on location, action, and events. Table XV presents an overview of the popular datasets commonly utilized in trajectory prediction tasks. The table includes information about the sensors used, scene descriptions, applications, and the research articles that have utilized these datasets.
%Table 16 provides a summary of the common datasets used in trajectory prediction tasks. The table includes the sensors, scene description , applications, and used by research articles.  
The majority of the techniques described in this paper employ trajectories as input, however, some also make use of vehicle states or map data.

\paragraph{Evaluation Metrics} 
Evaluation Metrics (EMs) are crucial for assessing the effectiveness of vehicle trajectory prediction models. One common metric used for evaluating model output is the \emph{Average Displacement Error} (ADE), which measures the mean $l_2$ distance between the predicted trajectory's locations and the corresponding ground truth. Another metric, the \emph{Final Displacement Error} (FDE), calculates the same distance but only for the final predicted location and its ground truth at the prediction horizon.
%related to generative models include various versions of Negative Log Likelihood (NLL),
Probabilistic generative models that produce multi-modal predictions require additional metrics. The \emph{Best of N} metric calculates ADE and FDE for the best \emph{N} samples out of all generated trajectories. When \emph{N}  equals 1, the method is called minADE and minFDE, respectively, and only the generated trajectory which is closest to the ground truth is selected. Other metrics for the multi-modal distribution include various versions of \emph{Negative Log Likelihood} (NLL), which compares the distribution of generated trajectories against the ground truth. To evaluate the performance of the model on the ApolloScape \cite{huang2019apolloscape} trajectory dataset in the literature, two metrics, the \emph{Weighted Sum of Average Displacement Errors} (WSADE) and the \emph{Weighted Sum of Final Displacement Errors} (WSFDE), are frequently used. Table XVI highlights the commonly used Evaluation metrics for trajectory prediction tasks with their formula and description.

\begin{table*}[t]
 \caption{Comparison of prediction models}
  \resizebox{17.5cm}{2cm}{
    \centering
    \begin{tabular}{|p{2cm}|p{2.0cm}|p{2cm}|p{1.5cm}|p{1.5cm}|p{1.5cm}|p{1.5cm}|p{1.5cm}|p{1.5cm}|}
    \hline
     & Prediction Horizon & Explainability & Holism & Complexity & Data Dependency & Adaptivity & Computat-\newline ional Time & Accuracy\\
     \hline
         \textbf{Deep learning-based methods} & Long-term(5s-8s) &Low&High &High &High &Medium& High & High\\
         \textbf{Reinforcement learning-based methods}& Long-term(5s-8s) & Medium(IRL), Low(IL)& High &High & Very High&High, if reward function is learned (IRL) & High & Medium\\
         \hline
    \end{tabular}}
    \raggedright \tiny{Holism: Adopting object interaction and semantic data, 
Adaptivity: Robust application in unknowable situations}
    \label{tab:my_label}
\end{table*}

\section{Discussion}
In this section, a fair evaluation of the proposed models is presented through a comparison of representative models. The selected criteria encompass different factors that pertain to the task of trajectory prediction, as well as the overall prerequisites for utilizing the models in the field. Nonetheless, the comparison reveals prevailing patterns and provides an understanding of particular characteristics and scenarios of use. Deep Learning-based models and Reinforcement learning-based approaches shall be compared. The comparison results are summarized in Table XVII.

Deep learning-based models have demonstrated their ability to produce accurate predictions over an extended period, as they can conduct long-term predictions of up to 8 seconds. However, these models are typically comprised of neural networks and are therefore considered black-box models, which reduces their explainability and could pose challenges in terms of validation and approval. Despite this, these models have the advantage of being holistic since they can integrate various features from multiple sources, including object interaction and semantic data, into the neural network. However, to achieve good prediction performance, it is crucial to carefully select valid features. The use of spatial features and corresponding representation enables the consideration of the interaction between agents, which makes interaction awareness possible.
%both discrete maneuvers and 
Deep learning-based models have the capability to describe complex processes at varying levels of abstraction, with the ability to output trajectories as prediction results. However, these models require valid training data that reflects the specific field of application to enable comprehensive and robust predictions. As a result, these models are highly data-dependent. Additionally, the adaptivity of these models is limited to scenarios that fall within the data the model has been trained on. Due to their holistic approach, Deep learning-based models are typically associated with high computational costs, which are strongly influenced by the size of the neural networks used. Nevertheless, in the current state of the art, Deep learning-based models offer the highest prediction accuracy. 

Reinforcement learning-based methods are also capable of conducting long-term predictions. However, the degree of explainability varies depending on the specific approach used. Indirect models generate a cost function that is mapped to state-action tuples, which can be used to interpret the proposed output of a policy. Nevertheless, it is challenging to explain how the cost function is determined from an expert's demonstration. Direct models that output a policy do not explicitly derive a cost function from demonstration, making them less explainable. These models can directly consider the interaction between multiple objects as an input feature. Additionally, a wide range of features, including semantical information from road maps, can be used as input, making these models holistic.

Reinforcement learning-based models have the ability to describe complex maneuvers by utilizing the underlying policy. However, the model's output typically consists of discrete maneuvers because policies comprise state-action tuples that objects may execute. Although explicit trajectories can be derived from subsequent modules, such as a Recurrent Neural Network (RNN) demonstrated in \cite{bansal2018chauffeurnet}, these models heavily rely on diverse data, including demonstrations, for training. Extracting comprehensive cost functions or robust policies is particularly challenging as it strongly relies on expert behavior observations, making it difficult to train correctly. Reinforcement learning-based models are designed to reason about an object's motion, allowing them to adapt well to unknown scenarios. However, similar to Deep learning-based models, holistic models based on the reinforcement learning approach have high computational costs. Moreover, the complexity of learning a robust policy negatively affects prediction accuracy.
\section{Challenges and Future Research Directions}
Based on the above survey, this section highlights the research challenges and future research directions in the domain of Autonomous Vehicles (AVs) trajectory predictions.  
\subsection{Challenges}
%Trajectory prediction plays a crucial role in AV systems by allowing them to anticipate the future movements of various traffic participants, including vehicles, pedestrians, and cyclists in their surroundings. Nonetheless, the domain of AVs presents unique challenges that make trajectory prediction exceptionally challenging. Some of these challenges include:

%Trajectory prediction plays a crucial role in autonomous vehicle (AV) systems by allowing them to anticipate the future movements of various traffic agents like vehicles, pedestrians, and cyclists in their surroundings. However, the domain of AVs presents several unique challenges that make trajectory prediction particularly challenging:

%However, there are several challenges specific to the domain of AVs that make trajectory prediction exceptionally challenging:

Trajectory prediction is a critical component of AV systems, as it enables them to anticipate the future motion of traffic agents such as vehicles, pedestrians, and cyclists in their environment. However, there are several challenges specific to the domain of AVs that make trajectory prediction exceptionally challenging:

%However, there are several challenges specific to the domain of AVs that make trajectory prediction particularly difficult:
\begin{enumerate}
    \item \emph {Uncertainty:} The future trajectory of traffic agents is inherently uncertain, and it is impossible to predict it with 100 percent accuracy. Various factors such as noise in sensor measurements, unpredictable environmental changes, and unknown intentions of other traffic agents can contribute to this uncertainty.

    \item \emph {Complex dynamics:} The motion of traffic agents can be affected by various physical laws, including gravity, friction, and aerodynamic forces. These dynamics can be highly complex and nonlinear, making it difficult to model accurately.
   
     \item \emph {Limited sensor coverage:} Autonomous vehicles rely on a suite of sensors, including cameras, LiDAR, and radar, to perceive their environment. However, the coverage of these sensors is limited, as depicted in Fig. 18, and can be affected by occlusions, weather conditions, and other factors that can make it difficult to accurately track the motion of other traffic agents.
    \item \emph {Limited data:} In some cases, there may be limited or incomplete data available for trajectory prediction. This can occur when sensors are malfunctioning, or when the historical data is missing or corrupted.
     
    \item  \emph {Long-term prediction:} Predicting trajectories over a long time horizon (no less than 3 seconds) can be challenging, as small errors in the initial prediction can compound and result in significant deviations from the true trajectory.
    \item \emph {Complex road environments:} Autonomous vehicles operate in complex and dynamic road environments, which can include intersections, roundabouts, and crowded urban areas. Predicting trajectories in these environments requires models that can handle complex interactions between multiple agents, including other vehicles, pedestrians, and cyclists.
    \item \emph {Multimodal Outputs:} In autonomous driving, agents' behaviors exhibit multimodality, where a single past trajectory can have multiple potential future trajectories, as depicted in Fig. 19.

    \item \emph{Sparse and noisy data:} The data from sensors can be sparse and noisy, particularly in urban areas where buildings and other structures can obstruct the line of sight between the sensors and the objects being tracked. This can make it difficult to accurately model the motion of other traffic agents over time.
    \item \emph{Multi-agent interactions: }In many real-world scenarios, multiple agents interact with each other, and their trajectories are interdependent. Predicting the trajectory of one agent may depend on the actions of other agents, as depicted in Fig. 20, making the problem even more challenging.
    \item \emph{Heterogeneous environment:} A Heterogeneous environment refers to an environment that contains a diverse range of elements, such as various types of vehicles, pedestrians, cyclists, different road types, and complex interactions among them. In order to effectively predict trajectories in such environments, prediction models need to account for the different types of agents, incorporate contextual information, fuse sensor data, model interactions among multiple agents, estimate uncertainty, and enable adaptability. 
    
    %The prediction model has to consider the various agent types, incorporating contextual information, fusing sensor data, modeling multi-agent interactions, estimating uncertainty, and enabling adaptability, trajectory prediction systems can better navigate and anticipate the behaviors in diverse and complex environments.
     \item \emph{Safety-critical applications:} Autonomous vehicles are safety-critical systems, and errors in trajectory prediction can have serious consequences, including accidents and injuries. As a result, trajectory prediction algorithms need to be highly accurate and reliable, with well-defined safety margins.

 \item \emph{Real-time constraints:} Autonomous vehicles operate in real-time environments, and trajectory prediction algorithms need to be able to process data and generate predictions in real-time. This requires efficient algorithms and hardware architectures that can handle the large amounts of data generated by the sensors.
\end{enumerate}

\begin{figure}[bht!]
    \centering
    \includegraphics[width=0.48\textwidth]{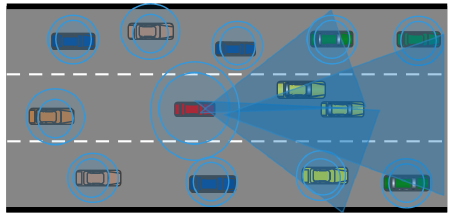}
    \caption{Depiction of limited visibility range of in-car sensors in the context of driving on a three-lane highway. The red vehicle fully covers the Yellow vehicle and partially covers the green vehicle.}
    \label{fig:my_label}
\end{figure}
\begin{figure}[]
    \centering
    \includegraphics[width=0.4\textwidth]{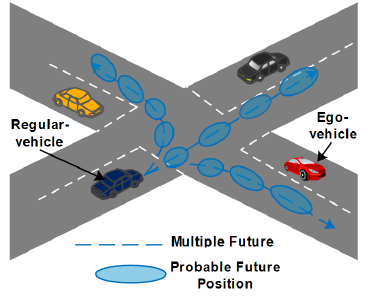}
    \caption{Depiction of the presence of multimodal nature of vehicle and uncertainties in the urban street setting. A common scenario where the self-driving car must decide on its next move while facing various uncertainties related to the anticipated movement of other regular vehicles.}
    \label{fig:my_label}
\end{figure}

\begin{figure}[]
    \centering
    \includegraphics[width=0.4\textwidth]{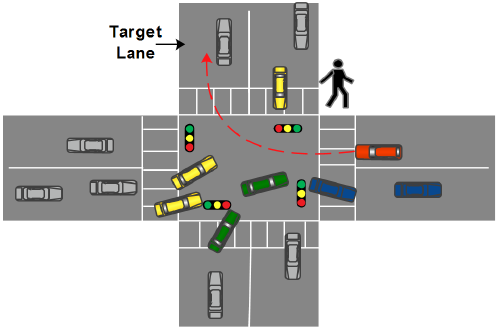}
    \caption{Depiction of the intersection scenarios: The Ego-Vehicle, represented by the Red vehicle, is about to change lanes to the target lane. To ensure socially acceptable driving behavior, the Ego-Vehicle must anticipate the future movements of the yellow vehicles. This prediction also includes taking into account the presence of green vehicles.}
    \label{fig:my_label}
\end{figure}
\subsection{Future Research Directions}
The field of trajectory prediction is undergoing rapid evolution in research, offering numerous opportunities for future investigations, particularly within the realm of autonomous vehicles. Several potential research directions can shape the trajectory prediction landscape. Here, we outline the identified futuristic research directions as follows:
\begin{enumerate}
    \item \emph{Incorporating context and intention:} One limitation of current trajectory prediction methods is that they often focus solely on the motion of other vehicles, without considering the context or intention behind that motion. Future research could explore how to incorporate contextual information, such as road layout and traffic rules, as well as the intention of other drivers, to improve trajectory prediction accuracy.

 \item \emph{Integration of multiple sensors:} 
 Autonomous vehicles rely on a suite of sensors to perceive their environment, and future research could explore how to integrate data from multiple sensors to improve trajectory prediction accuracy. This could involve developing new algorithms for fusing data from cameras, LiDAR, radar, and other sensors, as well as exploring new sensor modalities such as acoustic or thermal sensors.

 \item \emph{Uncertainty modeling:} Trajectory prediction is inherently uncertain, and future research could explore how to model and propagate uncertainty through the prediction pipeline. This could involve developing new probabilistic models, such as Bayesian neural networks, or exploring new techniques for uncertainty quantification and propagation.

 \item \emph{Human-aware trajectory prediction:} Autonomous vehicles operate in environments that include not only other vehicles but also pedestrians and cyclists. Future research could explore how to develop trajectory prediction methods that are aware of human behavior and can accurately predict the motion of pedestrians and cyclists in crowded urban environments.

 \item \emph{Real-time implementation \& Hardware acceleration:} Autonomous vehicles operate in real-time environments, and trajectory prediction algorithms need to be able to process data and generate predictions in real-time. Future research could explore how to optimize trajectory prediction algorithms for real-time performance, as well as developing new hardware architectures for efficient computation. 
 \item \emph{ Ensuring safety and robustness:}  Safety is of paramount importance in autonomous driving systems. Future research should aim to develop trajectory prediction methods that prioritize safety and robustness. This includes investigating techniques for handling rare or anomalous events, improving prediction accuracy in challenging weather conditions, and considering ethical aspects in trajectory prediction algorithms.
 \item \emph{Relative trajectory prediction:} Relative trajectory prediction refers to the task of predicting the future motion or path of surrounding objects or agents relative to the ego vehicle or coordinate system. Future research should focus on estimating the relative displacement, velocities, and trajectories of other vehicles, pedestrians, and cyclists with respect to the ego vehicle. 
 %Future research should focus on the integration of advanced technologies, machine learning approaches, and real-time adaptation capabilities will play a crucial role in achieving these goals.
% \item \textbf{Multi-sensor fusion-based trajectory prediction:} Multi-sensor fusion-based trajectory prediction in autonomous vehicles refers to the process of combining data from multiple sensors, such as lidar, radar, cameras, and more, to predict the future trajectories of surrounding objects. By fusing information from different sensors, the trajectory prediction system can obtain a more comprehensive and accurate understanding of the environment, enabling robustness, better predictions, and decision-making.
 \item \emph{Random obstacle aware trajectory prediction:} This approach refers to predicting the future trajectories of a vehicle while considering the presence of unexpected or random obstacles in the surrounding environment. These obstacles can be animals or objects in between roads, the sudden arrival of pedestrians, and road accidents that lead to an uncertain obstacle in between roads. Future research should focus on incorporating rare events into the prediction models and collecting and analyzing data related to these rare events to develop more comprehensive and robust prediction models.
\item \emph{Challenging Weather condition:}  Adverse weather conditions, such as heavy rain, snow, fog, or low visibility, can affect the performance of sensors and limit the availability of critical data for trajectory prediction. Future research should focus on involves incorporating techniques such as sensor fusion, adaptive filtering, probabilistic modeling, and machine learning to improve the reliability and accuracy of trajectory predictions under adverse weather conditions. 
\item \emph{Vehicle-to-Vehicle (V2V) communication and Vehicle-to-Everything (V2X) communication
strategies:} 

V2V communication refers to the exchange of information directly between vehicles. V2X communication expands beyond V2V and includes communication with other entities such as infrastructure, pedestrians, cyclists, and traffic management systems. By sharing real-time data such as position, speed, acceleration, and intentions, vehicles can collaborate and cooperate to enhance trajectory prediction.
\item \emph{Hybridization of several approaches:}

Multiple strategies are suggested in Sections 3, 4, and 5 for solving the task of trajectory prediction. Hybridization can take different forms depending on the specific context and requirements. This can lead to more accurate and robust trajectory predictions

%The various methods in deep learning can be combined and various methods in reinforcement learning can be combined. and they combine with 
%The hybridization of several approaches such as probabilistic models, deep learning-based models, and reinforcement learning models in trajectory prediction refers to the combination of different techniques, models, or algorithms to enhance the accuracy, robustness, and efficiency of trajectory prediction systems. 

\end{enumerate}

 %should focus the adding the rare events in the prediction model for increasing the safety of AVs.
%These are just a few of the potential research directions in trajectory prediction for autonomous vehicles. As the technology continues to evolve, researchers and engineers will continue to develop new methods and algorithms to improve the accuracy and reliability of trajectory prediction.

\section{Conclusion}

This paper provides an extensive survey of the current state-of-the-art Machine Learning (ML)-based trajectory prediction methods for Autonomous Vehicles (AVs). These ML-based approaches have demonstrated significant promise in accurately predicting trajectories, employing several techniques like deep learning-based methods and reinforcement learning-based methods. Deep learning-based methods including sequential models, vision-based models, and generative models are thoroughly explored, highlighting their respective strengths and weaknesses in trajectory prediction tasks. Furthermore, the review focuses on the discussion of reinforcement learning methods, including Inverse reinforcement learning, deep inverse reinforcement learning, and imitation learning techniques. Multiple informative tables and figures are provided to facilitate a comprehensive comparative study of various approaches used to address trajectory prediction tasks. The review paper includes an analysis of multiple datasets and evaluation metrics used to assess the accuracy of trajectory prediction tasks. This conducts a comparative analysis between deep learning-based methods and reinforcement learning methods across various characteristics. Recent advances in trajectory prediction for AVs show promise, but there are still several challenges that need to be addressed. The paper outlines potential research directions, emphasizing the need for more robust and interpretable models and the exploration of new sensor modalities. The survey aims to provide a valuable reference for researchers and practitioners in this field and guide future advancements in the trajectory prediction domain.

\label{conclusion}

\ifCLASSOPTIONcaptionsoff
  \newpage
\fi



%\bibliographystyle{IEEEtran}
\bibliographystyle{IEEEtran}
\bibliography{refs}

\begin{thebibliography}{100}
\providecommand{\url}[1]{#1}
\csname url@samestyle\endcsname
\providecommand{\newblock}{\relax}
\providecommand{\bibinfo}[2]{#2}
\providecommand{\BIBentrySTDinterwordspacing}{\spaceskip=0pt\relax}
\providecommand{\BIBentryALTinterwordstretchfactor}{4}
\providecommand{\BIBentryALTinterwordspacing}{\spaceskip=\fontdimen2\font plus
\BIBentryALTinterwordstretchfactor\fontdimen3\font minus
  \fontdimen4\font\relax}
\providecommand{\BIBforeignlanguage}[2]{{%
\expandafter\ifx\csname l@#1\endcsname\relax
\typeout{** WARNING: IEEEtran.bst: No hyphenation pattern has been}%
\typeout{** loaded for the language `#1'. Using the pattern for}%
\typeout{** the default language instead.}%
\else
\language=\csname l@#1\endcsname
\fi
#2}}
\providecommand{\BIBdecl}{\relax}
\BIBdecl

\bibitem{123}
``Global status report on road safety 2018: Summary, world health org., geneva,
  switzerland, 2018.''

\bibitem{345}
``Road trauma australia—annual summaries, australia, 2022.''

\bibitem{sae2018taxonomy}
S.~International, ``Taxonomy and definitions for terms related to driving
  automation systems for on-road motor vehicles,'' \emph{SAE international},
  vol. 4970, no. 724, pp. 1--5, 2018.

\bibitem{lefevre2014survey}
S.~Lef{\`e}vre, D.~Vasquez, and C.~Laugier, ``A survey on motion prediction and
  risk assessment for intelligent vehicles,'' \emph{ROBOMECH journal}, vol.~1,
  no.~1, pp. 1--14, 2014.

\bibitem{shirazi2016looking}
M.~S. Shirazi and B.~T. Morris, ``Looking at intersections: a survey of
  intersection monitoring, behavior and safety analysis of recent studies,''
  \emph{IEEE Transactions on Intelligent Transportation Systems}, vol.~18,
  no.~1, pp. 4--24, 2016.

\bibitem{mozaffari2020deep}
S.~Mozaffari, O.~Y. Al-Jarrah, M.~Dianati, P.~Jennings, and A.~Mouzakitis,
  ``Deep learning-based vehicle behavior prediction for autonomous driving
  applications: A review,'' \emph{IEEE Transactions on Intelligent
  Transportation Systems}, vol.~23, no.~1, pp. 33--47, 2020.

\bibitem{leon2021review}
F.~Leon and M.~Gavrilescu, ``A review of tracking and trajectory prediction
  methods for autonomous driving,'' \emph{Mathematics}, vol.~9, no.~6, p. 660,
  2021.

\bibitem{liu2021survey}
J.~Liu, X.~Mao, Y.~Fang, D.~Zhu, and M.~Q.-H. Meng, ``A survey on deep-learning
  approaches for vehicle trajectory prediction in autonomous driving,'' in
  \emph{2021 IEEE International Conference on Robotics and Biomimetics
  (ROBIO)}.\hskip 1em plus 0.5em minus 0.4em\relax IEEE, 2021, pp. 978--985.

\bibitem{karle2022scenario}
P.~Karle, M.~Geisslinger, J.~Betz, and M.~Lienkamp, ``Scenario understanding
  and motion prediction for autonomous vehicles-review and comparison,''
  \emph{IEEE Transactions on Intelligent Transportation Systems}, 2022.

\bibitem{gomes2022review}
I.~Gomes and D.~Wolf, ``A review on intention-aware and interaction-aware
  trajectory prediction for autonomous vehicles,'' 2022.

\bibitem{ghorai2022state}
P.~Ghorai, A.~Eskandarian, Y.-K. Kim, and G.~Mehr, ``State estimation and
  motion prediction of vehicles and vulnerable road users for cooperative
  autonomous driving: A survey,'' \emph{IEEE Transactions on Intelligent
  Transportation Systems}, vol.~23, no.~10, pp. 16\,983--17\,002, 2022.

\bibitem{huang2022survey}
Y.~Huang, J.~Du, Z.~Yang, Z.~Zhou, L.~Zhang, and H.~Chen, ``A survey on
  trajectory-prediction methods for autonomous driving,'' \emph{IEEE
  Transactions on Intelligent Vehicles}, vol.~7, no.~3, pp. 652--674, 2022.

\bibitem{benrachou2022use}
D.~E. Benrachou, S.~Glaser, M.~Elhenawy, and A.~Rakotonirainy, ``Use of social
  interaction and intention to improve motion prediction within automated
  vehicle framework: A review,'' \emph{IEEE Transactions on Intelligent
  Transportation Systems}, 2022.

\bibitem{xie2017vehicle}
G.~Xie, H.~Gao, L.~Qian, B.~Huang, K.~Li, and J.~Wang, ``Vehicle trajectory
  prediction by integrating physics-and maneuver-based approaches using
  interactive multiple models,'' \emph{IEEE Transactions on Industrial
  Electronics}, vol.~65, no.~7, pp. 5999--6008, 2017.

\bibitem{batz2009recognition}
T.~Batz, K.~Watson, and J.~Beyerer, ``Recognition of dangerous situations
  within a cooperative group of vehicles,'' in \emph{2009 IEEE Intelligent
  Vehicles Symposium}.\hskip 1em plus 0.5em minus 0.4em\relax IEEE, 2009, pp.
  907--912.

\bibitem{lefkopoulos2020interaction}
V.~Lefkopoulos, M.~Menner, A.~Domahidi, and M.~N. Zeilinger,
  ``Interaction-aware motion prediction for autonomous driving: A multiple
  model kalman filtering scheme,'' \emph{IEEE Robotics and Automation Letters},
  vol.~6, no.~1, pp. 80--87, 2020.

\bibitem{wang2021decision}
Y.~Wang, C.~Wang, W.~Zhao, and C.~Xu, ``Decision-making and planning method for
  autonomous vehicles based on motivation and risk assessment,'' \emph{IEEE
  Transactions on Vehicular Technology}, vol.~70, no.~1, pp. 107--120, 2021.

\bibitem{zhang2020research}
S.~Zhang, Y.~Zhi, R.~He, and J.~Li, ``Research on traffic vehicle behavior
  prediction method based on game theory and hmm,'' \emph{IEEE Access}, vol.~8,
  pp. 30\,210--30\,222, 2020.

\bibitem{li2019dynamic}
J.~Li, B.~Dai, X.~Li, X.~Xu, and D.~Liu, ``A dynamic bayesian network for
  vehicle maneuver prediction in highway driving scenarios: Framework and
  verification,'' \emph{Electronics}, vol.~8, no.~1, p.~40, 2019.

\bibitem{graves2013generating}
A.~Graves, ``Generating sequences with recurrent neural networks,'' \emph{arXiv
  preprint arXiv:1308.0850}, 2013.

\bibitem{deo2018convolutional}
N.~Deo and M.~M. Trivedi, ``Convolutional social pooling for vehicle trajectory
  prediction,'' in \emph{Proceedings of the IEEE conference on computer vision
  and pattern recognition workshops}, 2018, pp. 1468--1476.

\bibitem{zhao2019multi}
T.~Zhao, Y.~Xu, M.~Monfort, W.~Choi, C.~Baker, Y.~Zhao, Y.~Wang, and Y.~N. Wu,
  ``Multi-agent tensor fusion for contextual trajectory prediction,'' in
  \emph{Proceedings of the IEEE/CVF Conference on Computer Vision and Pattern
  Recognition}, 2019, pp. 12\,126--12\,134.

\bibitem{bhattacharyya2019conditional}
A.~Bhattacharyya, M.~Hanselmann, M.~Fritz, B.~Schiele, and C.-N. Straehle,
  ``Conditional flow variational autoencoders for structured sequence
  prediction,'' \emph{arXiv preprint arXiv:1908.09008}, 2019.

\bibitem{quintanar2021predicting}
A.~Quintanar, D.~Fern{\'a}ndez-Llorca, I.~Parra, R.~Izquierdo, and M.~Sotelo,
  ``Predicting vehicles trajectories in urban scenarios with transformer
  networks and augmented information,'' in \emph{2021 IEEE Intelligent Vehicles
  Symposium (IV)}.\hskip 1em plus 0.5em minus 0.4em\relax IEEE, 2021, pp.
  1051--1056.

\bibitem{ziebart2008maximum}
B.~D. Ziebart, A.~L. Maas, J.~A. Bagnell, A.~K. Dey \emph{et~al.}, ``Maximum
  entropy inverse reinforcement learning.'' in \emph{Aaai}, vol.~8.\hskip 1em
  plus 0.5em minus 0.4em\relax Chicago, IL, USA, 2008, pp. 1433--1438.

\bibitem{you2019advanced}
C.~You, J.~Lu, D.~Filev, and P.~Tsiotras, ``Advanced planning for autonomous
  vehicles using reinforcement learning and deep inverse reinforcement
  learning,'' \emph{Robotics and Autonomous Systems}, vol. 114, pp. 1--18,
  2019.

\bibitem{hu2022model}
A.~Hu, G.~Corrado, N.~Griffiths, Z.~Murez, C.~Gurau, H.~Yeo, A.~Kendall,
  R.~Cipolla, and J.~Shotton, ``Model-based imitation learning for urban
  driving,'' \emph{Advances in Neural Information Processing Systems}, vol.~35,
  pp. 20\,703--20\,716, 2022.

\bibitem{rajamani2011vehicle}
R.~Rajamani, \emph{Vehicle dynamics and control}.\hskip 1em plus 0.5em minus
  0.4em\relax Springer Science \& Business Media, 2011.

\bibitem{brannstrom2010model}
M.~Br{\"a}nnstr{\"o}m, E.~Coelingh, and J.~Sj{\"o}berg, ``Model-based threat
  assessment for avoiding arbitrary vehicle collisions,'' \emph{IEEE
  Transactions on Intelligent Transportation Systems}, vol.~11, no.~3, pp.
  658--669, 2010.

\bibitem{lin2000vehicle}
C.-F. Lin, A.~G. Ulsoy, and D.~J. LeBlanc, ``Vehicle dynamics and external
  disturbance estimation for vehicle path prediction,'' \emph{IEEE Transactions
  on Control Systems Technology}, vol.~8, no.~3, pp. 508--518, 2000.

\bibitem{huang2006vehicle}
J.~Huang and H.-S. Tan, ``Vehicle future trajectory prediction with a
  dgps/ins-based positioning system,'' in \emph{2006 American Control
  Conference}.\hskip 1em plus 0.5em minus 0.4em\relax IEEE, 2006, pp. 6--pp.

\bibitem{pepy2006reducing}
R.~Pepy, A.~Lambert, and H.~Mounier, ``Reducing navigation errors by planning
  with realistic vehicle model,'' in \emph{2006 IEEE Intelligent Vehicles
  Symposium}.\hskip 1em plus 0.5em minus 0.4em\relax IEEE, 2006, pp. 300--307.

\bibitem{kaempchen2009situation}
N.~Kaempchen, B.~Schiele, and K.~Dietmayer, ``Situation assessment of an
  autonomous emergency brake for arbitrary vehicle-to-vehicle collision
  scenarios,'' \emph{IEEE Transactions on Intelligent Transportation Systems},
  vol.~10, no.~4, pp. 678--687, 2009.

\bibitem{ammoun2009real}
S.~Ammoun and F.~Nashashibi, ``Real time trajectory prediction for collision
  risk estimation between vehicles,'' in \emph{2009 IEEE 5th International
  Conference on Intelligent Computer Communication and Processing}.\hskip 1em
  plus 0.5em minus 0.4em\relax IEEE, 2009, pp. 417--422.

\bibitem{schubert2008comparison}
R.~Schubert, E.~Richter, and G.~Wanielik, ``Comparison and evaluation of
  advanced motion models for vehicle tracking,'' in \emph{2008 11th
  international conference on information fusion}.\hskip 1em plus 0.5em minus
  0.4em\relax IEEE, 2008, pp. 1--6.

\bibitem{polychronopoulos2007sensor}
A.~Polychronopoulos, M.~Tsogas, A.~J. Amditis, and L.~Andreone, ``Sensor fusion
  for predicting vehicles' path for collision avoidance systems,'' \emph{IEEE
  Transactions on Intelligent Transportation Systems}, vol.~8, no.~3, pp.
  549--562, 2007.

\bibitem{lytrivis2008cooperative}
P.~Lytrivis, G.~Thomaidis, and A.~Amditis, ``Cooperative path prediction in
  vehicular environments,'' in \emph{2008 11th International IEEE Conference on
  Intelligent Transportation Systems}.\hskip 1em plus 0.5em minus 0.4em\relax
  IEEE, 2008, pp. 803--808.

\bibitem{barth2008will}
A.~Barth and U.~Franke, ``Where will the oncoming vehicle be the next second?''
  in \emph{2008 IEEE Intelligent Vehicles Symposium}.\hskip 1em plus 0.5em
  minus 0.4em\relax IEEE, 2008, pp. 1068--1073.

\bibitem{zhang2017method}
R.~Zhang, L.~Cao, S.~Bao, and J.~Tan, ``A method for connected vehicle
  trajectory prediction and collision warning algorithm based on v2v
  communication,'' \emph{International Journal of Crashworthiness}, vol.~22,
  no.~1, pp. 15--25, 2017.

\bibitem{okamoto2017driver}
K.~Okamoto, K.~Berntorp, and S.~Di~Cairano, ``Driver intention-based vehicle
  threat assessment using random forests and particle filtering,''
  \emph{IFAC-PapersOnLine}, vol.~50, no.~1, pp. 13\,860--13\,865, 2017.

\bibitem{wang2019trajectory}
Y.~Wang, Z.~Liu, Z.~Zuo, Z.~Li, L.~Wang, and X.~Luo, ``Trajectory planning and
  safety assessment of autonomous vehicles based on motion prediction and model
  predictive control,'' \emph{IEEE Transactions on Vehicular Technology},
  vol.~68, no.~9, pp. 8546--8556, 2019.

\bibitem{houenou2013vehicle}
A.~Houenou, P.~Bonnifait, V.~Cherfaoui, and W.~Yao, ``Vehicle trajectory
  prediction based on motion model and maneuver recognition,'' in \emph{2013
  IEEE/RSJ international conference on intelligent robots and systems}.\hskip
  1em plus 0.5em minus 0.4em\relax IEEE, 2013, pp. 4363--4369.

\bibitem{tran2014online}
Q.~Tran and J.~Firl, ``Online maneuver recognition and multimodal trajectory
  prediction for intersection assistance using non-parametric regression,'' in
  \emph{2014 ieee intelligent vehicles symposium proceedings}.\hskip 1em plus
  0.5em minus 0.4em\relax IEEE, 2014, pp. 918--923.

\bibitem{wissing2018interaction}
C.~Wissing, T.~Nattermann, K.-H. Glander, and T.~Bertram, ``Interaction-aware
  long-term driving situation prediction,'' in \emph{2018 21st International
  Conference on Intelligent Transportation Systems (ITSC)}.\hskip 1em plus
  0.5em minus 0.4em\relax IEEE, 2018, pp. 137--143.

\bibitem{albeaik2022limitations}
S.~Albeaik, A.~Bayen, M.~T. Chiri, X.~Gong, A.~Hayat, N.~Kardous, A.~Keimer,
  S.~T. McQuade, B.~Piccoli, and Y.~You, ``Limitations and improvements of the
  intelligent driver model (idm),'' \emph{SIAM Journal on Applied Dynamical
  Systems}, vol.~21, no.~3, pp. 1862--1892, 2022.

\bibitem{hu2018probabilistic}
Y.~Hu, W.~Zhan, and M.~Tomizuka, ``Probabilistic prediction of vehicle semantic
  intention and motion,'' in \emph{2018 IEEE Intelligent Vehicles Symposium
  (IV)}.\hskip 1em plus 0.5em minus 0.4em\relax IEEE, 2018, pp. 307--313.

\bibitem{augustin2019prediction}
D.~Augustin, M.~Hofmann, and U.~Konigorski, ``Prediction of highway lane
  changes based on prototype trajectories,'' \emph{Forschung im
  Ingenieurwesen}, vol.~83, no.~2, pp. 149--161, 2019.

\bibitem{wirthmuller2020teaching}
F.~Wirthm{\"u}ller, J.~Schlechtriemen, J.~Hipp, and M.~Reichert, ``Teaching
  vehicles to anticipate: A systematic study on probabilistic behavior
  prediction using large data sets,'' \emph{IEEE Transactions on Intelligent
  Transportation Systems}, vol.~22, no.~11, pp. 7129--7144, 2020.

\bibitem{deo2018would}
N.~Deo, A.~Rangesh, and M.~M. Trivedi, ``How would surround vehicles move? a
  unified framework for maneuver classification and motion prediction,''
  \emph{IEEE Transactions on Intelligent Vehicles}, vol.~3, no.~2, pp.
  129--140, 2018.

\bibitem{laugier2011probabilistic}
C.~Laugier, I.~E. Paromtchik, M.~Perrollaz, M.~Yong, J.-D. Yoder, C.~Tay,
  K.~Mekhnacha, and A.~N{\`e}gre, ``Probabilistic analysis of dynamic scenes
  and collision risks assessment to improve driving safety,'' \emph{IEEE
  Intelligent Transportation Systems Magazine}, vol.~3, no.~4, pp. 4--19, 2011.

\bibitem{trautman2010unfreezing}
P.~Trautman and A.~Krause, ``Unfreezing the robot: Navigation in dense,
  interacting crowds,'' in \emph{2010 IEEE/RSJ International Conference on
  Intelligent Robots and Systems}.\hskip 1em plus 0.5em minus 0.4em\relax IEEE,
  2010, pp. 797--803.

\bibitem{guo2019modeling}
Y.~Guo, V.~V. Kalidindi, M.~Arief, W.~Wang, J.~Zhu, H.~Peng, and D.~Zhao,
  ``Modeling multi-vehicle interaction scenarios using gaussian random field,''
  in \emph{2019 IEEE Intelligent Transportation Systems Conference
  (ITSC)}.\hskip 1em plus 0.5em minus 0.4em\relax IEEE, 2019, pp. 3974--3980.

\bibitem{vasquez2004motion}
D.~Vasquez and T.~Fraichard, ``Motion prediction for moving objects: a
  statistical approach,'' in \emph{IEEE International Conference on Robotics
  and Automation, 2004. Proceedings. ICRA'04. 2004}, vol.~4.\hskip 1em plus
  0.5em minus 0.4em\relax IEEE, 2004, pp. 3931--3936.

\bibitem{hermes2009long}
C.~Hermes, C.~Wohler, K.~Schenk, and F.~Kummert, ``Long-term vehicle motion
  prediction,'' in \emph{2009 IEEE intelligent vehicles symposium}.\hskip 1em
  plus 0.5em minus 0.4em\relax IEEE, 2009, pp. 652--657.

\bibitem{qiao2014self}
S.~Qiao, D.~Shen, X.~Wang, N.~Han, and W.~Zhu, ``A self-adaptive parameter
  selection trajectory prediction approach via hidden markov models,''
  \emph{IEEE Transactions on Intelligent Transportation Systems}, vol.~16,
  no.~1, pp. 284--296, 2014.

\bibitem{deng2018improved}
Q.~Deng and D.~S{\"o}ffker, ``Improved driving behaviors prediction based on
  fuzzy logic-hidden markov model (fl-hmm),'' in \emph{2018 IEEE Intelligent
  Vehicles Symposium (IV)}.\hskip 1em plus 0.5em minus 0.4em\relax IEEE, 2018,
  pp. 2003--2008.

\bibitem{murphy2002dynamic}
K.~P. Murphy, \emph{Dynamic bayesian networks: representation, inference and
  learning}.\hskip 1em plus 0.5em minus 0.4em\relax University of California,
  Berkeley, 2002.

\bibitem{gindele2015learning}
T.~Gindele, S.~Brechtel, and R.~Dillmann, ``Learning driver behavior models
  from traffic observations for decision making and planning,'' \emph{IEEE
  Intelligent Transportation Systems Magazine}, vol.~7, no.~1, pp. 69--79,
  2015.

\bibitem{schreier2016integrated}
M.~Schreier, V.~Willert, and J.~Adamy, ``An integrated approach to
  maneuver-based trajectory prediction and criticality assessment in arbitrary
  road environments,'' \emph{IEEE Transactions on Intelligent Transportation
  Systems}, vol.~17, no.~10, pp. 2751--2766, 2016.

\bibitem{bahram2015game}
M.~Bahram, A.~Lawitzky, J.~Friedrichs, M.~Aeberhard, and D.~Wollherr, ``A
  game-theoretic approach to replanning-aware interactive scene prediction and
  planning,'' \emph{IEEE Transactions on Vehicular Technology}, vol.~65, no.~6,
  pp. 3981--3992, 2015.

\bibitem{he2019probabilistic}
G.~He, X.~Li, Y.~Lv, B.~Gao, and H.~Chen, ``Probabilistic intention prediction
  and trajectory generation based on dynamic bayesian networks,'' in \emph{2019
  Chinese Automation Congress (CAC)}.\hskip 1em plus 0.5em minus 0.4em\relax
  IEEE, 2019, pp. 2646--2651.

\bibitem{jiang2022vehicle}
Y.~Jiang, B.~Zhu, S.~Yang, J.~Zhao, and W.~Deng, ``Vehicle trajectory
  prediction considering driver uncertainty and vehicle dynamics based on
  dynamic bayesian network,'' \emph{IEEE Transactions on Systems, Man, and
  Cybernetics: Systems}, 2022.

\bibitem{hewing2020simulation}
L.~Hewing, E.~Arcari, L.~P. Fr{\"o}hlich, and M.~N. Zeilinger, ``On simulation
  and trajectory prediction with gaussian process dynamics,'' in \emph{Learning
  for Dynamics and Control}.\hskip 1em plus 0.5em minus 0.4em\relax PMLR, 2020,
  pp. 424--434.

\bibitem{li2022autonomous}
T.~Li, L.~Chen, and Y.~Wang, ``Autonomous driving behavior prediction method
  based on improved hidden markov model,'' in \emph{2022 IEEE 25th
  International Conference on Computer Supported Cooperative Work in Design
  (CSCWD)}.\hskip 1em plus 0.5em minus 0.4em\relax IEEE, 2022, pp. 758--762.

\bibitem{ren2022method}
Y.-Y. Ren, L.~Zhao, X.-L. Zheng, and X.-S. Li, ``A method for predicting
  diverse lane-changing trajectories of surrounding vehicles based on early
  detection of lane change,'' \emph{IEEE access}, vol.~10, pp.
  17\,451--17\,472, 2022.

\bibitem{zyner2018recurrent}
A.~Zyner, S.~Worrall, and E.~Nebot, ``A recurrent neural network solution for
  predicting driver intention at unsignalized intersections,'' \emph{IEEE
  Robotics and Automation Letters}, vol.~3, no.~3, pp. 1759--1764, 2018.

\bibitem{ding2019online}
W.~Ding and S.~Shen, ``Online vehicle trajectory prediction using policy
  anticipation network and optimization-based context reasoning,'' in
  \emph{2019 International Conference on Robotics and Automation (ICRA)}.\hskip
  1em plus 0.5em minus 0.4em\relax IEEE, 2019, pp. 9610--9616.

\bibitem{chang2019argoverse}
M.-F. Chang, J.~Lambert, P.~Sangkloy, J.~Singh, S.~Bak, A.~Hartnett, D.~Wang,
  P.~Carr, S.~Lucey, D.~Ramanan \emph{et~al.}, ``Argoverse: 3d tracking and
  forecasting with rich maps,'' in \emph{Proceedings of the IEEE/CVF conference
  on computer vision and pattern recognition}, 2019, pp. 8748--8757.

\bibitem{zyner2019naturalistic}
A.~Zyner, S.~Worrall, and E.~Nebot, ``Naturalistic driver intention and path
  prediction using recurrent neural networks,'' \emph{IEEE transactions on
  intelligent transportation systems}, vol.~21, no.~4, pp. 1584--1594, 2019.

\bibitem{xing2019personalized}
Y.~Xing, C.~Lv, and D.~Cao, ``Personalized vehicle trajectory prediction based
  on joint time-series modeling for connected vehicles,'' \emph{IEEE
  Transactions on Vehicular Technology}, vol.~69, no.~2, pp. 1341--1352, 2019.

\bibitem{xin2018intention}
L.~Xin, P.~Wang, C.-Y. Chan, J.~Chen, S.~E. Li, and B.~Cheng, ``Intention-aware
  long horizon trajectory prediction of surrounding vehicles using dual lstm
  networks,'' in \emph{2018 21st International Conference on Intelligent
  Transportation Systems (ITSC)}.\hskip 1em plus 0.5em minus 0.4em\relax IEEE,
  2018, pp. 1441--1446.

\bibitem{deo2018multi}
N.~Deo and M.~M. Trivedi, ``Multi-modal trajectory prediction of surrounding
  vehicles with maneuver based lstms,'' in \emph{2018 IEEE intelligent vehicles
  symposium (IV)}.\hskip 1em plus 0.5em minus 0.4em\relax IEEE, 2018, pp.
  1179--1184.

\bibitem{dai2019modeling}
S.~Dai, L.~Li, and Z.~Li, ``Modeling vehicle interactions via modified lstm
  models for trajectory prediction,'' \emph{IEEE Access}, vol.~7, pp.
  38\,287--38\,296, 2019.

\bibitem{ding2019predicting}
W.~Ding, J.~Chen, and S.~Shen, ``Predicting vehicle behaviors over an extended
  horizon using behavior interaction network,'' in \emph{2019 International
  Conference on Robotics and Automation (ICRA)}.\hskip 1em plus 0.5em minus
  0.4em\relax IEEE, 2019, pp. 8634--8640.

\bibitem{tang2019multiple}
C.~Tang and R.~R. Salakhutdinov, ``Multiple futures prediction,''
  \emph{Advances in neural information processing systems}, vol.~32, 2019.

\bibitem{zhang2021vehicle}
T.~Zhang, W.~Song, M.~Fu, Y.~Yang, and M.~Wang, ``Vehicle motion prediction at
  intersections based on the turning intention and prior trajectories model,''
  \emph{IEEE/CAA Journal of Automatica Sinica}, vol.~8, no.~10, pp. 1657--1666,
  2021.

\bibitem{zhang2020lane}
Y.~Zhang, Y.~Zou, J.~Tang, and J.~Liang, ``A lane-changing prediction method
  based on temporal convolution network,'' \emph{arXiv preprint
  arXiv:2011.01224}, 2020.

\bibitem{strohbeck2020multiple}
J.~Strohbeck, V.~Belagiannis, J.~M{\"u}ller, M.~Schreiber, M.~Herrmann,
  D.~Wolf, and M.~Buchholz, ``Multiple trajectory prediction with deep temporal
  and spatial convolutional neural networks,'' in \emph{2020 IEEE/RSJ
  International Conference on Intelligent Robots and Systems (IROS)}.\hskip 1em
  plus 0.5em minus 0.4em\relax IEEE, 2020, pp. 1992--1998.

\bibitem{katariya2022deeptrack}
V.~Katariya, M.~Baharani, N.~Morris, O.~Shoghli, and H.~Tabkhi, ``Deeptrack:
  Lightweight deep learning for vehicle trajectory prediction in highways,''
  \emph{IEEE Transactions on Intelligent Transportation Systems}, vol.~23,
  no.~10, pp. 18\,927--18\,936, 2022.

\bibitem{li2022vehicle}
D.~Li, H.~Li, Y.~Xiao, B.~Li, and B.~Tang, ``Vehicle trajectory prediction for
  automated driving based on temporal convolution networks,'' in \emph{2022 WRC
  Symposium on Advanced Robotics and Automation (WRC SARA)}.\hskip 1em plus
  0.5em minus 0.4em\relax IEEE, 2022, pp. 257--262.

\bibitem{azadani2022novel}
M.~N. Azadani and A.~Boukerche, ``A novel multimodal vehicle path prediction
  method based on temporal convolutional networks,'' \emph{IEEE Transactions on
  Intelligent Transportation Systems}, vol.~23, no.~12, pp. 25\,384--25\,395,
  2022.

\bibitem{hao2020attention}
Z.~Hao, X.~Huang, K.~Wang, M.~Cui, and Y.~Tian, ``Attention-based gru for
  driver intention recognition and vehicle trajectory prediction,'' in
  \emph{2020 4th CAA international conference on vehicular control and
  intelligence (CVCI)}.\hskip 1em plus 0.5em minus 0.4em\relax IEEE, 2020, pp.
  86--91.

\bibitem{yan2020trajectory}
J.~Yan, Z.~Peng, H.~Yin, J.~Wang, X.~Wang, Y.~Shen, W.~Stechele, and
  D.~Cremers, ``Trajectory prediction for intelligent vehicles using
  spatial-attention mechanism,'' \emph{IET Intelligent Transport Systems},
  vol.~14, no.~13, pp. 1855--1863, 2020.

\bibitem{kim2021lapred}
B.~Kim, S.~H. Park, S.~Lee, E.~Khoshimjonov, D.~Kum, J.~Kim, J.~S. Kim, and
  J.~W. Choi, ``Lapred: Lane-aware prediction of multi-modal future
  trajectories of dynamic agents,'' in \emph{Proceedings of the IEEE/CVF
  Conference on Computer Vision and Pattern Recognition}, 2021, pp.
  14\,636--14\,645.

\bibitem{fu2021trajectory}
M.~Fu, T.~Zhang, W.~Song, Y.~Yang, and M.~Wang, ``Trajectory prediction-based
  local spatio-temporal navigation map for autonomous driving in dynamic
  highway environments,'' \emph{IEEE Transactions on Intelligent Transportation
  Systems}, vol.~23, no.~7, pp. 6418--6429, 2021.

\bibitem{yu2021dynamic}
J.~Yu, M.~Zhou, X.~Wang, G.~Pu, C.~Cheng, and B.~Chen, ``A dynamic and static
  context-aware attention network for trajectory prediction,'' \emph{ISPRS
  International Journal of Geo-Information}, vol.~10, no.~5, p. 336, 2021.

\bibitem{wu2021hsta}
Y.~Wu, G.~Chen, Z.~Li, L.~Zhang, L.~Xiong, Z.~Liu, and A.~Knoll, ``Hsta: A
  hierarchical spatio-temporal attention model for trajectory prediction,''
  \emph{IEEE Transactions on Vehicular Technology}, vol.~70, no.~11, pp.
  11\,295--11\,307, 2021.

\bibitem{meng2021intelligent}
Q.~Meng, B.~Shang, Y.~Liu, H.~Guo, and X.~Zhao, ``Intelligent vehicles
  trajectory prediction with spatial and temporal attention mechanism,''
  \emph{IFAC-PapersOnLine}, vol.~54, no.~10, pp. 454--459, 2021.

\bibitem{kim2020multi}
H.~Kim, D.~Kim, G.~Kim, J.~Cho, and K.~Huh, ``Multi-head attention based
  probabilistic vehicle trajectory prediction,'' in \emph{2020 IEEE Intelligent
  Vehicles Symposium (IV)}.\hskip 1em plus 0.5em minus 0.4em\relax IEEE, 2020,
  pp. 1720--1725.

\bibitem{messaoud2020attention}
K.~Messaoud, I.~Yahiaoui, A.~Verroust-Blondet, and F.~Nashashibi, ``Attention
  based vehicle trajectory prediction,'' \emph{IEEE Transactions on Intelligent
  Vehicles}, vol.~6, no.~1, pp. 175--185, 2020.

\bibitem{messaoud2021trajectory}
K.~Messaoud, N.~Deo, M.~M. Trivedi, and F.~Nashashibi, ``Trajectory prediction
  for autonomous driving based on multi-head attention with joint agent-map
  representation,'' in \emph{2021 IEEE Intelligent Vehicles Symposium
  (IV)}.\hskip 1em plus 0.5em minus 0.4em\relax IEEE, 2021, pp. 165--170.

\bibitem{lin2021vehicle}
L.~Lin, W.~Li, H.~Bi, and L.~Qin, ``Vehicle trajectory prediction using lstms
  with spatial--temporal attention mechanisms,'' \emph{IEEE Intelligent
  Transportation Systems Magazine}, vol.~14, no.~2, pp. 197--208, 2021.

\bibitem{yang2022lane}
S.~Yang, Y.~Chen, Y.~Cao, R.~Wang, R.~Shi, and J.~Lu, ``Lane change trajectory
  prediction based on spatiotemporal attention mechanism,'' in \emph{2022 IEEE
  25th International Conference on Intelligent Transportation Systems
  (ITSC)}.\hskip 1em plus 0.5em minus 0.4em\relax IEEE, 2022, pp. 2366--2371.

\bibitem{kim2022diverse}
S.~Kim, H.~Jeon, J.~W. Choi, and D.~Kum, ``Diverse multiple trajectory
  prediction using a two-stage prediction network trained with lane loss,''
  \emph{IEEE Robotics and Automation Letters}, 2022.

\bibitem{hu2022trajectory}
H.~Hu, Q.~Wang, M.~Cheng, and Z.~Gao, ``Trajectory prediction neural network
  and model interpretation based on temporal pattern attention,'' \emph{IEEE
  Transactions on Intelligent Transportation Systems}, 2022.

\bibitem{hasan2023mals}
F.~Hasan and H.~Huang, ``Mals-net: A multi-head attention-based lstm
  sequence-to-sequence network for socio-temporal interaction modelling and
  trajectory prediction,'' \emph{Sensors}, vol.~23, no.~1, p. 530, 2023.

\bibitem{sutskever2014sequence}
I.~Sutskever, O.~Vinyals, and Q.~V. Le, ``Sequence to sequence learning with
  neural networks,'' \emph{Advances in neural information processing systems},
  vol.~27, 2014.

\bibitem{zyner2017long}
A.~Zyner, S.~Worrall, J.~Ward, and E.~Nebot, ``Long short term memory for
  driver intent prediction,'' in \emph{2017 IEEE Intelligent Vehicles Symposium
  (IV)}.\hskip 1em plus 0.5em minus 0.4em\relax IEEE, 2017, pp. 1484--1489.

\bibitem{phillips2017generalizable}
D.~J. Phillips, T.~A. Wheeler, and M.~J. Kochenderfer, ``Generalizable
  intention prediction of human drivers at intersections,'' in \emph{2017 IEEE
  intelligent vehicles symposium (IV)}.\hskip 1em plus 0.5em minus 0.4em\relax
  IEEE, 2017, pp. 1665--1670.

\bibitem{altche2017lstm}
F.~Altch{\'e} and A.~de~La~Fortelle, ``An lstm network for highway trajectory
  prediction,'' in \emph{2017 IEEE 20th international conference on intelligent
  transportation systems (ITSC)}.\hskip 1em plus 0.5em minus 0.4em\relax IEEE,
  2017, pp. 353--359.

\bibitem{kawasaki2020multimodal}
A.~Kawasaki and A.~Seki, ``Multimodal trajectory predictions for urban
  environments using geometric relationships between a vehicle and lanes,'' in
  \emph{2020 IEEE International Conference on Robotics and Automation
  (ICRA)}.\hskip 1em plus 0.5em minus 0.4em\relax IEEE, 2020, pp. 9203--9209.

\bibitem{min2019rnn}
K.~Min, D.~Kim, J.~Park, and K.~Huh, ``Rnn-based path prediction of obstacle
  vehicles with deep ensemble,'' \emph{IEEE Transactions on Vehicular
  Technology}, vol.~68, no.~10, pp. 10\,252--10\,256, 2019.

\bibitem{mercat2020multi}
J.~Mercat, T.~Gilles, N.~El~Zoghby, G.~Sandou, D.~Beauvois, and G.~P. Gil,
  ``Multi-head attention for multi-modal joint vehicle motion forecasting,'' in
  \emph{2020 IEEE International Conference on Robotics and Automation
  (ICRA)}.\hskip 1em plus 0.5em minus 0.4em\relax IEEE, 2020, pp. 9638--9644.

\bibitem{xie2021congestion}
X.~Xie, C.~Zhang, Y.~Zhu, Y.~N. Wu, and S.-C. Zhu, ``Congestion-aware
  multi-agent trajectory prediction for collision avoidance,'' in \emph{2021
  IEEE International Conference on Robotics and Automation (ICRA)}.\hskip 1em
  plus 0.5em minus 0.4em\relax IEEE, 2021, pp. 13\,693--13\,700.

\bibitem{sharma2021towards}
A.~Sharma and D.~B. Jayagopi, ``Towards efficient unconstrained handwriting
  recognition using dilated temporal convolution network,'' \emph{Expert
  Systems with Applications}, vol. 164, p. 114004, 2021.

\bibitem{pandey2019tcnn}
A.~Pandey and D.~Wang, ``Tcnn: Temporal convolutional neural network for
  real-time speech enhancement in the time domain,'' in \emph{ICASSP 2019-2019
  IEEE International Conference on Acoustics, Speech and Signal Processing
  (ICASSP)}.\hskip 1em plus 0.5em minus 0.4em\relax IEEE, 2019, pp. 6875--6879.

\bibitem{zheng2022denoising}
Z.~Zheng, Z.~Zhang, L.~Wang, and X.~Luo, ``Denoising temporal convolutional
  recurrent autoencoders for time series classification,'' \emph{Information
  Sciences}, vol. 588, pp. 159--173, 2022.

\bibitem{bai2018empirical}
S.~Bai, J.~Z. Kolter, and V.~Koltun, ``An empirical evaluation of generic
  convolutional and recurrent networks for sequence modeling,'' \emph{arXiv
  preprint arXiv:1803.01271}, 2018.

\bibitem{liu2021multimodal}
Y.~Liu, J.~Zhang, L.~Fang, Q.~Jiang, and B.~Zhou, ``Multimodal motion
  prediction with stacked transformers,'' in \emph{Proceedings of the IEEE/CVF
  Conference on Computer Vision and Pattern Recognition}, 2021, pp. 7577--7586.

\bibitem{zhao2021spatial}
J.~Zhao, X.~Li, Q.~Xue, and W.~Zhang, ``Spatial-channel transformer network for
  trajectory prediction on the traffic scenes,'' \emph{arXiv preprint
  arXiv:2101.11472}, 2021.

\bibitem{chen2021s2tnet}
W.~Chen, F.~Wang, and H.~Sun, ``S2tnet: Spatio-temporal transformer networks
  for trajectory prediction in autonomous driving,'' in \emph{Asian Conference
  on Machine Learning}.\hskip 1em plus 0.5em minus 0.4em\relax PMLR, 2021, pp.
  454--469.

\bibitem{chen2022vehicle}
X.~Chen, H.~Zhang, F.~Zhao, Y.~Cai, H.~Wang, and Q.~Ye, ``Vehicle trajectory
  prediction based on intention-aware non-autoregressive transformer with
  multi-attention learning for internet of vehicles,'' \emph{IEEE Transactions
  on Instrumentation and Measurement}, vol.~71, pp. 1--12, 2022.

\bibitem{hou2022structural}
L.~Hou, S.~E. Li, B.~Yang, Z.~Wang, and K.~Nakano, ``Structural transformer
  improves speed-accuracy trade-off in interactive trajectory prediction of
  multiple surrounding vehicles,'' \emph{IEEE Transactions on Intelligent
  Transportation Systems}, vol.~23, no.~12, pp. 24\,778--24\,790, 2022.

\bibitem{huang2022multi}
Z.~Huang, X.~Mo, and C.~Lv, ``Multi-modal motion prediction with
  transformer-based neural network for autonomous driving,'' in \emph{2022
  International Conference on Robotics and Automation (ICRA)}.\hskip 1em plus
  0.5em minus 0.4em\relax IEEE, 2022, pp. 2605--2611.

\bibitem{ngiam2022scene}
J.~Ngiam, V.~Vasudevan, B.~Caine, Z.~Zhang, H.-T.~L. Chiang, J.~Ling,
  R.~Roelofs, A.~Bewley, C.~Liu, A.~Venugopal \emph{et~al.}, ``Scene
  transformer: A unified architecture for predicting future trajectories of
  multiple agents,'' in \emph{International Conference on Learning
  Representations}, 2022.

\bibitem{wang2023safety}
X.~Wang, K.~Tang, X.~Dai, J.~Xu, J.~Xi, R.~Ai, Y.~Wang, W.~Gu, and C.~Sun,
  ``Safety-balanced driving-style aware trajectory planning in intersection
  scenarios with uncertain environment,'' \emph{IEEE Transactions on
  Intelligent Vehicles}, 2023.

\bibitem{gao2023dual}
K.~Gao, X.~Li, B.~Chen, L.~Hu, J.~Liu, R.~Du, and Y.~Li, ``Dual transformer
  based prediction for lane change intentions and trajectories in mixed traffic
  environment,'' \emph{IEEE Transactions on Intelligent Transportation
  Systems}, 2023.

\bibitem{wang2023lane}
Z.~Wang, J.~Guo, Z.~Hu, H.~Zhang, J.~Zhang, and J.~Pu, ``Lane transformer: A
  high efficiency trajectory prediction model,'' \emph{IEEE Open Journal of
  Intelligent Transportation Systems}, 2023.

\bibitem{chorowski2015attention}
J.~K. Chorowski, D.~Bahdanau, D.~Serdyuk, K.~Cho, and Y.~Bengio,
  ``Attention-based models for speech recognition,'' \emph{Advances in neural
  information processing systems}, vol.~28, 2015.

\bibitem{mnih2014recurrent}
V.~Mnih, N.~Heess, A.~Graves \emph{et~al.}, ``Recurrent models of visual
  attention,'' \emph{Advances in neural information processing systems},
  vol.~27, 2014.

\bibitem{hu2020introductory}
D.~Hu, ``An introductory survey on attention mechanisms in nlp problems,'' in
  \emph{Intelligent Systems and Applications: Proceedings of the 2019
  Intelligent Systems Conference (IntelliSys) Volume 2}.\hskip 1em plus 0.5em
  minus 0.4em\relax Springer, 2020, pp. 432--448.

\bibitem{vaswani2017attention}
A.~Vaswani, N.~Shazeer, N.~Parmar, J.~Uszkoreit, L.~Jones, A.~N. Gomez,
  {\L}.~Kaiser, and I.~Polosukhin, ``Attention is all you need,''
  \emph{Advances in neural information processing systems}, vol.~30, 2017.

\bibitem{han2022few}
G.~Han, J.~Ma, S.~Huang, L.~Chen, and S.-F. Chang, ``Few-shot object detection
  with fully cross-transformer,'' in \emph{Proceedings of the IEEE/CVF
  conference on computer vision and pattern recognition}, 2022, pp. 5321--5330.

\bibitem{cheng2022masked}
B.~Cheng, I.~Misra, A.~G. Schwing, A.~Kirillov, and R.~Girdhar,
  ``Masked-attention mask transformer for universal image segmentation,'' in
  \emph{Proceedings of the IEEE/CVF Conference on Computer Vision and Pattern
  Recognition}, 2022, pp. 1290--1299.

\bibitem{tran2022combination}
T.-D. Tran, X.-T. Vo, D.-L. Nguyen, and K.-H. Jo, ``Combination of deep learner
  network and transformer for 3d human pose estimation,'' in \emph{2022 22nd
  International Conference on Control, Automation and Systems (ICCAS)}.\hskip
  1em plus 0.5em minus 0.4em\relax IEEE, 2022, pp. 174--178.

\bibitem{weng2022whose}
X.~Weng, B.~Ivanovic, K.~Kitani, and M.~Pavone, ``Whose track is it anyway?
  improving robustness to tracking errors with affinity-based trajectory
  prediction,'' in \emph{Proceedings of the IEEE/CVF Conference on Computer
  Vision and Pattern Recognition}, 2022, pp. 6573--6582.

\bibitem{bracsoveanu2020visualizing}
A.~M. Bra{\c{s}}oveanu and R.~Andonie, ``Visualizing transformers for nlp: a
  brief survey,'' in \emph{2020 24th International Conference Information
  Visualisation (IV)}.\hskip 1em plus 0.5em minus 0.4em\relax IEEE, 2020, pp.
  270--279.

\bibitem{teeti2022vision}
I.~Teeti, S.~Khan, A.~Shahbaz, A.~Bradley, and F.~Cuzzolin, ``Vision-based
  intention and trajectory prediction in autonomous vehicles: A survey,'' in
  \emph{Proceedings of the Thirty-First International Joint Conference on
  Artificial Intelligence, IJCAI-22, Lud De Raedt, Ed}, vol.~7, 2022, pp.
  5630--5637.

\bibitem{cui2019multimodal}
H.~Cui, V.~Radosavljevic, F.-C. Chou, T.-H. Lin, T.~Nguyen, T.-K. Huang,
  J.~Schneider, and N.~Djuric, ``Multimodal trajectory predictions for
  autonomous driving using deep convolutional networks,'' in \emph{2019
  International Conference on Robotics and Automation (ICRA)}.\hskip 1em plus
  0.5em minus 0.4em\relax IEEE, 2019, pp. 2090--2096.

\bibitem{sandler2018mobilenetv2}
M.~Sandler, A.~Howard, M.~Zhu, A.~Zhmoginov, and L.-C. Chen, ``Mobilenetv2:
  Inverted residuals and linear bottlenecks,'' in \emph{Proceedings of the IEEE
  conference on computer vision and pattern recognition}, 2018, pp. 4510--4520.

\bibitem{djuric2020uncertainty}
N.~Djuric, V.~Radosavljevic, H.~Cui, T.~Nguyen, F.-C. Chou, T.-H. Lin,
  N.~Singh, and J.~Schneider, ``Uncertainty-aware short-term motion prediction
  of traffic actors for autonomous driving,'' in \emph{Proceedings of the
  IEEE/CVF Winter Conference on Applications of Computer Vision}, 2020, pp.
  2095--2104.

\bibitem{phan2020covernet}
T.~Phan-Minh, E.~C. Grigore, F.~A. Boulton, O.~Beijbom, and E.~M. Wolff,
  ``Covernet: Multimodal behavior prediction using trajectory sets,'' in
  \emph{Proceedings of the IEEE/CVF Conference on Computer Vision and Pattern
  Recognition}, 2020, pp. 14\,074--14\,083.

\bibitem{targ2016resnet}
S.~Targ, D.~Almeida, and K.~Lyman, ``Resnet in resnet: Generalizing residual
  architectures,'' \emph{arXiv preprint arXiv:1603.08029}, 2016.

\bibitem{marchetti2020mantra}
F.~Marchetti, F.~Becattini, L.~Seidenari, and A.~D. Bimbo, ``Mantra: Memory
  augmented networks for multiple trajectory prediction,'' in \emph{Proceedings
  of the IEEE/CVF conference on computer vision and pattern recognition}, 2020,
  pp. 7143--7152.

\bibitem{zhang2021resnet}
Z.~Zhang, ``Resnet-based model for autonomous vehicles trajectory prediction,''
  in \emph{2021 IEEE International Conference on Consumer Electronics and
  Computer Engineering (ICCECE)}.\hskip 1em plus 0.5em minus 0.4em\relax IEEE,
  2021, pp. 565--568.

\bibitem{li2019grip}
X.~Li, X.~Ying, and M.~C. Chuah, ``Grip: Graph-based interaction-aware
  trajectory prediction,'' in \emph{2019 IEEE Intelligent Transportation
  Systems Conference (ITSC)}.\hskip 1em plus 0.5em minus 0.4em\relax IEEE,
  2019, pp. 3960--3966.

\bibitem{li2019grip++}
------, ``Grip++: Enhanced graph-based interaction-aware trajectory prediction
  for autonomous driving,'' \emph{arXiv preprint arXiv:1907.07792}, 2019.

\bibitem{jeon2020scale}
H.~Jeon, J.~Choi, and D.~Kum, ``Scale-net: Scalable vehicle trajectory
  prediction network under random number of interacting vehicles via
  edge-enhanced graph convolutional neural network,'' in \emph{2020 IEEE/RSJ
  International Conference on Intelligent Robots and Systems (IROS)}.\hskip 1em
  plus 0.5em minus 0.4em\relax IEEE, 2020, pp. 2095--2102.

\bibitem{chandra2020forecasting}
R.~Chandra, T.~Guan, S.~Panuganti, T.~Mittal, U.~Bhattacharya, A.~Bera, and
  D.~Manocha, ``Forecasting trajectory and behavior of road-agents using
  spectral clustering in graph-lstms,'' \emph{IEEE Robotics and Automation
  Letters}, vol.~5, no.~3, pp. 4882--4890, 2020.

\bibitem{zhao2020gisnet}
Z.~Zhao, H.~Fang, Z.~Jin, and Q.~Qiu, ``Gisnet: Graph-based information sharing
  network for vehicle trajectory prediction,'' in \emph{2020 International
  Joint Conference on Neural Networks (IJCNN)}.\hskip 1em plus 0.5em minus
  0.4em\relax IEEE, 2020, pp. 1--7.

\bibitem{sheng2022graph}
Z.~Sheng, Y.~Xu, S.~Xue, and D.~Li, ``Graph-based spatial-temporal
  convolutional network for vehicle trajectory prediction in autonomous
  driving,'' \emph{IEEE Transactions on Intelligent Transportation Systems},
  vol.~23, no.~10, pp. 17\,654--17\,665, 2022.

\bibitem{xu2022group}
D.~Xu, X.~Shang, Y.~Liu, H.~Peng, and H.~Li, ``Group vehicle trajectory
  prediction with global spatio-temporal graph,'' \emph{IEEE Transactions on
  Intelligent Vehicles}, 2022.

\bibitem{xu2023mvhgn}
D.~Xu, X.~Shang, H.~Peng, and H.~Li, ``Mvhgn: Multi-view adaptive hierarchical
  spatial graph convolution network based trajectory prediction for
  heterogeneous traffic-agents,'' \emph{IEEE Transactions on Intelligent
  Transportation Systems}, 2023.

\bibitem{gehring2016convolutional}
J.~Gehring, M.~Auli, D.~Grangier, and Y.~N. Dauphin, ``A convolutional encoder
  model for neural machine translation,'' \emph{arXiv preprint
  arXiv:1611.02344}, 2016.

\bibitem{bhatt2021cnn}
D.~Bhatt, C.~Patel, H.~Talsania, J.~Patel, R.~Vaghela, S.~Pandya, K.~Modi, and
  H.~Ghayvat, ``Cnn variants for computer vision: history, architecture,
  application, challenges and future scope,'' \emph{Electronics}, vol.~10,
  no.~20, p. 2470, 2021.

\bibitem{nikhil2018convolutional}
N.~Nikhil and B.~Tran~Morris, ``Convolutional neural network for trajectory
  prediction,'' in \emph{Proceedings of the European Conference on Computer
  Vision (ECCV) Workshops}, 2018, pp. 0--0.

\bibitem{chou2020predicting}
F.-C. Chou, T.-H. Lin, H.~Cui, V.~Radosavljevic, T.~Nguyen, T.-K. Huang,
  M.~Niedoba, J.~Schneider, and N.~Djuric, ``Predicting motion of vulnerable
  road users using high-definition maps and efficient convnets,'' in \emph{2020
  IEEE Intelligent Vehicles Symposium (IV)}.\hskip 1em plus 0.5em minus
  0.4em\relax IEEE, 2020, pp. 1655--1662.

\bibitem{gilles2021home}
T.~Gilles, S.~Sabatini, D.~Tsishkou, B.~Stanciulescu, and F.~Moutarde, ``Home:
  Heatmap output for future motion estimation,'' in \emph{2021 IEEE
  International Intelligent Transportation Systems Conference (ITSC)}.\hskip
  1em plus 0.5em minus 0.4em\relax IEEE, 2021, pp. 500--507.

\bibitem{ye2021tpcn}
M.~Ye, T.~Cao, and Q.~Chen, ``Tpcn: Temporal point cloud networks for motion
  forecasting,'' in \emph{Proceedings of the IEEE/CVF Conference on Computer
  Vision and Pattern Recognition}, 2021, pp. 11\,318--11\,327.

\bibitem{mo2020recog}
X.~Mo, Y.~Xing, and C.~Lv, ``Recog: A deep learning framework with
  heterogeneous graph for interaction-aware trajectory prediction,''
  \emph{arXiv preprint arXiv:2012.05032}, 2020.

\bibitem{ding2021ra}
Z.~Ding, Z.~Yao, and H.~Zhao, ``Ra-gat: Repulsion and attraction graph
  attention for trajectory prediction,'' in \emph{2021 IEEE International
  Intelligent Transportation Systems Conference (ITSC)}.\hskip 1em plus 0.5em
  minus 0.4em\relax IEEE, 2021, pp. 734--741.

\bibitem{mo2022multi}
X.~Mo, Z.~Huang, Y.~Xing, and C.~Lv, ``Multi-agent trajectory prediction with
  heterogeneous edge-enhanced graph attention network,'' \emph{IEEE
  Transactions on Intelligent Transportation Systems}, vol.~23, no.~7, pp.
  9554--9567, 2022.

\bibitem{liu2022multi}
Y.~Liu, X.~Qi, E.~A. Sisbot, and K.~Oguchi, ``Multi-agent trajectory prediction
  with graph attention isomorphism neural network,'' in \emph{2022 IEEE
  Intelligent Vehicles Symposium (IV)}.\hskip 1em plus 0.5em minus 0.4em\relax
  IEEE, 2022, pp. 273--279.

\bibitem{zhang2022ai}
K.~Zhang, L.~Zhao, C.~Dong, L.~Wu, and L.~Zheng, ``Ai-tp: Attention-based
  interaction-aware trajectory prediction for autonomous driving,'' \emph{IEEE
  Transactions on Intelligent Vehicles}, 2022.

\bibitem{zhang2022trajectory}
K.~Zhang, X.~Feng, L.~Wu, and Z.~He, ``Trajectory prediction for autonomous
  driving using spatial-temporal graph attention transformer,'' \emph{IEEE
  Transactions on Intelligent Transportation Systems}, vol.~23, no.~11, pp.
  22\,343--22\,353, 2022.

\bibitem{meng2023trajectory}
Q.~Meng, H.~Guo, Y.~Liu, H.~Chen, and D.~Cao, ``Trajectory prediction for
  automated vehicles on roads with lanes partially covered by ice or snow,''
  \emph{IEEE Transactions on Vehicular Technology}, 2023.

\bibitem{wu2020comprehensive}
Z.~Wu, S.~Pan, F.~Chen, G.~Long, C.~Zhang, and S.~Y. Philip, ``A comprehensive
  survey on graph neural networks,'' \emph{IEEE transactions on neural networks
  and learning systems}, vol.~32, no.~1, pp. 4--24, 2020.

\bibitem{diehl2019graph}
F.~Diehl, T.~Brunner, M.~T. Le, and A.~Knoll, ``Graph neural networks for
  modelling traffic participant interaction,'' in \emph{2019 IEEE Intelligent
  Vehicles Symposium (IV)}.\hskip 1em plus 0.5em minus 0.4em\relax IEEE, 2019,
  pp. 695--701.

\bibitem{huang2019apolloscape}
X.~Huang, P.~Wang, X.~Cheng, D.~Zhou, Q.~Geng, and R.~Yang, ``The apolloscape
  open dataset for autonomous driving and its application,'' \emph{IEEE
  transactions on pattern analysis and machine intelligence}, vol.~42, no.~10,
  pp. 2702--2719, 2019.

\bibitem{gong2019exploiting}
L.~Gong and Q.~Cheng, ``Exploiting edge features for graph neural networks,''
  in \emph{Proceedings of the IEEE/CVF conference on computer vision and
  pattern recognition}, 2019, pp. 9211--9219.

\bibitem{gao2020vectornet}
J.~Gao, C.~Sun, H.~Zhao, Y.~Shen, D.~Anguelov, C.~Li, and C.~Schmid,
  ``Vectornet: Encoding hd maps and agent dynamics from vectorized
  representation,'' in \emph{Proceedings of the IEEE/CVF Conference on Computer
  Vision and Pattern Recognition}, 2020, pp. 11\,525--11\,533.

\bibitem{liang2020learning}
M.~Liang, B.~Yang, R.~Hu, Y.~Chen, R.~Liao, S.~Feng, and R.~Urtasun, ``Learning
  lane graph representations for motion forecasting,'' in \emph{Computer
  Vision--ECCV 2020: 16th European Conference, Glasgow, UK, August 23--28,
  2020, Proceedings, Part II 16}.\hskip 1em plus 0.5em minus 0.4em\relax
  Springer, 2020, pp. 541--556.

\bibitem{zhao2021tnt}
H.~Zhao, J.~Gao, T.~Lan, C.~Sun, B.~Sapp, B.~Varadarajan, Y.~Shen, Y.~Shen,
  Y.~Chai, C.~Schmid \emph{et~al.}, ``Tnt: Target-driven trajectory
  prediction,'' in \emph{Conference on Robot Learning}.\hskip 1em plus 0.5em
  minus 0.4em\relax PMLR, 2021, pp. 895--904.

\bibitem{gu2021densetnt}
J.~Gu, C.~Sun, and H.~Zhao, ``Densetnt: End-to-end trajectory prediction from
  dense goal sets,'' in \emph{Proceedings of the IEEE/CVF International
  Conference on Computer Vision}, 2021, pp. 15\,303--15\,312.

\bibitem{zeng2021lanercnn}
W.~Zeng, M.~Liang, R.~Liao, and R.~Urtasun, ``Lanercnn: Distributed
  representations for graph-centric motion forecasting,'' in \emph{2021
  IEEE/RSJ International Conference on Intelligent Robots and Systems
  (IROS)}.\hskip 1em plus 0.5em minus 0.4em\relax IEEE, 2021, pp. 532--539.

\bibitem{veličković2018graph}
P.~Veličković, G.~Cucurull, A.~Casanova, A.~Romero, P.~Liò, and Y.~Bengio,
  ``Graph attention networks,'' 2018.

\bibitem{ziegler2014making}
J.~Ziegler, P.~Bender, M.~Schreiber, H.~Lategahn, T.~Strauss, C.~Stiller,
  T.~Dang, U.~Franke, N.~Appenrodt, C.~G. Keller \emph{et~al.}, ``Making bertha
  drive—an autonomous journey on a historic route,'' \emph{IEEE Intelligent
  transportation systems magazine}, vol.~6, no.~2, pp. 8--20, 2014.

\bibitem{schreiber2019long}
M.~Schreiber, S.~Hoermann, and K.~Dietmayer, ``Long-term occupancy grid
  prediction using recurrent neural networks,'' in \emph{2019 International
  Conference on Robotics and Automation (ICRA)}.\hskip 1em plus 0.5em minus
  0.4em\relax IEEE, 2019, pp. 9299--9305.

\bibitem{chandra2019traphic}
R.~Chandra, U.~Bhattacharya, A.~Bera, and D.~Manocha, ``Traphic: Trajectory
  prediction in dense and heterogeneous traffic using weighted interactions,''
  in \emph{Proceedings of the IEEE/CVF Conference on Computer Vision and
  Pattern Recognition}, 2019, pp. 8483--8492.

\bibitem{xie2020motion}
G.~Xie, A.~Shangguan, R.~Fei, W.~Ji, W.~Ma, and X.~Hei, ``Motion trajectory
  prediction based on a cnn-lstm sequential model,'' \emph{Science China
  Information Sciences}, vol.~63, pp. 1--21, 2020.

\bibitem{xu4135360vehicle}
X.~Xu, X.~Wang, Y.~Wang, and J.~Cao, ``Vehicle trajectory prediction
  considering multi-feature independent encoding,'' \emph{Available at SSRN
  4135360}.

\bibitem{hegde2020vehicle}
C.~Hegde, S.~Dash, and P.~Agarwal, ``Vehicle trajectory prediction using gan,''
  in \emph{2020 Fourth International Conference on I-SMAC (IoT in Social,
  Mobile, Analytics and Cloud)(I-SMAC)}.\hskip 1em plus 0.5em minus 0.4em\relax
  IEEE, 2020, pp. 502--507.

\bibitem{zhao2020novel}
L.~Zhao, Y.~Liu, A.~Y. Al-Dubai, A.~Y. Zomaya, G.~Min, and A.~Hawbani, ``A
  novel generation-adversarial-network-based vehicle trajectory prediction
  method for intelligent vehicular networks,'' \emph{IEEE Internet of Things
  Journal}, vol.~8, no.~3, pp. 2066--2077, 2020.

\bibitem{li2021vehicle}
X.~Li, G.~Rosman, I.~Gilitschenski, C.-I. Vasile, J.~A. DeCastro, S.~Karaman,
  and D.~Rus, ``Vehicle trajectory prediction using generative adversarial
  network with temporal logic syntax tree features,'' \emph{IEEE Robotics and
  Automation Letters}, vol.~6, no.~2, pp. 3459--3466, 2021.

\bibitem{guo2023map}
H.~Guo, Q.~Meng, X.~Zhao, J.~Liu, D.~Cao, and H.~Chen, ``Map-enhanced
  generative adversarial trajectory prediction method for automated vehicles,''
  \emph{Information Sciences}, vol. 622, pp. 1033--1049, 2023.

\bibitem{wang2020multi}
Y.~Wang, S.~Zhao, R.~Zhang, X.~Cheng, and L.~Yang, ``Multi-vehicle
  collaborative learning for trajectory prediction with spatio-temporal tensor
  fusion,'' \emph{IEEE Transactions on Intelligent Transportation Systems},
  vol.~23, no.~1, pp. 236--248, 2020.

\bibitem{wang2021multi}
Y.~Wang and S.~Chen, ``Multi-agent trajectory prediction with spatio-temporal
  sequence fusion,'' \emph{IEEE Transactions on Multimedia}, 2021.

\bibitem{zhou2018unet++}
Z.~Zhou, M.~M. Rahman~Siddiquee, N.~Tajbakhsh, and J.~Liang, ``Unet++: A nested
  u-net architecture for medical image segmentation,'' in \emph{Deep Learning
  in Medical Image Analysis and Multimodal Learning for Clinical Decision
  Support: 4th International Workshop, DLMIA 2018, and 8th International
  Workshop, ML-CDS 2018, Held in Conjunction with MICCAI 2018, Granada, Spain,
  September 20, 2018, Proceedings 4}.\hskip 1em plus 0.5em minus 0.4em\relax
  Springer, 2018, pp. 3--11.

\bibitem{cho2019deep}
K.~Cho, T.~Ha, G.~Lee, and S.~Oh, ``Deep predictive autonomous driving using
  multi-agent joint trajectory prediction and traffic rules,'' in \emph{2019
  IEEE/RSJ International Conference on Intelligent Robots and Systems
  (IROS)}.\hskip 1em plus 0.5em minus 0.4em\relax IEEE, 2019, pp. 2076--2081.

\bibitem{hu2019multi}
Y.~Hu, W.~Zhan, L.~Sun, and M.~Tomizuka, ``Multi-modal probabilistic prediction
  of interactive behavior via an interpretable model,'' in \emph{2019 IEEE
  Intelligent Vehicles Symposium (IV)}.\hskip 1em plus 0.5em minus 0.4em\relax
  IEEE, 2019, pp. 557--563.

\bibitem{zhang2020multimodal}
H.~Zhang, Y.~Wang, J.~Liu, C.~Li, T.~Ma, and C.~Yin, ``A multi-modal states
  based vehicle descriptor and dilated convolutional social pooling for vehicle
  trajectory prediction,'' 2020.

\bibitem{sriram2020smart}
N.~Sriram, B.~Liu, F.~Pittaluga, and M.~Chandraker, ``Smart: Simultaneous
  multi-agent recurrent trajectory prediction,'' in \emph{Computer Vision--ECCV
  2020: 16th European Conference, Glasgow, UK, August 23--28, 2020,
  Proceedings, Part XXVII 16}.\hskip 1em plus 0.5em minus 0.4em\relax Springer,
  2020, pp. 463--479.

\bibitem{dulian2021multi}
A.~Dulian and J.~C. Murray, ``Multi-modal anticipation of stochastic
  trajectories in a dynamic environment with conditional variational
  autoencoders,'' \emph{arXiv preprint arXiv:2103.03912}, 2021.

\bibitem{liu2022interactive}
X.~Liu, Y.~Wang, K.~Jiang, Z.~Zhou, K.~Nam, and C.~Yin, ``Interactive
  trajectory prediction using a driving risk map-integrated deep learning
  method for surrounding vehicles on highways,'' \emph{IEEE Transactions on
  Intelligent Transportation Systems}, vol.~23, no.~10, pp. 19\,076--19\,087,
  2022.

\bibitem{wang2017generative}
K.~Wang, C.~Gou, Y.~Duan, Y.~Lin, X.~Zheng, and F.-Y. Wang, ``Generative
  adversarial networks: introduction and outlook,'' \emph{IEEE/CAA Journal of
  Automatica Sinica}, vol.~4, no.~4, pp. 588--598, 2017.

\bibitem{song2022learning}
H.~Song, D.~Luan, W.~Ding, M.~Y. Wang, and Q.~Chen, ``Learning to predict
  vehicle trajectories with model-based planning,'' in \emph{Conference on
  Robot Learning}.\hskip 1em plus 0.5em minus 0.4em\relax PMLR, 2022, pp.
  1035--1045.

\bibitem{wang2018reinforcement}
P.~Wang, C.-Y. Chan, and A.~de~La~Fortelle, ``A reinforcement learning based
  approach for automated lane change maneuvers,'' in \emph{2018 IEEE
  Intelligent Vehicles Symposium (IV)}.\hskip 1em plus 0.5em minus 0.4em\relax
  IEEE, 2018, pp. 1379--1384.

\bibitem{guan2018markov}
Y.~Guan, S.~E. Li, J.~Duan, W.~Wang, and B.~Cheng, ``Markov probabilistic
  decision making of self-driving cars in highway with random traffic flow: a
  simulation study,'' \emph{Journal of Intelligent and Connected Vehicles},
  vol.~1, no.~2, pp. 77--84, 2018.

\bibitem{zou2018inverse}
Q.~Zou, H.~Li, and R.~Zhang, ``Inverse reinforcement learning via neural
  network in driver behavior modeling,'' in \emph{2018 IEEE Intelligent
  Vehicles Symposium (IV)}.\hskip 1em plus 0.5em minus 0.4em\relax IEEE, 2018,
  pp. 1245--1250.

\bibitem{sun2018probabilistic}
L.~Sun, W.~Zhan, and M.~Tomizuka, ``Probabilistic prediction of interactive
  driving behavior via hierarchical inverse reinforcement learning,'' in
  \emph{2018 21st International Conference on Intelligent Transportation
  Systems (ITSC)}.\hskip 1em plus 0.5em minus 0.4em\relax IEEE, 2018, pp.
  2111--2117.

\bibitem{gonzalez2018modeling}
D.~S. Gonz{\'a}lez, O.~Erkent, V.~Romero-Cano, J.~Dibangoye, and C.~Laugier,
  ``Modeling driver behavior from demonstrations in dynamic environments using
  spatiotemporal lattices,'' in \emph{2018 IEEE International Conference on
  Robotics and Automation (ICRA)}.\hskip 1em plus 0.5em minus 0.4em\relax IEEE,
  2018, pp. 3384--3390.

\bibitem{xin2019accelerated}
L.~Xin, S.~E. Li, P.~Wang, W.~Cao, B.~Nie, C.-Y. Chan, and B.~Cheng,
  ``Accelerated inverse reinforcement learning with randomly pre-sampled
  policies for autonomous driving reward design,'' in \emph{2019 IEEE
  Intelligent Transportation Systems Conference (ITSC)}.\hskip 1em plus 0.5em
  minus 0.4em\relax IEEE, 2019, pp. 2757--2764.

\bibitem{xu2019learning}
Y.~Xu, T.~Zhao, C.~Baker, Y.~Zhao, and Y.~N. Wu, ``Learning trajectory
  prediction with continuous inverse optimal control via langevin sampling of
  energy-based models,'' \emph{arXiv preprint arXiv:1904.05453}, 2019.

\bibitem{xu2020learning}
D.~Xu, Z.~Ding, X.~He, H.~Zhao, M.~Moze, F.~Aioun, and F.~Guillemard,
  ``Learning from naturalistic driving data for human-like autonomous highway
  driving,'' \emph{IEEE Transactions on Intelligent Transportation Systems},
  vol.~22, no.~12, pp. 7341--7354, 2020.

\bibitem{wu2020efficient}
Z.~Wu, L.~Sun, W.~Zhan, C.~Yang, and M.~Tomizuka, ``Efficient sampling-based
  maximum entropy inverse reinforcement learning with application to autonomous
  driving,'' \emph{IEEE Robotics and Automation Letters}, vol.~5, no.~4, pp.
  5355--5362, 2020.

\bibitem{deo2020trajectory}
N.~Deo and M.~M. Trivedi, ``Trajectory forecasts in unknown environments
  conditioned on grid-based plans,'' \emph{arXiv preprint arXiv:2001.00735},
  2020.

\bibitem{zhu2020off}
Z.~Zhu, N.~Li, R.~Sun, D.~Xu, and H.~Zhao, ``Off-road autonomous vehicles
  traversability analysis and trajectory planning based on deep inverse
  reinforcement learning,'' in \emph{2020 IEEE Intelligent Vehicles Symposium
  (IV)}.\hskip 1em plus 0.5em minus 0.4em\relax IEEE, 2020, pp. 971--977.

\bibitem{jung2021incorporating}
C.~Jung and D.~H. Shim, ``Incorporating multi-context into the traversability
  map for urban autonomous driving using deep inverse reinforcement learning,''
  \emph{IEEE Robotics and Automation Letters}, vol.~6, no.~2, pp. 1662--1669,
  2021.

\bibitem{choi2021trajgail}
S.~Choi, J.~Kim, and H.~Yeo, ``Trajgail: Generating urban vehicle trajectories
  using generative adversarial imitation learning,'' \emph{Transportation
  Research Part C: Emerging Technologies}, vol. 128, p. 103091, 2021.

\bibitem{cheng2022mpnp}
J.~Cheng, R.~Xin, S.~Wang, and M.~Liu, ``Mpnp: Multi-policy neural planner for
  urban driving,'' in \emph{2022 IEEE/RSJ International Conference on
  Intelligent Robots and Systems (IROS)}.\hskip 1em plus 0.5em minus
  0.4em\relax IEEE, 2022, pp. 10\,549--10\,554.

\bibitem{phan2022driving}
T.~Phan-Minh, F.~Howington, T.-S. Chu, S.~U. Lee, M.~S. Tomov, N.~Li, C.~Dicle,
  S.~Findler, F.~Suarez-Ruiz, R.~Beaudoin \emph{et~al.}, ``Driving in real life
  with inverse reinforcement learning,'' \emph{arXiv preprint
  arXiv:2206.03004}, 2022.

\bibitem{bronstein2022hierarchical}
E.~Bronstein, M.~Palatucci, D.~Notz, B.~White, A.~Kuefler, Y.~Lu, S.~Paul,
  P.~Nikdel, P.~Mougin, H.~Chen \emph{et~al.}, ``Hierarchical model-based
  imitation learning for planning in autonomous driving,'' in \emph{2022
  IEEE/RSJ International Conference on Intelligent Robots and Systems
  (IROS)}.\hskip 1em plus 0.5em minus 0.4em\relax IEEE, 2022, pp. 8652--8659.

\bibitem{hjaltason2019predicting}
B.~Hjaltason, ``Predicting vehicle trajectories with inverse reinforcement
  learning,'' 2019.

\bibitem{kiran2021deep}
B.~R. Kiran, I.~Sobh, V.~Talpaert, P.~Mannion, A.~A. Al~Sallab, S.~Yogamani,
  and P.~P{\'e}rez, ``Deep reinforcement learning for autonomous driving: A
  survey,'' \emph{IEEE Transactions on Intelligent Transportation Systems},
  vol.~23, no.~6, pp. 4909--4926, 2021.

\bibitem{fernando2020deep}
T.~Fernando, S.~Denman, S.~Sridharan, and C.~Fookes, ``Deep inverse
  reinforcement learning for behavior prediction in autonomous driving:
  Accurate forecasts of vehicle motion,'' \emph{IEEE Signal Processing
  Magazine}, vol.~38, no.~1, pp. 87--96, 2020.

\bibitem{abbeel2004apprenticeship}
P.~Abbeel and A.~Y. Ng, ``Apprenticeship learning via inverse reinforcement
  learning,'' in \emph{Proceedings of the twenty-first international conference
  on Machine learning}, 2004, p.~1.

\bibitem{wulfmeier2015maximum}
M.~Wulfmeier, P.~Ondruska, and I.~Posner, ``Maximum entropy deep inverse
  reinforcement learning,'' \emph{arXiv preprint arXiv:1507.04888}, 2015.

\bibitem{wulfmeier2016watch}
M.~Wulfmeier, D.~Z. Wang, and I.~Posner, ``Watch this: Scalable cost-function
  learning for path planning in urban environments,'' in \emph{2016 IEEE/RSJ
  International Conference on Intelligent Robots and Systems (IROS)}.\hskip 1em
  plus 0.5em minus 0.4em\relax IEEE, 2016, pp. 2089--2095.

\bibitem{pomerleau1988alvinn}
D.~A. Pomerleau, ``Alvinn: An autonomous land vehicle in a neural network,''
  \emph{Advances in neural information processing systems}, vol.~1, 1988.

\bibitem{devi2020behaviour}
T.~K. Devi, A.~Srivatsava, K.~K. Mudgal, R.~R. Jayanti, and T.~Karthick,
  ``Behaviour cloning for autonomous driving.'' \emph{Webology}, vol.~17,
  no.~2, pp. 694--705, 2020.

\bibitem{ho2016generative}
J.~Ho and S.~Ermon, ``Generative adversarial imitation learning,''
  \emph{Advances in neural information processing systems}, vol.~29, 2016.

\bibitem{somani2013despot}
A.~Somani, N.~Ye, D.~Hsu, and W.~S. Lee, ``Despot: Online pomdp planning with
  regularization,'' \emph{Advances in neural information processing systems},
  vol.~26, 2013.

\bibitem{silver2010monte}
D.~Silver and J.~Veness, ``Monte-carlo planning in large pomdps,''
  \emph{Advances in neural information processing systems}, vol.~23, 2010.

\bibitem{kuefler2017imitating}
A.~Kuefler, J.~Morton, T.~Wheeler, and M.~Kochenderfer, ``Imitating driver
  behavior with generative adversarial networks,'' in \emph{2017 IEEE
  Intelligent Vehicles Symposium (IV)}.\hskip 1em plus 0.5em minus 0.4em\relax
  IEEE, 2017, pp. 204--211.

\bibitem{bansal2018chauffeurnet}
M.~Bansal, A.~Krizhevsky, and A.~Ogale, ``Chauffeurnet: Learning to drive by
  imitating the best and synthesizing the worst,'' \emph{arXiv preprint
  arXiv:1812.03079}, 2018.

\bibitem{caesar2020nuscenes}
H.~Caesar, V.~Bankiti, A.~H. Lang, S.~Vora, V.~E. Liong, Q.~Xu, A.~Krishnan,
  Y.~Pan, G.~Baldan, and O.~Beijbom, ``nuscenes: A multimodal dataset for
  autonomous driving,'' in \emph{Proceedings of the IEEE/CVF conference on
  computer vision and pattern recognition}, 2020, pp. 11\,621--11\,631.

\bibitem{houston2021one}
J.~Houston, G.~Zuidhof, L.~Bergamini, Y.~Ye, L.~Chen, A.~Jain, S.~Omari,
  V.~Iglovikov, and P.~Ondruska, ``One thousand and one hours: Self-driving
  motion prediction dataset,'' in \emph{Conference on Robot Learning}.\hskip
  1em plus 0.5em minus 0.4em\relax PMLR, 2021, pp. 409--418.

\bibitem{ettinger2021large}
S.~Ettinger, S.~Cheng, B.~Caine, C.~Liu, H.~Zhao, S.~Pradhan, Y.~Chai, B.~Sapp,
  C.~R. Qi, Y.~Zhou \emph{et~al.}, ``Large scale interactive motion forecasting
  for autonomous driving: The waymo open motion dataset,'' in \emph{Proceedings
  of the IEEE/CVF International Conference on Computer Vision}, 2021, pp.
  9710--9719.

\bibitem{wilson2023argoverse}
B.~Wilson, W.~Qi, T.~Agarwal, J.~Lambert, J.~Singh, S.~Khandelwal, B.~Pan,
  R.~Kumar, A.~Hartnett, J.~K. Pontes \emph{et~al.}, ``Argoverse 2: Next
  generation datasets for self-driving perception and forecasting,''
  \emph{arXiv preprint arXiv:2301.00493}, 2023.

\bibitem{zhan2019interaction}
W.~Zhan, L.~Sun, D.~Wang, H.~Shi, A.~Clausse, M.~Naumann, J.~Kummerle,
  H.~Konigshof, C.~Stiller, A.~de~La~Fortelle \emph{et~al.}, ``Interaction
  dataset: An international, adversarial and cooperative motion dataset in
  interactive driving scenarios with semantic maps,'' \emph{arXiv preprint
  arXiv:1910.03088}, 2019.

\bibitem{krajewski2018highd}
R.~Krajewski, J.~Bock, L.~Kloeker, and L.~Eckstein, ``The highd dataset: A
  drone dataset of naturalistic vehicle trajectories on german highways for
  validation of highly automated driving systems,'' in \emph{2018 21st
  International Conference on Intelligent Transportation Systems (ITSC)}.\hskip
  1em plus 0.5em minus 0.4em\relax IEEE, 2018, pp. 2118--2125.

\bibitem{geiger2013vision}
A.~Geiger, P.~Lenz, C.~Stiller, and R.~Urtasun, ``Vision meets robotics: The
  kitti dataset,'' \emph{The International Journal of Robotics Research},
  vol.~32, no.~11, pp. 1231--1237, 2013.

\bibitem{coifman2017critical}
B.~Coifman and L.~Li, ``A critical evaluation of the next generation simulation
  (ngsim) vehicle trajectory dataset,'' \emph{Transportation Research Part B:
  Methodological}, vol. 105, pp. 362--377, 2017.

\bibitem{chen2017multi}
X.~Chen, H.~Ma, J.~Wan, B.~Li, and T.~Xia, ``Multi-view 3d object detection
  network for autonomous driving,'' in \emph{Proceedings of the IEEE conference
  on Computer Vision and Pattern Recognition}, 2017, pp. 1907--1915.

\bibitem{seif2016autonomous}
H.~G. Seif and X.~Hu, ``Autonomous driving in the icity—hd maps as a key
  challenge of the automotive industry,'' \emph{Engineering}, vol.~2, no.~2,
  pp. 159--162, 2016.

\end{thebibliography}
\begin{IEEEbiography} 
[{\includegraphics[width=1in,height=1in,clip,keepaspectratio]{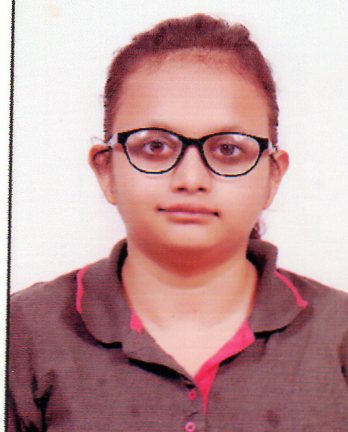}}]{Vibha Bharilya} received his M.Sc degree in Mathematics and Scientific Computing from the National Institute of Technology, Warangal in 2019, and an M.Tech. degree in Computer Science \& Technology from Jawaharlal Nehru University, New Delhi. Currently, she is pursuing Ph.D. in the Department of Computer Science and Engineering from Indian Institute of Technology, Roorkee, India. His research interests include Autonomous Vehicles motion, Autonomous Vehicle Driving, and Self Driving Cars.
\end{IEEEbiography}

\begin{IEEEbiography}[{\includegraphics[width=1in,height=1in,clip,keepaspectratio]{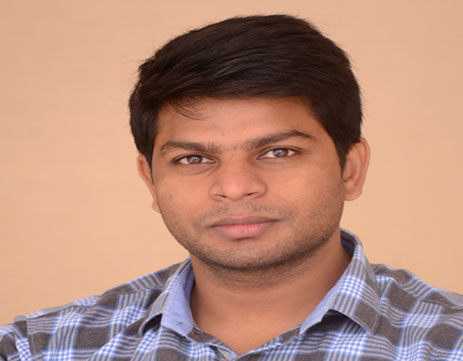}}]{Dr. N. Kumar} received his M.tech and Ph.D. degrees from the School of Computer and Systems Sciences, Jawaharlal Nehru University, New Delhi. He has received Special Mention Awarded Xerox Research Centre Bangalore and several best paper awrads. Currently, he is working as a faculty member at IIT-Roorkee, India. He has published sevral research publications in world’s top tier publishers like IEEE journals and transactions, Elsevier journals (including Future Generation Computer System (FGCS), Information Sciences etc.) and Springer Journals. One of his articles published in IEEE sensors journal has been notified by IEEE council in the list of world’s top 15 most downloaded articles in the month of Oct-Nov 2018. He is also acting as a lead PI for four sponsored projects from IIT Roorkee/DST/CSIR agencies, Government of India, and few proposals are in under process by several agencies. He has also filed four patents individual and in collaboration as well. He has been a technical program committee member in several conferences. He has also been invited as a keynote speaker in the conference held at Amity Gwalior, Conference Chair at IEEE CICT-2019 at IIIT Allahabad, IIT Indore etc. Broadly, his research interests include Algorithm Design, IoT, High Performance Computing (Cloud, Fog and Parallel computing), Applied Evolutionary Computing, Software Defined Networking, WSN and Intelligent Transportation Systems.
\end{IEEEbiography}

\end{document}